\documentclass{article}
\usepackage[eandd, preprint]{neurips_2026}

\usepackage[utf8]{inputenc} %
\usepackage[T1]{fontenc}    %
\usepackage{url}            %
\usepackage{booktabs}       %
\usepackage{nicefrac}       %
\usepackage{microtype}      %
\usepackage{xcolor}         %

\usepackage{graphicx}
\usepackage{subcaption}

\usepackage{amsmath}
\usepackage{amssymb}
\usepackage{mathtools}
\usepackage{amsthm}
\usepackage{amsfonts}       %
\usepackage{paralist}
\usepackage{bbm}

\usepackage[colorlinks=true,linkcolor=gray,citecolor=gray]{hyperref}
\usepackage[capitalize,noabbrev]{cleveref}
\crefname{appsec}{Appendix Section}{Appendix Sections}
\Crefname{appsec}{Appendix Section}{Appendix Sections}
\crefname{appsubsec}{Appendix Section}{Appendix Sections}
\Crefname{appsubsec}{Appendix Section}{Appendix Sections}
\crefname{appsubsubsec}{Appendix Section}{Appendix Sections}
\Crefname{appsubsubsec}{Appendix Section}{Appendix Sections}

\definecolor{tdblue}{HTML}{1F77B4}
\definecolor{reprorange}{HTML}{FF7F0E}

\ExplSyntaxOn

\tl_clear_new:N \l_agent_tl
\tl_clear_new:N \l_hps_tl
\tl_clear_new:N \l_exploreratio_tl
\tl_clear_new:N \l_prefix
\int_new:N \l_num_subfigs_int
\dim_new:N \l_subfig_width_dim

\cs_new_protected:Npn \agent_hp_setting_epsilon: {
  \dim_set:Nn \l_subfig_width_dim { ( \dim_eval:n {1\textwidth / \l_num_subfigs_int} ) }
  \begin{subfigure}[t]{\l_subfig_width_dim}
    \centering
    \includegraphics[width=\textwidth]{\tl_use:N \l_prefix \tl_use:N \l_agent_tl _antmaze-medium-explore\tl_use:N \l_exploreratio_tl navigate-v0_64c_\tl_use:N \l_hps_tl/igpr-(\tl_use:N \l_hps_tl)-normalized_goal_distance_return_epsilon_optimality-1.png}
    \caption{Phase~1}
  \end{subfigure}
  \begin{subfigure}[t]{\l_subfig_width_dim}
    \centering
    \includegraphics[width=\textwidth]{\tl_use:N \l_prefix \tl_use:N \l_agent_tl _antmaze-medium-explore\tl_use:N \l_exploreratio_tl navigate-v0_64c_\tl_use:N \l_hps_tl/igpr-(\tl_use:N \l_hps_tl)-normalized_goal_distance_return_epsilon_optimality-2.png}
    \caption{Phase~2}
  \end{subfigure}
  \begin{subfigure}[t]{\l_subfig_width_dim}
    \centering
    \includegraphics[width=\textwidth]{\tl_use:N \l_prefix \tl_use:N \l_agent_tl _antmaze-medium-explore\tl_use:N \l_exploreratio_tl navigate-v0_64c_\tl_use:N \l_hps_tl/igpr-(\tl_use:N \l_hps_tl)-normalized_goal_distance_return_epsilon_optimality-3.png}
    \caption{Phase~3}
  \end{subfigure}
  \begin{subfigure}[t]{\l_subfig_width_dim}
    \centering
    \includegraphics[width=\textwidth]{\tl_use:N \l_prefix \tl_use:N \l_agent_tl _antmaze-medium-explore\tl_use:N \l_exploreratio_tl navigate-v0_64c_\tl_use:N \l_hps_tl/igpr-(\tl_use:N \l_hps_tl)-normalized_goal_distance_return_epsilon_optimality-4.png}
    \caption{Phase~4}
  \end{subfigure}

}

\cs_new_protected:Npn \agent_hp_setting: {
  \dim_set:Nn \l_subfig_width_dim { ( \dim_eval:n {1\textwidth / \l_num_subfigs_int} ) }
  \begin{subfigure}[t]{\l_subfig_width_dim}
    \centering
    \includegraphics[width=\textwidth]{\tl_use:N \l_prefix \tl_use:N \l_agent_tl _antmaze-medium-explore\tl_use:N \l_exploreratio_tl navigate-v0_64c_\tl_use:N \l_hps_tl/igpr-(\tl_use:N \l_hps_tl)-normalized_goal_distance_return-1.png}
    \caption{Phase~1}
  \end{subfigure}
  \begin{subfigure}[t]{\l_subfig_width_dim}
    \centering
    \includegraphics[width=\textwidth]{\tl_use:N \l_prefix \tl_use:N \l_agent_tl _antmaze-medium-explore\tl_use:N \l_exploreratio_tl navigate-v0_64c_\tl_use:N \l_hps_tl/igpr-(\tl_use:N \l_hps_tl)-normalized_goal_distance_return-2.png}
    \caption{Phase~2}
  \end{subfigure}
  \begin{subfigure}[t]{\l_subfig_width_dim}
    \centering
    \includegraphics[width=\textwidth]{\tl_use:N \l_prefix \tl_use:N \l_agent_tl _antmaze-medium-explore\tl_use:N \l_exploreratio_tl navigate-v0_64c_\tl_use:N \l_hps_tl/igpr-(\tl_use:N \l_hps_tl)-normalized_goal_distance_return-3.png}
    \caption{Phase~3}
  \end{subfigure}
  \begin{subfigure}[t]{\l_subfig_width_dim}
    \centering
    \includegraphics[width=\textwidth]{\tl_use:N \l_prefix \tl_use:N \l_agent_tl _antmaze-medium-explore\tl_use:N \l_exploreratio_tl navigate-v0_64c_\tl_use:N \l_hps_tl/igpr-(\tl_use:N \l_hps_tl)-normalized_goal_distance_return-4.png}
    \caption{Phase~4}
  \end{subfigure}

}

\ExplSyntaxOff

\ExplSyntaxOn

\tl_clear_new:N \l_agent_tl
\tl_clear_new:N \l_hps_tl
\tl_clear_new:N \l_prefix

\tl_clear_new:N \l_dataset_path_tl

\tl_set:Nn \l_dataset_path_tl {
  antmaze-medium-explore100navigate-v0 ~ antmaze-medium-explore80navigate-v0 ~ antmaze-medium-explore40navigate-v0 ~ antmaze-medium-explore0navigate-v0
}

\cs_new_protected:Npn \agent_hp_scheduled_epsilon: {
  \dim_set:Nn \l_subfig_width_dim { ( \dim_eval:n {1\textwidth / \l_num_subfigs_int} ) }
  \begin{subfigure}[t]{\l_subfig_width_dim}
    \centering
    \includegraphics[width=\textwidth]{\tl_use:N \l_prefix \tl_use:N \l_agent_tl _ \tl_use:N \l_dataset_path_tl _64c_\tl_use:N \l_hps_tl/igpr-(\tl_use:N \l_hps_tl)-normalized_goal_distance_return_epsilon_optimality-1.png}
    \caption{Phase~1}
  \end{subfigure}
  \begin{subfigure}[t]{\l_subfig_width_dim}
    \centering
    \includegraphics[width=\textwidth]{\tl_use:N \l_prefix \tl_use:N \l_agent_tl _ \tl_use:N \l_dataset_path_tl _64c_\tl_use:N \l_hps_tl/igpr-(\tl_use:N \l_hps_tl)-normalized_goal_distance_return_epsilon_optimality-2.png}
    \caption{Phase~2}
  \end{subfigure}
  \begin{subfigure}[t]{\l_subfig_width_dim}
    \centering
    \includegraphics[width=\textwidth]{\tl_use:N \l_prefix \tl_use:N \l_agent_tl _ \tl_use:N \l_dataset_path_tl _64c_\tl_use:N \l_hps_tl/igpr-(\tl_use:N \l_hps_tl)-normalized_goal_distance_return_epsilon_optimality-3.png}
    \caption{Phase~3}
  \end{subfigure}
  \begin{subfigure}[t]{\l_subfig_width_dim}
    \centering
    \includegraphics[width=\textwidth]{\tl_use:N \l_prefix \tl_use:N \l_agent_tl _ \tl_use:N \l_dataset_path_tl _64c_\tl_use:N \l_hps_tl/igpr-(\tl_use:N \l_hps_tl)-normalized_goal_distance_return_epsilon_optimality-4.png}
    \caption{Phase~4}
  \end{subfigure}

}

\cs_new_protected:Npn \agent_hp_scheduled: {
  \dim_set:Nn \l_subfig_width_dim { ( \dim_eval:n {1\textwidth / \l_num_subfigs_int} ) }
  \begin{subfigure}[t]{\l_subfig_width_dim}
    \centering
    \includegraphics[width=\textwidth]{\tl_use:N \l_prefix \tl_use:N \l_agent_tl _ \tl_use:N \l_dataset_path_tl _64c_\tl_use:N \l_hps_tl/igpr-(\tl_use:N \l_hps_tl)-normalized_goal_distance_return-1.png}
    \caption{Phase~1}
  \end{subfigure}
  \begin{subfigure}[t]{\l_subfig_width_dim}
    \centering
    \includegraphics[width=\textwidth]{\tl_use:N \l_prefix \tl_use:N \l_agent_tl _ \tl_use:N \l_dataset_path_tl _64c_\tl_use:N \l_hps_tl/igpr-(\tl_use:N \l_hps_tl)-normalized_goal_distance_return-2.png}
    \caption{Phase~2}
  \end{subfigure}
  \begin{subfigure}[t]{\l_subfig_width_dim}
    \centering
    \includegraphics[width=\textwidth]{\tl_use:N \l_prefix \tl_use:N \l_agent_tl _ \tl_use:N \l_dataset_path_tl _64c_\tl_use:N \l_hps_tl/igpr-(\tl_use:N \l_hps_tl)-normalized_goal_distance_return-3.png}
    \caption{Phase~3}
  \end{subfigure}
  \begin{subfigure}[t]{\l_subfig_width_dim}
    \centering
    \includegraphics[width=\textwidth]{\tl_use:N \l_prefix \tl_use:N \l_agent_tl _ \tl_use:N \l_dataset_path_tl _64c_\tl_use:N \l_hps_tl/igpr-(\tl_use:N \l_hps_tl)-normalized_goal_distance_return-4.png}
    \caption{Phase~4}
  \end{subfigure}

}

\ExplSyntaxOff

\ExplSyntaxOn

\tl_clear_new:N \l_agent_tl
\tl_clear_new:N \l_hps_tl
\tl_clear_new:N \l_prefix

\cs_new_protected:Npn \agent_hp_lastphase_epsilon: {
  \dim_set:Nn \l_subfig_width_dim { ( \dim_eval:n {1\textwidth / 5} ) }
  \begin{subfigure}[t]{\l_subfig_width_dim}
    \centering
    \includegraphics[width=\textwidth]{\tl_use:N \l_prefix \tl_use:N \l_agent_tl _antmaze-medium-explore100navigate-v0_64c_\tl_use:N \l_hps_tl/igpr-(\tl_use:N \l_hps_tl)-normalized_goal_distance_return_epsilon_optimality-4.png}
    \caption{100\%\ Noise}
  \end{subfigure}
  \begin{subfigure}[t]{\l_subfig_width_dim}
    \centering
    \includegraphics[width=\textwidth]{\tl_use:N \l_prefix \tl_use:N \l_agent_tl _antmaze-medium-explore90navigate-v0_64c_\tl_use:N \l_hps_tl/igpr-(\tl_use:N \l_hps_tl)-normalized_goal_distance_return_epsilon_optimality-4.png}
    \caption{90\%\ Noise}
  \end{subfigure}
  \begin{subfigure}[t]{\l_subfig_width_dim}
    \centering
    \includegraphics[width=\textwidth]{\tl_use:N \l_prefix \tl_use:N \l_agent_tl _antmaze-medium-explore80navigate-v0_64c_\tl_use:N \l_hps_tl/igpr-(\tl_use:N \l_hps_tl)-normalized_goal_distance_return_epsilon_optimality-4.png}
    \caption{80\%\ Noise}
  \end{subfigure}
  \begin{subfigure}[t]{\l_subfig_width_dim}
    \centering
    \includegraphics[width=\textwidth]{\tl_use:N \l_prefix \tl_use:N \l_agent_tl _antmaze-medium-explore40navigate-v0_64c_\tl_use:N \l_hps_tl/igpr-(\tl_use:N \l_hps_tl)-normalized_goal_distance_return_epsilon_optimality-4.png}
    \caption{40\%\ Noise}
  \end{subfigure}
  \begin{subfigure}[t]{\l_subfig_width_dim}
    \centering
    \includegraphics[width=\textwidth]{\tl_use:N \l_prefix \tl_use:N \l_agent_tl _antmaze-medium-explore0navigate-v0_64c_\tl_use:N \l_hps_tl/igpr-(\tl_use:N \l_hps_tl)-normalized_goal_distance_return_epsilon_optimality-4.png}
    \caption{0\%\ Noise}
  \end{subfigure}
}

\cs_new_protected:Npn \agent_hp_lastphase: {
  \dim_set:Nn \l_subfig_width_dim { ( \dim_eval:n {1\textwidth / 5} ) }
  \begin{subfigure}[t]{\l_subfig_width_dim}
    \centering
    \includegraphics[width=\textwidth]{\tl_use:N \l_prefix \tl_use:N \l_agent_tl _antmaze-medium-explore100navigate-v0_64c_\tl_use:N \l_hps_tl/igpr-(\tl_use:N \l_hps_tl)-normalized_goal_distance_return-4.png}
    \caption{100\%\ Noise}
  \end{subfigure}
  \begin{subfigure}[t]{\l_subfig_width_dim}
    \centering
    \includegraphics[width=\textwidth]{\tl_use:N \l_prefix \tl_use:N \l_agent_tl _antmaze-medium-explore90navigate-v0_64c_\tl_use:N \l_hps_tl/igpr-(\tl_use:N \l_hps_tl)-normalized_goal_distance_return-4.png}
    \caption{90\%\ Noise}
  \end{subfigure}
  \begin{subfigure}[t]{\l_subfig_width_dim}
    \centering
    \includegraphics[width=\textwidth]{\tl_use:N \l_prefix \tl_use:N \l_agent_tl _antmaze-medium-explore80navigate-v0_64c_\tl_use:N \l_hps_tl/igpr-(\tl_use:N \l_hps_tl)-normalized_goal_distance_return-4.png}
    \caption{80\%\ Noise}
  \end{subfigure}
  \begin{subfigure}[t]{\l_subfig_width_dim}
    \centering
    \includegraphics[width=\textwidth]{\tl_use:N \l_prefix \tl_use:N \l_agent_tl _antmaze-medium-explore40navigate-v0_64c_\tl_use:N \l_hps_tl/igpr-(\tl_use:N \l_hps_tl)-normalized_goal_distance_return-4.png}
    \caption{40\%\ Noise}
  \end{subfigure}
  \begin{subfigure}[t]{\l_subfig_width_dim}
    \centering
    \includegraphics[width=\textwidth]{\tl_use:N \l_prefix \tl_use:N \l_agent_tl _antmaze-medium-explore0navigate-v0_64c_\tl_use:N \l_hps_tl/igpr-(\tl_use:N \l_hps_tl)-normalized_goal_distance_return-4.png}
    \caption{0\%\ Noise}
  \end{subfigure}
}

\ExplSyntaxOff

\usepackage{latexml} %
\iflatexml\makeatletter
\newif\iflx@fig
\def\lx@tw{\textwidth}
\newcommand\lx@mpbegin[2][]{\def\lx@w{#2}%
  \ifx\lx@w\lx@tw\begin{figure}\lx@figtrue\else\par\begingroup\lx@figfalse\fi}
\def\lx@mpend{\iflx@fig\end{figure}\else\endgroup\par\fi}
\expandafter\let\csname begin{minipage}\endcsname\lx@mpbegin
\expandafter\let\csname end{minipage}\endcsname\lx@mpend
\makeatother
\renewcommand\subcaptionbox[2]{%
  \begin{subfigure}{0.48\textwidth}\centering#2\caption{#1}\end{subfigure}}
\fi
\usepackage{tcolorbox}
\iflatexml
\newenvironment{takeawaybox}{\begin{quote}}{\end{quote}}
\else
\newtcolorbox{takeawaybox}{
    colback=gray!10,
    colframe=black,
    boxrule=0.9pt,
    arc=3pt,
    left=4pt, right=4pt, top=2pt, bottom=2pt, boxsep=2pt,
    before skip=6pt,
    after skip=6pt
}
\fi

\usepackage{multirow}

\usepackage{expl3}

\usepackage{xparse}
\usepackage{siunitx}
\usepackage{minitoc}

\newcommand{\AlgGCIQL}{\textnormal{\textsc{GCIQL}}}
\newcommand{\AlgGCIVL}{\textnormal{\textsc{GCIVL}}}
\newcommand{\AlgQRL}{\textnormal{\textsc{QRL}}}
\newcommand{\AlgCRL}{\textnormal{\textsc{CRL}}}

\newcommand{\EnvAntMazeM}{\texttt{AntMaze-M}}
\newcommand{\EnvAntMazeL}{\texttt{AntMaze-L}}
\newcommand{\EnvCube}{\texttt{Cube}}
\newcommand{\EnvScene}{\texttt{Scene}}

\newcommand{\AWR}{\textnormal{\textsc{AWR}}}
\newcommand{\GCRL}{\textnormal{\textsc{GCRL}}}
\newcommand{\OGBench}{\textnormal{\textsc{OGBench}}}

\newcommand{\Rdist}{R_{\mathrm{dist}}}
\newcommand{\rhoeps}{\rho_{\varepsilon}}
\newcommand{\rhonine}{\rho_{0.9}}
\newcommand{\rhoeight}{\rho_{0.8}}
\newcommand{\FRAUC}{\textnormal{\textsc{FR-AUC}}}
\newcommand{\MRR}{\textnormal{\textsc{MRR}}}
\newcommand{\ESS}{\textnormal{\textsc{ESS}}}

\usepackage{lmodern}
\usepackage{tabularx}
\usepackage{enumitem}

\usepackage{wrapfig}
\usepackage[colorinlistoftodos,textsize=footnotesize]{todonotes}

\hypersetup{
  pdftitle={Beyond Success Rates: Trainability and Extractability for Offline GCRL},
  pdfauthor={Jan Malte T\"opperwien, Aditya Mohan, Marius Lindauer}
}

\newcommand{\FAD}{future-association discrimination}
\newcommand{\FADcap}{Future-association discrimination}
\newcommand{\TopFivePctMass}{top-5\% mass}

\title{Beyond Success Rates: Trainability and Extractability for Offline GCRL}

\author{%
  Jan Malte Töpperwien$^{1}$\thanks{Equal contribution. Correspondence to \texttt{m.toepperwien@ai.uni-hannover.de} and \texttt{a.mohan@ai.uni-hannover.de}.}\\
  \texttt{m.toepperwien@ai.uni-hannover.de}
  \And
  Aditya Mohan$^{1}$\footnotemark[1]\\
  \texttt{a.mohan@ai.uni-hannover.de}
  \And
  Marius Lindauer$^{1,2}$ \\
  \texttt{m.lindauer@ai.uni-hannover.de} \\
  $^1$Institute of Artificial Intelligence (LUH$|$AI),\\
  Leibniz University Hannover \\
  $^2$L3S Research Center
}

\begin{document}

\maketitle

\doparttoc %
\faketableofcontents %

\begin{abstract}
Offline goal-conditioned reinforcement learning (GCRL) methods are typically benchmarked by the success rates of optimized hyperparameter configurations.
However, success rate by itself does not provide information about the sensitivity of algorithms to hyperparameters for value learning or policy extraction, since an algorithm could obtain the same success rate across a wide range of hyperparameters, or might succeed only on specific settings.
We study this phenomenon in the offline GCRL algorithms GCIQL, GCIVL, QRL, and CRL under a shared policy extractor, by constructing landscapes of interaction between the learning rate and policy extraction hyperparameters such as the temperature of advantage-weighted regression (AWR).
These landscapes reveal large, task-dependent differences in hyperparameter sensitivity. 
On AntMaze-M, 59\% of QRL configurations and 66\% of CRL configurations achieve at least 90\% of the method's best sampled success rate, compared with 3\% for both GCIQL and GCIVL. 
On Cube and Scene, GCIQL and GCIVL instead attain higher success over broader hyperparameter regions than QRL and CRL. 
To interpret these differences, we analyze how reliably learned advantages rank actual goals from the same trajectory above random goals, and how strongly AWR concentrates policy updates on a small subset of transitions. 
Our findings indicate strongly differing behavior of these algorithms at policy-extraction time, even under the same success rates.
Given these results, we argue that offline GCRL methods should be evaluated by both attainable success and the needed hyperparameter range to sustain it, complemented by post-hoc diagnostics to understand the signal used to train the policy.
The overview and code are available at \url{https://luh-ai.github.io/GCRL-Landscapes/}.
\end{abstract}

\section{Introduction}

\begin{figure*}[!t]
    \centering
    \includegraphics[width=0.98\textwidth]{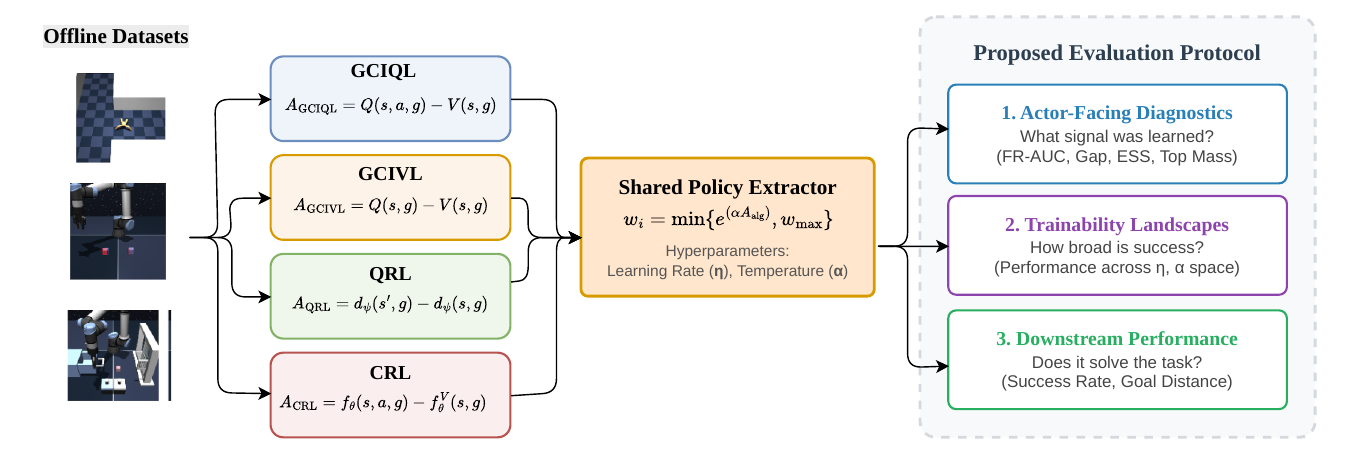}
    \caption{\textbf{Proposed evaluation framework.}
    \AlgGCIQL{}, \AlgGCIVL{}, \AlgQRL{}, and \AlgCRL{} produce different advantage signals, which are passed to a shared \AWR{} extractor.
    We evaluate downstream performance, trainability landscapes over learning rate \((\eta)\) and \AWR{} temperature \((\alpha)\), and policy-extraction diagnostics for \FAD{} and \AWR{} weight concentration.}
    \label{fig:method_overview}
\end{figure*}

Offline goal-conditioned reinforcement learning (GCRL) learns goal-reaching policies from unlabeled trajectories by relabeling dataset states as goals~\citep{schaul-icml15a,andrychowicz-neurips17a,ghosh-iclr21a,ma-neurips22a,park-neurips23a,jackson-neurips25a}. 
Recent benchmarks such as OGBench~\citep{park-iclr25a} evaluate these methods across locomotion and manipulation tasks by reporting the final success rates for the best-tuned hyperparameter configurations~\citep{paine-arxiv20a}. 
However, a success rate obtained after tuning does not reveal how sensitive that performance is to hyperparameter choices. We ask: \emph{how broadly do successful settings extend across the hyperparameters governing value learning and policy extraction?}

This question matters because offline RL performance depends on both the learned values and the policy extracted from them~\citep{levine-arxiv20a,park-neurips24a}.
Advantage-weighted regression (\AWR{}) learns a policy by imitating dataset actions, giving more weight to actions with a higher advantage~\citep{peng-arxiv19a}. 
Hyperparameters affect both how these estimates are learned and how strongly they influence the policy. 
We therefore examine whether different GCRL methods achieve high success across many settings or require careful tuning.

To answer this question, we construct \emph{trainability landscapes} following \citet{mohan-automlconf23a}, mapping hyperparameter configurations to performance at different stages of training. 
We vary the learning rate ($\eta$), which affects both value learning and policy learning, and the \AWR{} temperature ($\alpha$), which controls how strongly the policy favors high-advantage transitions.
The resulting landscapes show whether high performance persists across many sampled settings or requires a narrow range of settings, which we quantify as \emph{trainability} ($\rho_\varepsilon(t)$).
We then characterize \emph{extractability} using two diagnostics on the training signal: \FAD{}, which measures whether learned advantages rank future goals from the same trajectory above random goals, and \AWR{} weight concentration, which measures how much weight is assigned to a small subset of transitions. 

A shared policy extractor, however, does not make the advantage signals of different GCRL methods equivalent.
\AlgGCIQL{}~\citep{kostrikov-iclr22a,park-neurips23a} estimates an action's value relative to a state-value baseline, whereas \AlgGCIVL{}~\citep{kostrikov-iclr22a,park-neurips23a} and \AlgQRL{}~\citep{wang-icml23a} measure changes in value and quasimetric distance, respectively.
\AlgCRL{}~\citep{eysenbach-neurips22a} derives its advantage from contrastive reachability estimates.
These quantities can differ in how they rank goals and how they weight transitions during policy learning.
A method may rank future goals above random goals yet produce an unsuccessful policy; another may achieve high success despite weak future-goal discrimination.
We therefore examine how goal-ranking scores and \AWR{} weights relate to policy success, because ranking goals for a dataset transition does not directly test whether the resulting policy can reach them.

Our results show that both peak success rates and sensitivity to hyperparameters depend on the task and the algorithm.
On AntMaze, \AlgQRL{} and \AlgCRL{} achieve higher success across broader ranges of settings, whereas \AlgGCIQL{} and \AlgGCIVL{} perform better on Cube and Scene.
The diagnostics provide additional information about the advantages and weights used to learn these policies, but strong goal-ranking scores do not guarantee success.
For example, on Scene, \AlgQRL{} ranks future goals above random goals more reliably than \AlgGCIQL{}, yet achieves substantially lower success.
These findings support evaluating methods by both their success and their sensitivity to tuning, with policy-extraction diagnostics helping interpret the results.

\paragraph{Contributions.}
\begin{inparaenum}[(i)]
    \item We introduce an evaluation protocol for offline GCRL that combines policy success, trainability landscapes, and policy-extraction diagnostics under a shared \AWR{} extractor.
    \item We compare \AlgGCIQL{}, \AlgGCIVL{}, \AlgQRL{}, and \AlgCRL{} across \EnvAntMazeM{}, \EnvAntMazeL{}, \EnvCube{}, and \EnvScene{}, showing that both peak success and the fraction of sampled settings achieving near-peak success depend on the task and method.
    \item We characterize how advantage rankings and \AWR{} weight concentration relate to policy performance, identifying cases where stronger \FAD{} accompanies lower success.
\end{inparaenum}

\section{Preliminaries}
\label{sec:preliminaries}

We consider a reward-free controlled Markov process \(\mathcal{M}=(\mathcal{S},\mathcal{A},\mu,p)\), with state space
\(\mathcal{S}\), action space \(\mathcal{A}\), initial-state distribution \(\mu\), and transition kernel \(p(s'\mid s,a)\).
The learner receives an unlabeled offline dataset \(\mathcal{D}=\{\tau^{(n)}\}_{n=1}^{N}\), where \(\tau=(s_0,a_0,s_1,\ldots,a_{T-1},s_T)\), collected by an unknown behavior policy.
The goal is to learn a goal-conditioned policy \(\pi(a\mid s,g)\), where \(g\in\mathcal{G}\) may be a full state or a task-specific state projection, such as an object position.

\paragraph{Self-supervised goal relabeling.}
Offline \GCRL{} methods in \OGBench{} train on goals relabeled from the dataset.
For a transition, a goal \(g\) is sampled from current states, future states in the same trajectory, or random states from the dataset marginal, with algorithm-specific probabilities.
The methods use these goals differently during value learning: \AlgGCIVL{} and \AlgGCIQL{} use mixed current/future/random value goals, \AlgCRL{} uses future trajectory goals with minibatch contrastive negatives, and \AlgQRL{} uses random value goals while treating one-step transitions as local positives.
For actor extraction, all four methods use future goals from the same trajectory.

\paragraph{Policy extraction with \AWR{}.}
We use the same \AWR{} extractor for all evaluated algorithms.
Given an algorithm-specific advantage \(A_{\mathrm{alg}}(s,a,g)\), \AWR{} trains the policy by weighted behavioral cloning:
\begin{gather*}
\mathcal{L}_{\mathrm{AWR}}(\pi) = -\mathbb{E}_{(s,a,g)\sim\mathcal{D}_{\mathrm{rel}}} \left[ w_\alpha(s,a,g)\log\pi(a\mid s,g) \right], \\
w_\alpha(s,a,g) = \min\{e^{(\alpha A_{\mathrm{alg}}(s,a,g))},w_{\max}\}.
\end{gather*}
The relabeled data distribution is denoted by \(\mathcal{D}_{\mathrm{rel}}\).
The temperature \(\alpha\) controls the concentration of imitation weights on high-advantage transitions.
Under this shared objective, the imitation weights depend on each method's advantage estimates and the \AWR{} temperature \(\alpha\).

\paragraph{Advantage signals.}
The four methods compute advantages differently, as summarized in \cref{tab:advantages}.
These signals encode different notions of goal-directed progress: bootstrapped action-value margins, temporal value progress, quasimetric distance reduction, and contrastive reachability.
Therefore, no single geometric metric is a universal target for all methods.

\begin{table}[t]
\centering
\small
\setlength{\tabcolsep}{3pt}
\begin{tabular}{lll}
\toprule
\textbf{Algorithm} & \textbf{Advantage signal} $A_{\mathrm{alg}}(s,a,g)$ & \textbf{Informal meaning} \\
\midrule
GCIQL & $Q(s,a,g)-V(s,g)$                   & Action value above the state-goal baseline\\
GCIVL & $V(s',g)-V(s,g)$                    & One-step value progress \\
QRL   & $d_\phi(s,g)-d_\phi(s',g)$          & One-step distance reduction\\
CRL   & $f_\theta(s,a,g)-f_\theta^V(s,g)$   & Reachability score relative to a state-goal baseline \\
\bottomrule
\end{tabular}
\vspace{0.5em}
\caption{\textbf{Advantages.} All methods use the same \AWR{} extractor in our controlled evaluation, but the extractor receives semantically different advantage signals.}
\label{tab:advantages}
\end{table}

\paragraph{Trainability landscapes.}
We study trainability through performance surfaces at different training stages \(f_t:\Lambda\to\mathbb{R}\), where each configuration \(\lambda\in\Lambda\) maps to downstream performance at phase \(t\).
The shared search space is
\[
    \Lambda=\Lambda_\eta\times\Lambda_\alpha,
\]
where \(\eta\) is the optimizer learning rate and \(\alpha\) is the \AWR{} temperature.
With the offline dataset fixed, these landscapes show how performance varies with the learning rate and \AWR{} temperature.
Broad regions of high performance indicate success across many sampled settings; narrow peaks indicate success over a small range of settings.\footnote{Fitted surfaces are used only for visualization; all landscape masses and diagnostics are computed from raw configuration evaluations.}

\section{Analysis Protocol and Experimental Setup}
\label{sec:method}

Our protocol treats downstream performance as the outcome of three components:
the learned advantage signal, the \AWR{} weighting rule used for actor extraction, and the region of hyperparameter space where these components produce successful behavior.
We measure these components through downstream evaluation, trainability landscapes, and extractor-facing diagnostics.

\paragraph{Datasets and algorithms.}
We evaluate \AlgGCIQL, \AlgGCIVL, \AlgQRL, and \AlgCRL{} on static \OGBench{} datasets (MIT license).
We restrict the controlled comparison to methods expressible as flat \AWR{} over per-transition advantages; 
HIQL~\citep{park-neurips23a} is excluded because it uses hierarchical extraction from a value function.
The navigation tasks are \EnvAntMazeM{} and \EnvAntMazeL{} (\texttt{antmaze-\{medium,large\}-navigate-v0}).
The manipulation tasks are \EnvCube{} and \EnvScene{} (\texttt{cube-single-play-v0}, \texttt{scene-play-v0}).
\Cref{app:generalization} relaxes part of this setup: on \EnvAntMazeM{} and \EnvCube{} it replaces \AWR{} with alternative extractors and compares against FQL, a separate non-\AWR{} algorithm, and on \EnvAntMazeM{} it adds the \texttt{stitch} dataset.

\paragraph{Search space and phase construction.}
For each method and environment, we evaluate a two-dimensional search space
\(C\subset\Lambda_\eta\times\Lambda_\alpha\).
The set \(C\) contains 64 configurations sampled with a scrambled Sobol sequence~\citep{sobol-cmmp67a,joe-sjsci08a}.
We sample the learning rate \(\eta\) on a log scale and the \AWR{} temperature \(\alpha\) on a linear scale.
Each configuration is run across five random seeds.

We divide training into four phases and evaluate every configuration at the end of each phase.
To compare hyperparameter choices from a common starting point, all configurations in the next phase start from the same reference checkpoint.
We select the configuration with the highest interquartile mean (IQM) success over seeds in the previous phase and use the checkpoint of its seed whose success is closest to that IQM.
Thus, all configurations within a phase start from the same checkpoint.
After restoring it, the five seeds of a configuration differ only in minibatch sampling and goal relabeling; in the first phase, they additionally differ in network initialization.
Later-phase landscapes therefore measure how hyperparameter choices affect continued training from this selected checkpoint.
Phase boundaries are set per algorithm and dataset as fixed fractions of the training time a reference run needs to reach 95\% of its maximum success.
The reference run uses the OGBench default hyperparameters except for the policy extractor, which is \AWR{} for all methods (\Cref{sec:appendix:phases}).

\paragraph{Downstream performance and landscape breadth.}
We measure downstream performance using success rate and normalized goal-distance return \(\Rdist\).
Success rate is used for checkpoint selection, final performance, and landscape breadth, and \(\Rdist\) is used for smoother landscape visualization where appropriate.
We fit independent Gaussian process regressors only to draw smooth response surfaces.\footnote{``Independent'' means that each plotted surface is fit separately as a smooth regressor. We do not use GP posterior uncertainty for landscape breadth, bootstrap intervals, or diagnostics.}
All quantitative landscape statistics are computed from raw evaluated configurations.

For phase \(t\), let \(\bar S_t(\lambda)\) denote the IQM success of configuration \(\lambda\) over its five seeds, i.e., the mean of the middle three seeds.
We define the \(\varepsilon\)-optimality mass as
\[
    \rho_{\varepsilon}(t) = \frac{1}{|C|} \sum_{\lambda\in C} \mathbf{1} \left[ \bar S_t(\lambda) \ge \varepsilon \max_{\lambda'\in C}\bar S_t(\lambda') \right].
\]
We report \(\rhonine\) for the near-peak region and \(\rhoeight\) for the broader moderate-performance basin.
Because \(\rho_{\varepsilon}\) is relative to each method's best sampled configuration, a broad basin can still have low absolute success.
We therefore report \(\rho_{\varepsilon}\) together with mean and maximum success. In the appendix robustness checks, where an absolute performance threshold \(c\) is meaningful, we additionally report the absolute-optimality mass
\[
    \rho_{\mathrm{abs}}(c) = \frac{1}{|C|} \sum_{\lambda\in C} \mathbf{1} \left[ \bar S_t(\lambda) \ge c \right],
\]
the fraction of sampled configurations reaching success at least \(c\) in absolute terms, independent of the best sampled configuration.
High \(\rhoeps\) together with high maximum success indicates accessible success; low \(\rhoeps\) with high maximum success indicates a narrow but attainable peak.

\paragraph{Policy-extraction diagnostics.}
We diagnose two parts of policy extraction: the learned advantage signal and the \AWR{} weights it induces.
\FADcap{} and cross-goal ranking measure whether the learned advantage assigns higher scores to corresponding future goals than to random goals.
\AWR{} concentration measures whether actor updates draw weight from many transitions or collapse onto a small high-advantage tail.

\paragraph{\FADcap{}.}
For a batch of \(B\) dataset transitions \((s_i,a_i,s_i')\), let \(A_{\mathrm{alg}}(s_i,a_i,g)\) denote the algorithm's advantage estimate for transition \(i\) and goal \(g\).
For each transition, we sample a future goal \(g_i^+\) from the same trajectory and \(K\) random goals \(g_{ij}^-\).
We compute the advantage difference:
\[
\Delta A_{ij}^{\mathrm{FR}} = A_{\mathrm{alg}}(s_i,a_i,g_i^+) - A_{\mathrm{alg}}(s_i,a_i,g_{ij}^-)
\]
A positive difference means that the future goal receives a higher advantage than the random goal for the same transition.
We report 
\[
\FRAUC = \Pr\bigl(\Delta A^{\mathrm{FR}} > 0\bigr)
\quad\text{and}\quad
\mathrm{Gap} = \mathbb{E}\bigl[\Delta A^{\mathrm{FR}}\bigr].
\]
For unclipped \AWR{} weights, \(\alpha\Delta A_{ij}^{\mathrm{FR}}\) equals their log-weight ratio.
The gap depends on the advantage scale and compares goals within a transition. It does not, however, measure weight concentration across transitions.

\paragraph{Cross-goal ranking.}
For each transition, we rank the future goal and its \(K\) sampled random goals by \(A_{\mathrm{alg}}(s_i,a_i,g)\).
Let \(r_i\) be the rank of \(g_i^+\), where \(r_i=1\) denotes the highest advantage.
We report mean reciprocal rank:
\[
\MRR=B^{-1}\sum_{i=1}^{B}1/r_i.
\]
We use \(K=255\), so each candidate set contains 256 goals and random ranking gives \(\MRR\approx H_{256}/256\approx0.024\).

\paragraph{AWR weight concentration.}
Using the future-goal advantage for each transition, we compute
\[
    w_i=\min\{\exp(\alpha A_{\mathrm{alg}}(s_i,a_i,g_i^+)),w_{\max}\}.
\]
We measure the normalized effective sample size,
\[
    \ESS = \frac{\left(\sum_i w_i\right)^2} {B\sum_i w_i^2}.
\]
High \(\ESS\) indicates that \AWR{} imitates many transitions and low \(\ESS\) indicates that the actor update is concentrated on a small subset of high-weight transitions.
We also measure the fraction of total weight assigned to the highest-weighted 5\% of transitions (\TopFivePctMass{}),
\[
    \mathrm{Top5} = \frac{\sum_{i\in \mathcal{T}_{0.05}} w_i} {\sum_i w_i},
\]
where \(\mathcal{T}_{0.05}\) contains the largest 5\% of weights.
Concentration is useful only when the highest-weighted samples provide reliable supervision for the downstream task.

\paragraph{Diagnostic measurement.}
\Cref{tab:compact_diagnostics} reports final-phase diagnostics for each method and environment, averaged over all sampled configurations, seeds, and validation batches.
Configuration-level diagnostics, including success--diagnostic scatter plots, are reported in \cref{app:diagnostic_scatter}.
These diagnostics are computed post hoc from trained checkpoints and do not require constructing new trainability landscapes.

\paragraph{Implementation scope.}
The shared \AWR{} extractor puts all methods under the same policy-extraction rule.
This lets us compare learned signals under a common extraction interface.
All non-swept implementation choices, including target update rates, batch sizes, architecture sizes, and representation dimensions, follow standard \OGBench{} defaults within each method.
Full hyperparameters and reproducibility details are provided in \cref{sec:appendix:hps}.

\begin{table}[t]
\centering
\small
\setlength{\tabcolsep}{4pt}
\begin{tabularx}{\textwidth}{@{}lXl@{}}
\toprule
\textbf{Evaluation component} & \textbf{What it measures} & \textbf{Where used} \\
\midrule
Downstream performance & Mean and maximum success & \Cref{tab:final_landscape_ci} \\
Landscape breadth & Relative near-optimal mass \(\rhonine,\rhoeight\) & \Cref{tab:final_landscape_ci} \\
\FADcap{} & Pairwise future-vs-random goal comparison \((\FRAUC, \mathrm{Gap})\) & \Cref{tab:compact_diagnostics} \\
Cross-goal ranking & Future-goal rank among 256 candidates \((\MRR)\) & \Cref{tab:compact_diagnostics} \\
\AWR{} concentration & Policy-update shape \((\ESS,\) \TopFivePctMass{}\()\) & \Cref{tab:compact_diagnostics,tab:antmaze_ess_phase4} \\
Robustness checks & Seed/configuration stability and extractor sensitivity & \Cref{app:robustness} \\
\bottomrule
\end{tabularx}
\vspace{0.5em}
\caption{\textbf{Evaluation components.}
The protocol reports downstream performance, landscape breadth, and policy-extraction diagnostics.
Diagnostics describe the learned signal and extractor behavior.
Downstream evaluation measures task success.
}
\label{tab:evaluation_components}
\end{table}

\section{Experiments}
\label{sec:experiments}

We evaluate \AlgGCIQL, \AlgGCIVL, \AlgQRL, and \AlgCRL{} under the shared \AWR{} extractor from \cref{sec:method}.
The experiments address three questions:
\begin{inparaenum}[(i)]
    \item Does peak success hide trainability differences?
    \item How do advantage rankings and AWR weight concentration relate to performance?
    \item When do goal-discrimination diagnostics diverge from downstream control?
\end{inparaenum}
Experiments used roughly \(10{,}000\) GPU-hours on A100 and H100 GPUs, with peak memory below approximately \(100\)GB.
We report additional checks for seed variance, phase mobility, configuration subsampling, \(w_{\max}\) sensitivity, advantage normalization, and sampled-space sensitivity in \Cref{app:robustness}.

\subsection{Landscape breadth separates accessible and brittle success}
\label{sec:experiments:landscapes}

\begin{figure}[t]
  \centering
  \begin{minipage}[c]{0.66\textwidth}
    \centering
    \begin{tabular}{@{}cc@{}}
      \includegraphics[width=0.48\linewidth]{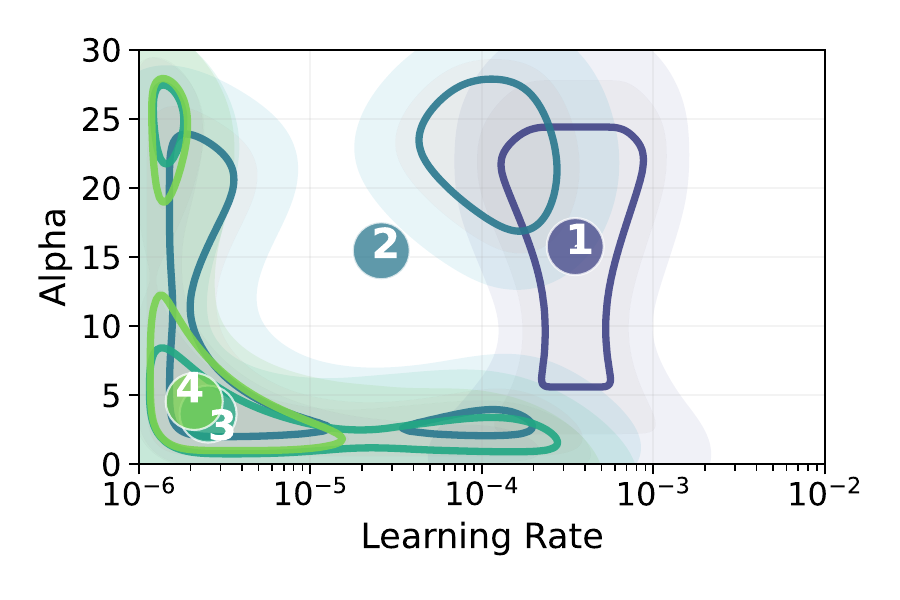}
      &
      \includegraphics[width=0.48\linewidth]{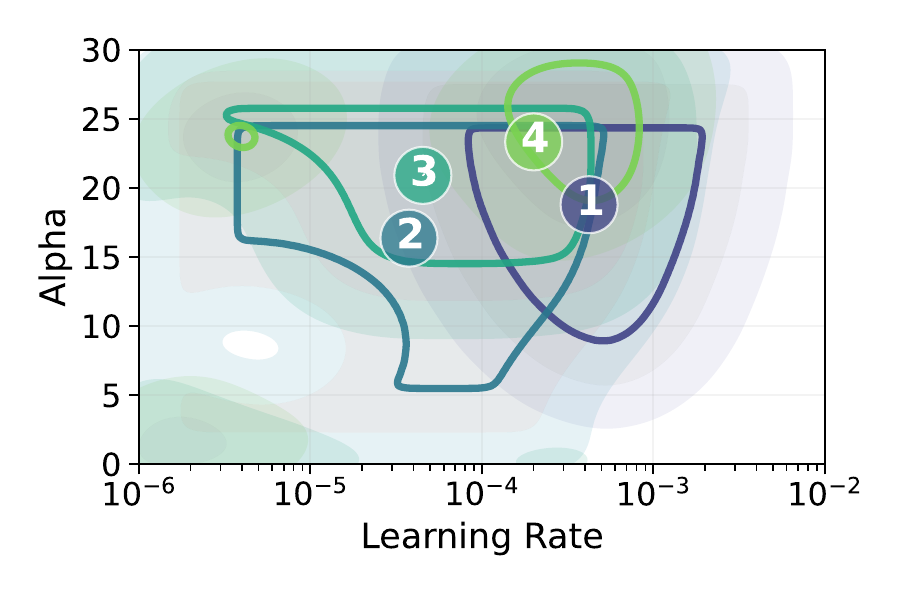}
      \\
      \footnotesize (a) \AlgGCIQL{} & \footnotesize (b) \AlgGCIVL{} \\ [4pt]
      \includegraphics[width=0.48\linewidth]{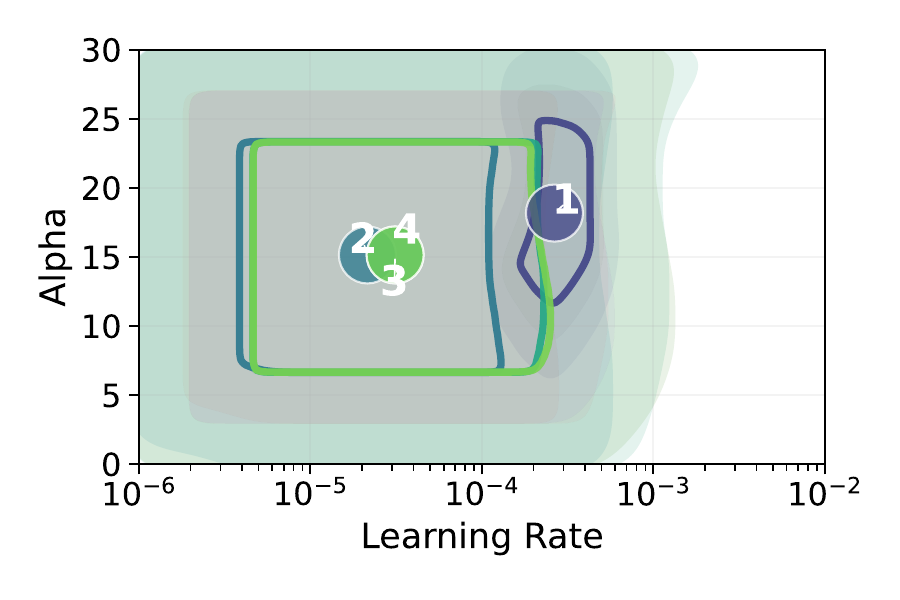}
      &
      \includegraphics[width=0.48\linewidth]{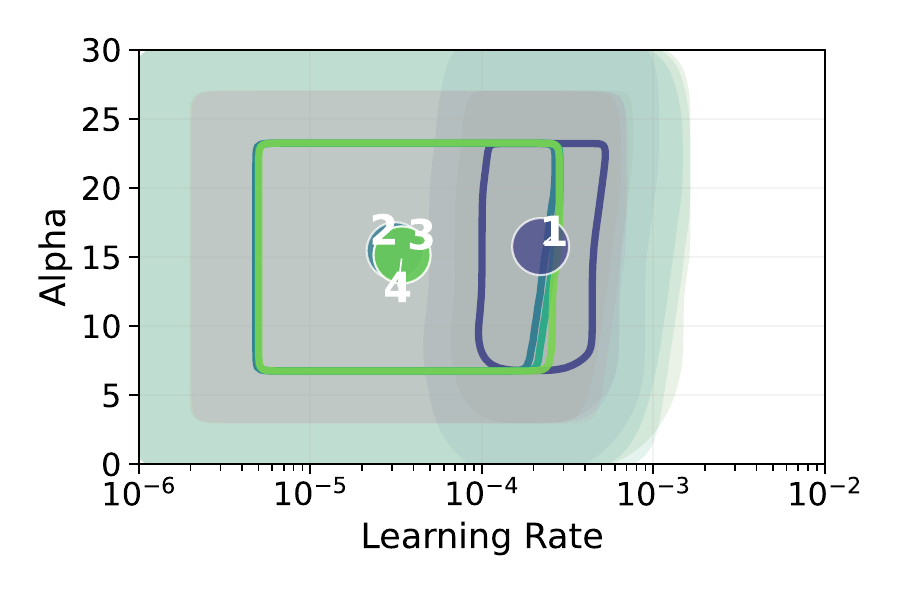}
      \\
      \footnotesize (c) \AlgCRL{} & \footnotesize (d) \AlgQRL{}
    \end{tabular}
  \end{minipage}\hfill
  \begin{minipage}[c]{0.32\textwidth}
    \captionsetup{
      type=figure,
      justification=raggedright,
      singlelinecheck=false,
      font=small
    }
    \caption{\textbf{Trainability landscapes on \EnvAntMazeM.}
        We show top-20\% regions in the learning-rate \((\eta)\) -- \AWR{}-temperature \((\alpha)\) plane.
        Numbers indicate training progress. \\
        \AlgGCIQL{} produces narrower and less stationary landscapes than \AlgGCIVL{}, \AlgCRL{}, and \AlgQRL{} on \EnvAntMazeM{}.
    }
    \label{fig:mobility:phases_vs_quality-antmaze}
  \end{minipage}
\end{figure}

\begin{table}[t]
\centering
\small
\setlength{\tabcolsep}{10pt}
\begin{tabular}{llrrrr}
\toprule
Environment     & Algorithm & Max succ. & Mean succ. & $\rhonine$ & $\rhoeight$ \\
\midrule
\EnvAntMazeM{}  & \AlgCRL{}   & 0.83 & 0.67 & 0.66 & 0.80 \\
                & \AlgGCIQL{} & 0.70 & 0.34 & 0.03 & 0.05 \\
                & \AlgGCIVL{} & 0.74 & 0.49 & 0.03 & 0.28 \\
                & \AlgQRL{}   & 0.84 & 0.66 & 0.59 & 0.70 \\
\midrule
\EnvAntMazeL{}  & \AlgCRL{}   & 0.54 & 0.36 & 0.17 & 0.59 \\
                & \AlgGCIQL{} & 0.24 & 0.16 & 0.14 & 0.39 \\
                & \AlgGCIVL{} & 0.17 & 0.09 & 0.02 & 0.11 \\
                & \AlgQRL{}   & 0.55 & 0.38 & 0.23 & 0.59 \\
\midrule
\EnvCube{}      & \AlgCRL{}   & 0.28 & 0.12 & 0.08 & 0.17 \\
                & \AlgGCIQL{} & 0.77 & 0.53 & 0.28 & 0.56 \\
                & \AlgGCIVL{} & 0.91 & 0.70 & 0.56 & 0.67 \\
                & \AlgQRL{}   & 0.18 & 0.07 & 0.02 & 0.06 \\
\midrule
\EnvScene{}     & \AlgCRL{}   & 0.19 & 0.10 & 0.03 & 0.25 \\
                & \AlgGCIQL{} & 0.87 & 0.64 & 0.47 & 0.66 \\
                & \AlgGCIVL{} & 0.73 & 0.55 & 0.41 & 0.64 \\
                & \AlgQRL{}   & 0.22 & 0.13 & 0.08 & 0.19 \\
\bottomrule
\end{tabular}%
\vspace{0.5em}
\caption{\textbf{Final-phase success and landscape breadth.}
\(\rhonine\) and \(\rhoeight\) are relative to each method's best sampled configuration; mean and maximum success give the absolute scale.
}
\label{tab:final_landscape_ci}
\end{table}

\Cref{tab:final_landscape_ci} separates peak performance from trainability.
On \EnvAntMazeM, \AlgCRL{} and \AlgQRL{} reach the highest maximum success and also occupy broad near-optimal regions, with \(\rhonine=0.66\) and \(0.59\), respectively.
\AlgGCIQL{} and \AlgGCIVL{} reach lower maximum success, but the more important distinction is breadth:
both have small \(\rhonine\), while \AlgGCIVL{} retains a larger moderate-performance basin at \(\rhoeight=0.28\).

\Cref{fig:mobility:phases_vs_quality-antmaze} shows the same pattern geometrically.
High-performing regions do not occupy the \((\eta,\alpha)\) plane in the same way across methods.
\AlgCRL{} and \AlgQRL{} have broad top-20\% regions over the sampled space.
\AlgGCIVL{} has a smaller region, consistent with its small near-peak mass (\(\rhonine=0.03\)).
\AlgGCIQL{} has a sharper and more mobile region, showing sensitivity to the interaction between learning rate and \AWR{} temperature.
Peak success alone cannot distinguish these cases.

The manipulation tasks (see \Cref{sec:appendix:landscapes} for mobility plots) show why relative breadth must be read together with absolute performance.
On \EnvScene, \AlgCRL{} has a nonzero moderate-performance basin, \(\rhoeight=0.25\), but its maximum success is only \(0.19\).
This is not a broadly useful region; it is a relative basin around a low ceiling.
For this reason, we report \(\rho_{\varepsilon}\), mean success, and maximum success together.

We further test whether landscape breadth carries over to a different policy extractor, DDPG+BC~\citep{fujimoto-neurips21a}: on \EnvAntMazeM{}, \AlgGCIQL{} and \AlgCRL{} retain their breadth classes (\Cref{app:generalization}).
\AlgGCIVL{} is incompatible with DDPG+BC, and \AlgQRL{} requires an additional latent dynamics model, which changes the algorithm.
Under this modified variant, \AlgQRL{} loses its broad near-peak region (\(\rho_{0.9}\colon 0.59\to0.02\)).
For methods whose critic carries over unchanged, breadth therefore reflects the learned signal, not only the choice of extractor.

\begin{takeawaybox}
\textbf{Takeaway (Landscapes).}
Peak success hides trainability: similar maxima can come from a broad region of good configurations or from a narrow, sensitive peak.
\end{takeawaybox}

\subsection{Diagnostics describe the signal used by \AWR{}}
\label{sec:experiments:diagnostics}

\begin{table}[t]
\centering
\small
\setlength{\tabcolsep}{4pt}
\resizebox{\textwidth}{!}{%
\begin{tabular}{llrrrrrr}
\toprule
 & & \multicolumn{3}{c}{\FADcap{}} & \multicolumn{2}{c}{\AWR{} weights} & Policy \\
\cmidrule(lr){3-5}\cmidrule(lr){6-7}
Environment & Algorithm & \FRAUC{} & Gap & \MRR{} & \ESS{} & \TopFivePctMass{} & Max Success \\
\midrule
\EnvAntMazeM{} & \AlgCRL{}   & 0.530 &  0.169 & 0.022 & 0.329 & 0.252 & 0.83 \\
\EnvAntMazeM{} & \AlgGCIQL{} & 0.471 & -0.102 & 0.045 & 0.140 & 0.494 & 0.70 \\
\EnvAntMazeM{} & \AlgGCIVL{} & 0.564 &  0.143 & 0.078 & 0.468 & 0.150 & 0.74 \\
\EnvAntMazeM{} & \AlgQRL{}   & 0.546 &  0.358 & 0.072 & 0.543 & 0.114 & 0.84 \\
\midrule
\EnvAntMazeL{} & \AlgCRL{}   & 0.527 &  0.174 & 0.019 & 0.327 & 0.223 & 0.54 \\
\EnvAntMazeL{} & \AlgGCIQL{} & 0.472 & -0.101 & 0.042 & 0.144 & 0.517 & 0.24 \\
\EnvAntMazeL{} & \AlgGCIVL{} & 0.575 &  0.136 & 0.084 & 0.459 & 0.154 & 0.17 \\
\EnvAntMazeL{} & \AlgQRL{}   & 0.582 &  0.680 & 0.080 & 0.601 & 0.101 & 0.55 \\
\midrule
\EnvCube{}      & \AlgCRL{}   & 0.562 &  0.875 & 0.022 & 0.429 & 0.194 & 0.28 \\
\EnvCube{}      & \AlgGCIQL{} & 0.531 & -0.039 & 0.040 & 0.159 & 0.481 & 0.77 \\
\EnvCube{}      & \AlgGCIVL{} & 0.644 &  0.160 & 0.064 & 0.559 & 0.134 & 0.91 \\
\EnvCube{}      & \AlgQRL{}   & 0.523 &  0.167 & 0.092 & 0.470 & 0.150 & 0.18 \\
\midrule
\EnvScene{}     & \AlgCRL{}   & 0.555 &  0.695 & 0.022 & 0.446 & 0.181 & 0.19 \\
\EnvScene{}     & \AlgGCIQL{} & 0.411 & -0.192 & 0.041 & 0.134 & 0.562 & 0.87 \\
\EnvScene{}     & \AlgGCIVL{} & 0.672 &  0.185 & 0.134 & 0.524 & 0.143 & 0.73 \\
\EnvScene{}     & \AlgQRL{}   & 0.580 &  0.379 & 0.115 & 0.469 & 0.152 & 0.22 \\
\bottomrule
\end{tabular}%
}
\vspace{0.5em}
\caption{\textbf{Compact diagnostic suite.}
\FRAUC{} and gap measure pairwise \FAD{}.
MRR measures the mean reciprocal rank of the on-trajectory goal among 256 candidates; random MRR is approximately \(0.024\).
\TopFivePctMass{} and \ESS{} describe the \AWR{} weight distribution.
Max success gives downstream context.
}
\label{tab:compact_diagnostics}
\end{table}

\Cref{tab:compact_diagnostics} reports extractor-facing diagnostics at the end of the final phase, averaged over all sampled configurations and seeds.
\FRAUC{} and gap measure \FAD{}, \MRR{} measures future-goal ranking, and \ESS{} and \TopFivePctMass{} measure \AWR{} concentration.
The following subsections show when these diagnostics explain downstream behavior and when they separate from it.

\subsection{\EnvAntMazeM: weight concentration and performance}
\label{sec:experiments:antmaze}

\begin{figure}[t]
  \centering
  \begin{minipage}[t]{0.66\textwidth}
    \vspace{0pt}
    \centering
    \captionsetup[sub]{skip=1pt}
    \subcaptionbox{\AlgCRL{}\label{fig:qvd:antmaze-medium:crl}}%
      {\includegraphics[width=0.49\linewidth]{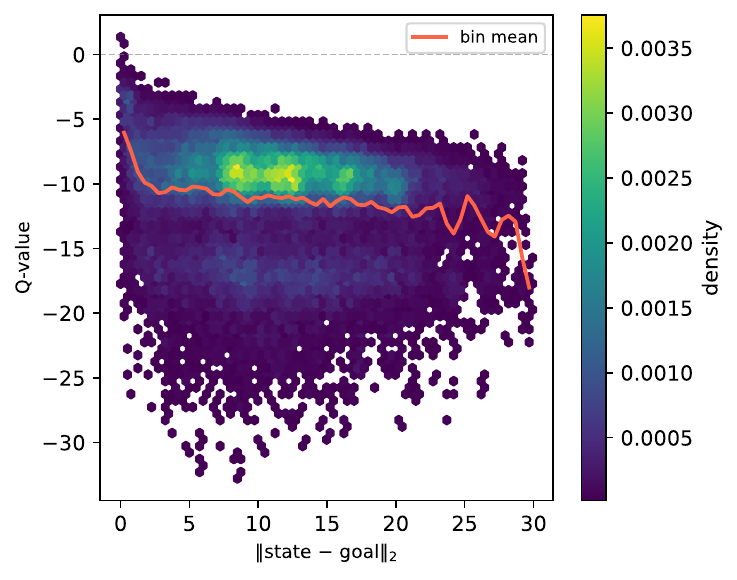}}\hspace{2pt} %
    \subcaptionbox{\AlgGCIQL{}\label{fig:qvd:antmaze-medium:gciql}}%
      {\includegraphics[width=0.49\linewidth]{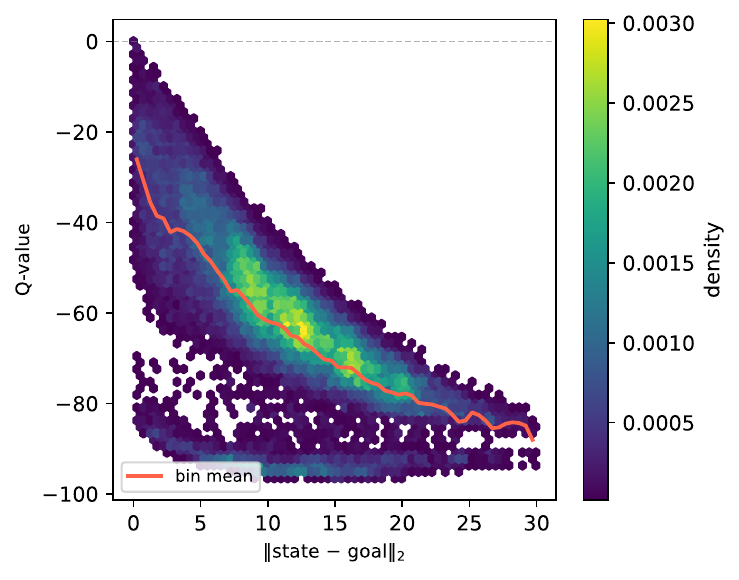}}

    \vspace{6pt} %

    \subcaptionbox{\AlgGCIVL{}\label{fig:qvd:antmaze-medium:gcivl}}%
      {\includegraphics[width=0.49\linewidth]{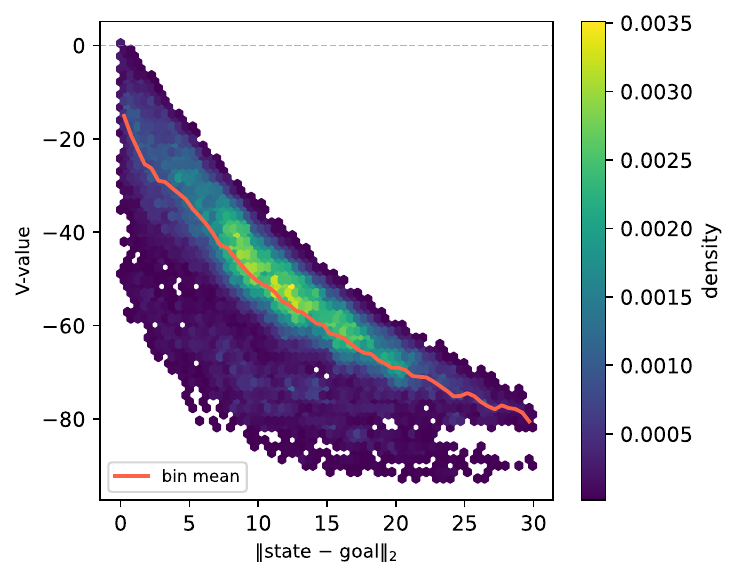}}\hspace{2pt}
    \subcaptionbox{\AlgQRL{}\label{fig:qvd:antmaze-medium:qrl}}%
      {\includegraphics[width=0.49\linewidth]{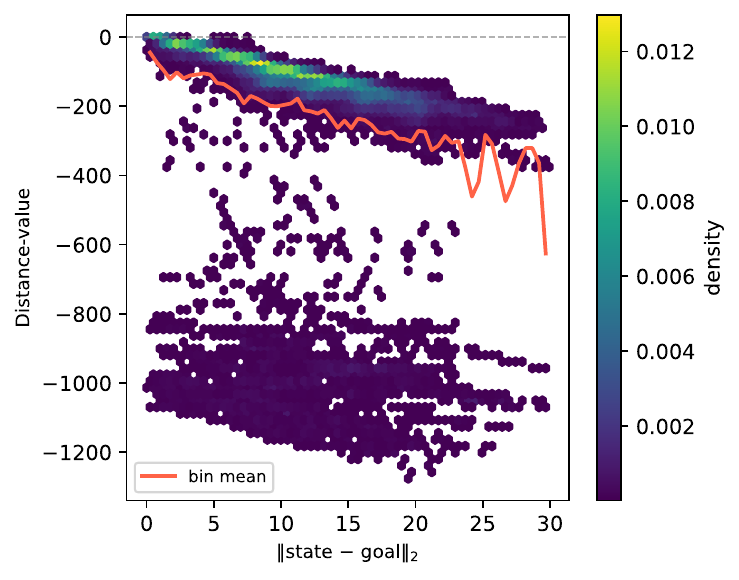}}
  \end{minipage}
  \hspace{8pt} %
  \begin{minipage}[t]{0.30\textwidth}
    \vspace{0pt}
    \caption{\textbf{Score geometry across goal distance on \EnvAntMazeM.}
    Hexbin density from the $256\times256$ cross-product of states and goals from a validation batch, indexed by agent $xy$-distance to the goal.
    A distance-aligned score produces a negative trend.
    \AlgCRL{} is aligned with Euclidean distance only for nearby goals because its score reflects empirical reachability.
    This diagnostic is coarse: Euclidean $xy$-distance ignores maze walls and robot state.}
    \label{fig:qvd:antmaze-medium}
  \end{minipage}
\end{figure}

\begin{table}[t]
\centering
\small
\begin{minipage}[t]{0.5\textwidth}
\vspace{0pt}
\centering
\small
\begin{tabular}{lccc}
\toprule
& \ESS{} & \multicolumn{2}{c}{Corr.\ with return} \\ \cmidrule(lr){3-4}
Algorithm & Mean$\pm$Std & \(r\) & \(p\) \\
\midrule
\AlgCRL{}    & \(0.32\pm0.26\) & \(-0.86\) & \(<0.01\) \\
\AlgGCIQL{}  & \(0.14\pm0.12\) & \( 0.10\) & \(0.07\) \\
\AlgGCIVL{}  & \(0.49\pm0.14\) & \( 0.02\) & \(0.74\) \\
\AlgQRL{}    & \(0.57\pm0.11\) & \( 0.05\) & \(0.38\) \\
\bottomrule
\end{tabular}
\vspace{0.5em}
\caption{\textbf{Final-phase \AWR{} concentration on \EnvAntMazeM.}
Lower \(\ESS\) strongly correlates with higher return for \AlgCRL{}. 
The observed association is weak for \AlgGCIQL{}.
The full phase-resolved table is reported in \cref{app:antmaze_ess_full}.
}
\label{tab:antmaze_ess_phase4}
\end{minipage}\hfill
\begin{minipage}[t]{0.45\textwidth}
\vspace{0pt}
\small
\centering
\begin{tabular}{lrr}
\toprule
Algorithm & Seed \(r\) & Distance-proxy \(r\) \\
\midrule
\AlgCRL{}    & 0.760 & 0.091 \\
\AlgGCIQL{}  & 0.754 & 0.071 \\
\AlgGCIVL{}  & 0.902 & 0.122 \\
\AlgQRL{}    & 0.954 & 0.420 \\
\bottomrule
\end{tabular}
\vspace{2em}
\caption{\textbf{Advantage-rank stability and alignment on \EnvAntMazeM.}
Seed Spearman \(r\) measures stability across seeds.
Distance-proxy Spearman \(r\) measures alignment with a one-step Euclidean-distance reduction.
}
\label{tab:adv_corr}
\end{minipage}
\end{table}

\EnvAntMazeM{} is the clearest case because future-goal relabeling, reachability, and physical progress are closely related.
In \cref{tab:compact_diagnostics}, \AlgQRL{} and \AlgGCIVL{} both produce positive future-association signals with relatively diffuse \AWR{} weights.
\AlgQRL{} has gap \(0.358\) and \(\ESS=0.543\).
\AlgGCIVL{} has \(\FRAUC=0.564\) and \(\ESS=0.468\).
Both methods distribute \AWR{} weight across many transitions.

\Cref{fig:qvd:antmaze-medium} checks whether learned state--goal scores vary with geometric distance, and shows that the methods learn qualitatively different scores.
Ideally, a progress-like score should decrease as \(xy\)-distance to the goal increases.
However, Euclidean distance ignores maze topology, since two nearby points may be separated by a wall, and it also ignores orientation, joint configuration, and velocity.
Despite these limitations, the plot shows that \AlgGCIQL{} and \AlgGCIVL{} decrease smoothly and monotonically with distance.
\AlgQRL{} follows the same trend for most pairs, but also assigns very large distances to a separate cluster of pairs at all \(xy\)-distances.
These pairs may differ in robot state or be separated by walls, which the Euclidean proxy does not capture.
\AlgCRL{} decreases only for nearby goals and is almost flat beyond that, as expected for a reachability score.

The score-geometry plot and the advantage-correlation table measure different objects.
\Cref{fig:qvd:antmaze-medium} evaluates state--goal score geometry.
\Cref{tab:adv_corr} evaluates one-step advantage ranking.
A critic can assign lower scores to farther goals while still giving poorly ordered advantages.

\Cref{tab:adv_corr} shows that \AlgQRL{} has the strongest alignment to the distance proxy, \(r=0.420\), and the most stable seed-wise ranking, \(r=0.954\).
\AlgGCIVL{} is also stable across seeds.
\AlgCRL{} and \AlgGCIQL{} have weaker geometric alignment, which is expected from their signal definitions:
\AlgCRL{} estimates empirical reachability, while \AlgGCIQL{} exposes a bootstrapped action-value margin.

In \cref{tab:antmaze_ess_phase4}, lower \(\ESS\) is strongly associated with higher return for \AlgCRL{} in the final phase.
The reachability signal appears to concentrate \AWR{} on useful behavior-supported actions.
\AlgGCIQL{} has even lower \(\ESS\), but the association with return is absent.
Its negative future-association gap, weak alignment with the distance proxy, and high \TopFivePctMass{} indicate concentration on a narrow \(Q-V\) tail that is less aligned with the navigation diagnostics.

\begin{takeawaybox}
\textbf{Takeaway (AntMaze).}
In AntMaze, where reaching future goals means physical progress, the diagnostics explain the landscapes: \AlgGCIVL{} and \AlgQRL{} spread \AWR{} weight broadly, \AlgCRL{} focuses it on reachable goals, and \AlgGCIQL{} puts most weight on a few samples.
\end{takeawaybox}

\subsection{\EnvCube{} and \EnvScene{}: goal ranking and control misalignment}
\label{sec:experiments:manipulation}

\EnvCube{} and \EnvScene{} separate goal ranking from control.
The goal specifies an object or scene target, while the policy acts through robot state and contact.
A transition can prepare a grasp before the object moves closer to the goal.
This may help explain why goal-ranking scores and policy success differ.

\AlgGCIVL{} gives the most consistent manipulation signal.
On \EnvCube, it has \(\FRAUC=0.644\), \(\ESS=0.559\), and maximum success \(0.91\).
On \EnvScene, it has \(\FRAUC=0.672\), \(\MRR=0.134\), \(\ESS=0.524\), and maximum success \(0.733\).
Across both tasks, one-step value progress gives a discriminative signal that remains broadly extractable.

\AlgGCIQL{} shows the limitation of future-association diagnostics.
On \EnvScene, it reaches maximum success \(0.873\) despite \(\FRAUC=0.411\), gap \(-0.192\), and \(\ESS=0.134\).
These diagnostics test only \FAD{}.
The successful actor update instead appears to rely on an action-conditioned \(Q(s,a,g)-V(s,g)\) margin that identifies useful manipulation actions even when future goals are not ranked above random goals.
Its low \(\ESS\) and high \TopFivePctMass{} show that the \AWR{} weights are highly concentrated.

\AlgQRL{} and \AlgCRL{} show the opposite pattern.
On \EnvScene, \AlgQRL{} has strong ranking metrics, including \(\MRR=0.115\), but maximum success only \(0.220\).
\AlgCRL{} has positive pairwise gaps on \EnvCube{} and \EnvScene{}, but near-random \(\MRR\) and low downstream success.
These results show that \FAD{} is not sufficient for contact-rich control.
In line with prior observations~\citep{mohan-rlbrew26a,shen-arxiv26a}, a method can rank goals well and still fail to extract the actions needed to manipulate the object.
Configuration-level scatter plots in \cref{app:diagnostic_scatter} show the same separation across the sampled configurations.
Manipulation performance depends on the task structure and on the advantage signal used by the extractor, not only on \FAD{}.

\begin{takeawaybox}
\textbf{Takeaway (Manipulation).}
In manipulation, ranking goals well is not enough: a critic can rank goals correctly yet miss the precise contact actions, and an action-conditioned critic can produce good policies despite weak goal-ranking scores.
\end{takeawaybox}

\section{Conclusion}
\label{sec:conclusion}
We evaluated offline \GCRL{} using policy success, trainability landscapes over the learning rate and \AWR{} temperature, and policy-extraction diagnostics.
Peak success alone does not reveal how many hyperparameter settings achieve comparable performance, since similar maxima can arise from broad high-performing regions or narrow sensitive peaks.

Across the tested tasks, \AlgQRL{} and \AlgCRL{} achieve higher success across broader ranges of settings on AntMaze, while \AlgGCIQL{} and \AlgGCIVL{} do so on Cube and Scene.
The relationship between diagnostics and performance also varies across tasks and methods: stronger \FAD{} does not necessarily accompany higher policy success.
On \EnvAntMazeM{}, where future goals coincide with physical progress, the diagnostics help explain the landscapes, while on manipulation tasks, goal ranking and contact-rich control separate.

Thus, even under a shared extractor, methods differ in their learned signals and in the landscapes these signals induce.
We recommend reporting downstream success together with landscape breadth, goal-ranking diagnostics, and \AWR{} weight concentration.

\paragraph{Limitations.}
Our findings are limited to four tasks, two swept hyperparameters, and the sampled parameter ranges.
The shared \AWR{} objective controls policy extraction, while learning-rate changes affect both value and policy learning.
Future work should extend this study to cover hierarchical extractors, action chunking, or test-time planning, and build these landscapes for broader task families and extractors.
\Cref{app:generalization} takes a first step on \EnvAntMazeM{} and \EnvCube{}: breadth patterns largely carry over to other extractors, and most configuration rankings stay positively correlated on a harder dataset.

\begin{ack}
Jan Malte Töpperwien acknowledges funding by the German Research Foundation (DFG) under LI 2801/7-1.
Aditya Mohan and Marius Lindauer acknowledge funding by the European Union (ERC, ``ixAutoML'', grant no.\ 101041029). Views and opinions expressed are however those of the author(s) only and do not necessarily reflect those of the European Union or the European Research Council Executive Agency. Neither the European Union nor the granting authority can be held responsible for them.

\begin{center}
  \includegraphics[height=2.5cm]{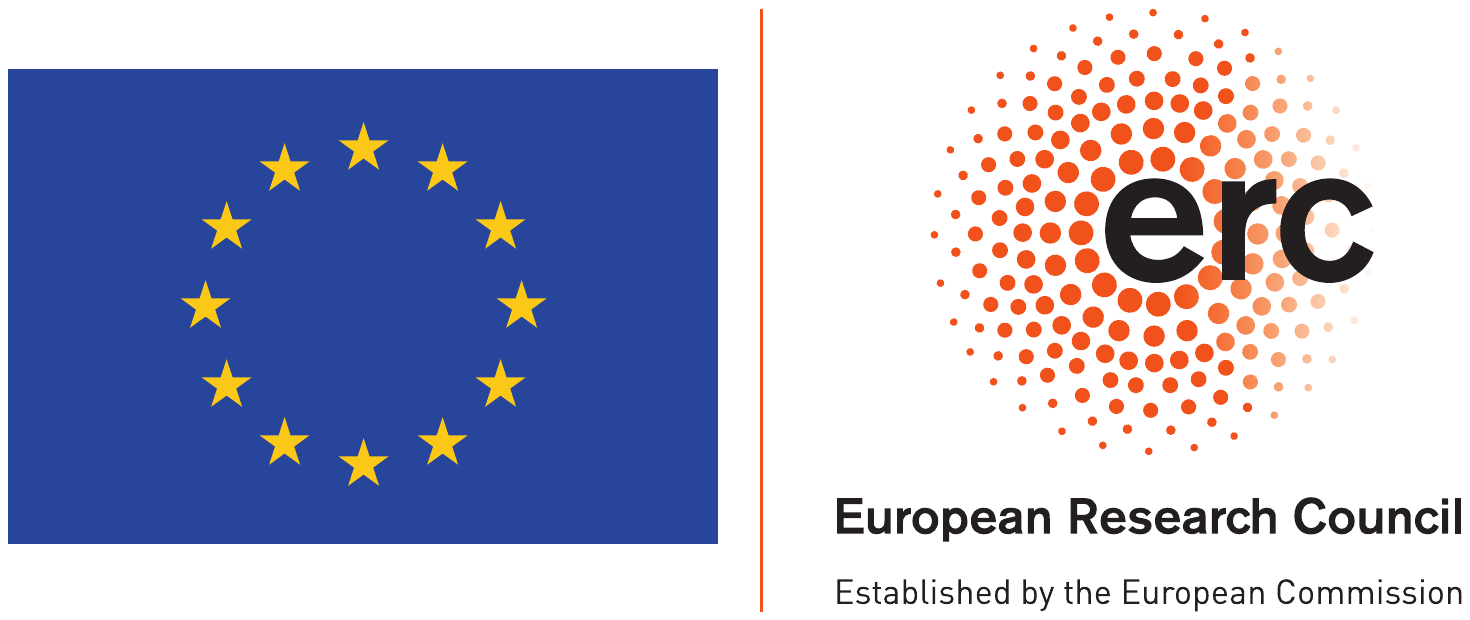}
\end{center}
\end{ack}

\bibliographystyle{plainnat}
\bibliography{bib/strings,bib/lib,bib/local,bib/proc}

\begin{thebibliography}{72}
\providecommand{\natexlab}[1]{#1}
\providecommand{\url}[1]{\texttt{#1}}
\expandafter\ifx\csname urlstyle\endcsname\relax
  \providecommand{\doi}[1]{doi: #1}\else
  \providecommand{\doi}{doi: \begingroup \urlstyle{rm}\Url}\fi

\bibitem[neu(2024)]{neurips24}
\emph{Proceedings of the 38th International Conference on Advances in Neural
  Information Processing Systems ({N}eur{IPS}'24)}, 2024. Curran Associates.

\bibitem[Adkins et~al.(2024)Adkins, Bowling, and White]{adkins-neurips24a}
J.~Adkins, M.~Bowling, and A.~White.
\newblock A method for evaluating hyperparameter sensitivity in reinforcement
  learning.
\newblock In \emph{Proceedings of the 38th International Conference on Advances
  in Neural Information Processing Systems ({N}eur{IPS}'24)} \citet{neurips24}.

\bibitem[Agarwal et~al.(2021)Agarwal, Schwarzer, Castro, Courville, and
  Bellemare]{agarwal-neurips21a}
R.~Agarwal, M.~Schwarzer, P.~Samuel Castro, A.~C. Courville, and M.~G.
  Bellemare.
\newblock Deep reinforcement learning at the edge of the statistical precipice.
\newblock In  \citet{neurips21}.

\bibitem[Ahn et~al.(2025)Ahn, Choi, Han, and Moon]{ahn-neurips25a}
H.~Ahn, H.~Choi, J.~Han, and T.~Moon.
\newblock Option-aware temporally abstracted value for offline goal-conditioned
  reinforcement learning.
\newblock In  \citet{neurips25}.

\bibitem[Andrychowicz et~al.(2017)Andrychowicz, Wolski, Ray, Schneider, Fong,
  Welinder, McGrew, Tobin, Abbeel, and Zaremba]{andrychowicz-neurips17a}
M.~Andrychowicz, F.~Wolski, A.~Ray, J.~Schneider, R.~Fong, P.~Welinder,
  B.~McGrew, J.~Tobin, P.~Abbeel, and W.~Zaremba.
\newblock Hindsight experience replay.
\newblock In \emph{Proceedings of the 32nd International Conference on Advances
  in Neural Information Processing Systems ({N}eur{IPS}'17)}, 2017.

\bibitem[Belgrave et~al.(2025)Belgrave, Zhang, Lin, Pascanu, Koniusz, Ghassemi,
  and Chen]{neurips25}
D.~Belgrave, C.~Zhang, H.~Lin, R.~Pascanu, P.~Koniusz, M.~Ghassemi, and
  N.~Chen, editors.
\newblock \emph{Proceedings of the 39th International Conference on Advances in
  Neural Information Processing Systems ({N}eur{IPS}'25)}, 2025. Curran
  Associates.

\bibitem[Bortkiewicz et~al.(2025)Bortkiewicz, Pa\l{}ucki, Myers, Dziarmaga,
  Arczewski, Kuci\'{n}ski, and Eysenbach]{bortkiewicz-iclr25a}
M.~Bortkiewicz, W.~Pa\l{}ucki, V.~Myers, T.~Dziarmaga, T.~Arczewski,
  L.~Kuci\'{n}ski, and B.~Eysenbach.
\newblock Accelerating goal-conditioned rl algorithms and research.
\newblock In \emph{The Thirteenth International Conference on Learning
  Representations ({ICLR}'25)} \citet{iclr25}.

\bibitem[Chebotar et~al.(2021)Chebotar, Hausman, Lu, Xiao, Kalashnikov, Varley,
  Irpan, Eysenbach, Julian, Finn, and Levine]{chebotar-icml21a}
Y.~Chebotar, K.~Hausman, Y.~Lu, T.~Xiao, D.~Kalashnikov, J.~Varley, A.~Irpan,
  B.~Eysenbach, R.~Julian, C.~Finn, and S.~Levine.
\newblock Actionable models: Unsupervised offline reinforcement learning of
  robotic skills.
\newblock In M.~Meila and T.~Zhang, editors, \emph{Proceedings of the 38th
  International Conference on Machine Learning ({ICML}'21)}, volume 139 of
  \emph{Proceedings of Machine Learning Research}. PMLR, 2021.

\bibitem[Chen et~al.(2023)Chen, Lu, Ying, Su, and Zhu]{chen-iclr23a}
H.~Chen, C.~Lu, C.~Ying, H.~Su, and J.~Zhu.
\newblock Offline reinforcement learning via high-fidelity generative behavior
  modeling.
\newblock In \emph{The Eleventh International Conference on Learning
  Representations ({ICLR}'23)}. ICLR, 2023.

\bibitem[Choi et~al.(2026)Choi, Lee, and Seo]{choi-icml26a}
J.~Choi, S.~Lee, and S.~Seo.
\newblock Chain-of-goals hierarchical policy for long-horizon offline
  goal-conditioned {RL}.
\newblock In T.~Zhang, M.~Dudik, M.~Jaggi, A.~Agarwal, S.~Li, D.~Schuurmans,
  J.~Zhu, F.~Berkenkamp, H.~Dong, and A.~Bietti, editors, \emph{Proceedings of
  the 43rd International Conference on Machine Learning ({ICML}'26)}, volume
  306 of \emph{Proceedings of Machine Learning Research}, pages 20198--20219.
  PMLR, 2026.

\bibitem[Dierkes et~al.(2024)Dierkes, Cramer, Hoos, and Trimpe]{dierkes-rlj24a}
J.~Dierkes, E.~Cramer, H.~Hoos, and S.~Trimpe.
\newblock Combining automated optimisation of hyperparameters and reward shape.
\newblock \emph{Reinforcement Learning Journal}, 2024.

\bibitem[Eimer et~al.(2023)Eimer, Lindauer, and Raileanu]{eimer-icml23a}
T.~Eimer, M.~Lindauer, and R.~Raileanu.
\newblock Hyperparameters in reinforcement learning and how to tune them.
\newblock In  \citet{icml23}.

\bibitem[Engstrom et~al.(2020)Engstrom, Ilyas, Santurkar, Tsipras, Janoos,
  Rudolph, and Madry]{engstrom-iclr20a}
L.~Engstrom, A.~Ilyas, S.~Santurkar, D.~Tsipras, F.~Janoos, L.~Rudolph, and
  A.~Madry.
\newblock Implementation matters in deep {RL:} {A} case study on {PPO} and
  {TRPO}.
\newblock In \emph{The Eigth International Conference on Learning
  Representations ({ICLR}'20)}. ICLR, 2020.

\bibitem[Eysenbach et~al.(2021)Eysenbach, Salakhutdinov, and
  Levine]{eysenbach-iclr21a}
B.~Eysenbach, R.~Salakhutdinov, and S.~Levine.
\newblock C-learning: Learning to achieve goals via recursive classification.
\newblock In \emph{The Ninth International Conference on Learning
  Representations ({ICLR}'21)} \citet{iclr21}.

\bibitem[Eysenbach et~al.(2022)Eysenbach, Zhang, Levine, and
  Salakhutdinov]{eysenbach-neurips22a}
B.~Eysenbach, T.~Zhang, S.~Levine, and R.~Salakhutdinov.
\newblock Contrastive learning as goal-conditioned reinforcement learning.
\newblock In  \citet{neurips22}.

\bibitem[Fu et~al.(2020)Fu, Kumar, Nachum, Tucker, and Levine]{fu-arvix20a}
J.~Fu, A.~Kumar, O.~Nachum, G.~Tucker, and S.~Levine.
\newblock {D4RL}: Datasets for deep data-driven reinforcement learning.
\newblock \emph{arXiv preprint arXiv:2004.07219}, 2020.

\bibitem[Fujimoto and Gu(2021)]{fujimoto-neurips21a}
S.~Fujimoto and S.~Gu.
\newblock A minimalist approach to offline reinforcement learning.
\newblock In  \citet{neurips21}.

\bibitem[Fujimoto et~al.(2019)Fujimoto, Meger, and Precup]{fujimoto-icml19a}
S.~Fujimoto, D.~Meger, and D.~Precup.
\newblock Off-policy deep reinforcement learning without exploration.
\newblock In K.~Chaudhuri and R.~Salakhutdinov, editors, \emph{Proceedings of
  the 36th International Conference on Machine Learning ({ICML}'19)},
  volume~97. Proceedings of Machine Learning Research, 2019.

\bibitem[Ghosh et~al.(2019)Ghosh, Gupta, and Levine]{ghosh-iclr19a}
D.~Ghosh, A.~Gupta, and S.~Levine.
\newblock Learning actionable representations with goal-conditioned policies.
\newblock In \emph{The Seventh International Conference on Learning
  Representations ({ICLR}'19)}. ICLR, 2019.

\bibitem[Ghosh et~al.(2021)Ghosh, Gupta, Reddy, Fu, Devin, Eysenbach, and
  Levine]{ghosh-iclr21a}
D.~Ghosh, A.~Gupta, A.~Reddy, J.~Fu, C.~Devin, B.~Eysenbach, and S.~Levine.
\newblock Learning to reach goals via iterated supervised learning.
\newblock In \emph{The Ninth International Conference on Learning
  Representations ({ICLR}'21)} \citet{iclr21}.

\bibitem[Giammarino et~al.(2025)Giammarino, Ni, and
  Qureshi]{giammarino-arxiv25a}
V.~Giammarino, R.~Ni, and A.~Qureshi.
\newblock Physics-informed value learner for offline goal-conditioned
  reinforcement learning.
\newblock In  \citet{neurips25}.

\bibitem[Henderson et~al.(2018)Henderson, Islam, Bachman, Pineau, Precup, and
  Meger]{henderson-aaai18a}
P.~Henderson, R.~Islam, P.~Bachman, J.~Pineau, D.~Precup, and D.~Meger.
\newblock Deep reinforcement learning that matters.
\newblock In S.~McIlraith and K.~Weinberger, editors, \emph{Proceedings of the
  Thirty-Second Conference on Artificial Intelligence ({AAAI}'18)}. {AAAI}
  Press, 2018.

\bibitem[Hutter et~al.(2019)Hutter, Kotthoff, and Vanschoren]{hutter-book19a}
F.~Hutter, L.~Kotthoff, and J.~Vanschoren, editors.
\newblock \emph{Automated Machine Learning: Methods, Systems, Challenges}.
\newblock Springer, 2019.
\newblock Available for free at \url{http://automl.org/book}.

\bibitem[ICL(2021)]{iclr21}
\emph{Proceedings of the International Conference on Learning Representations
  ({ICLR}'21)}, 2021. ICLR.

\bibitem[ICL(2022)]{iclr22}
\emph{Proceedings of the International Conference on Learning Representations
  ({ICLR}'22)}, 2022. ICLR.

\bibitem[ICL(2025)]{iclr25}
\emph{Proceedings of the International Conference on Learning Representations
  ({ICLR}'25)}, 2025. ICLR.

\bibitem[Jackson et~al.(2025)Jackson, Berdica, Liesen, Whiteson, and
  Foerster]{jackson-neurips25a}
M.~Jackson, U.~Berdica, J.~Liesen, S.~Whiteson, and J.~Foerster.
\newblock A clean slate for offline reinforcement learning.
\newblock In  \citet{neurips25}.

\bibitem[Joe and Kuo(2008)]{joe-sjsci08a}
S.~Joe and F.~Kuo.
\newblock Constructing sobol sequences with better two-dimensional projections.
\newblock \emph{{SIAM} J. Sci. Comput.}, 2008.

\bibitem[Kaelbling(1993)]{kaelbling-ijcai93a}
L.~P. Kaelbling.
\newblock Learning to achieve goals.
\newblock In \emph{Proceedings of the 13th International Joint Conference on
  Artificial Intelligence ({IJCAI}'93)}, 1993.

\bibitem[Kobanda et~al.(2025)Kobanda, Radji, Petitbois, Maillard, and
  Portelas]{kobanda-arxiv25a}
A.~Kobanda, W.~Radji, M.~Petitbois, O.~Maillard, and R.~Portelas.
\newblock Offline goal-conditioned reinforcement learning with projective
  quasimetric planning.
\newblock \emph{CoRR}, abs/2506.18847, 2025.

\bibitem[Kostrikov et~al.(2022)Kostrikov, Nair, and Levine]{kostrikov-iclr22a}
I.~Kostrikov, A.~Nair, and S.~Levine.
\newblock Offline reinforcement learning with implicit q-learning.
\newblock In \emph{The Tenth International Conference on Learning
  Representations ({ICLR}'22)} \citet{iclr22}.

\bibitem[Koyejo et~al.(2022)Koyejo, Mohamed, Agarwal, Belgrave, Cho, and
  Oh]{neurips22}
S.~Koyejo, S.~Mohamed, A.~Agarwal, D.~Belgrave, K.~Cho, and A.~Oh, editors.
\newblock \emph{Proceedings of the 36th International Conference on Advances in
  Neural Information Processing Systems ({N}eur{IPS}'22)}, 2022. Curran
  Associates.

\bibitem[Krause et~al.(2023)Krause, Brunskill, Cho, Engelhardt, Sabato, and
  Scarlett]{icml23}
A.~Krause, E.~Brunskill, K.~Cho, B.~Engelhardt, S.~Sabato, and J.~Scarlett,
  editors.
\newblock \emph{Proceedings of the 40th International Conference on Machine
  Learning ({ICML}'23)}, volume 202 of \emph{Proceedings of Machine Learning
  Research}, 2023. PMLR.

\bibitem[Kumar et~al.(2019)Kumar, Fu, Soh, Tucker, and
  Levine]{kumar-neurips19a}
A.~Kumar, J.~Fu, M.~Soh, G.~Tucker, and S.~Levine.
\newblock Stabilizing off-policy q-learning via bootstrapping error reduction.
\newblock In  \citet{neurips19}.

\bibitem[Kumar et~al.(2020)Kumar, Zhou, Tucker, and Levine]{kumar-neurips20a}
A.~Kumar, A.~Zhou, G.~Tucker, and S.~Levine.
\newblock Conservative q-learning for offline reinforcement learning.
\newblock In H.~Larochelle, M.~Ranzato, R.~Hadsell, M.-F. Balcan, and H.~Lin,
  editors, \emph{Proceedings of the 34th International Conference on Advances
  in Neural Information Processing Systems ({N}eur{IPS}'20)}. Curran
  Associates, 2020.

\bibitem[Levine et~al.(2020)Levine, Kumar, Tucker, and Fu]{levine-arxiv20a}
S.~Levine, A.~Kumar, G.~Tucker, and J.~Fu.
\newblock Offline reinforcement learning: Tutorial, review, and perspectives on
  open problems.
\newblock \emph{arXiv:2005.01643 [cs.LG]}, 2020.

\bibitem[Ma et~al.(2022)Ma, Yan, Jayaraman, and Bastani]{ma-neurips22a}
Y.~Ma, J.~Yan, D.~Jayaraman, and O.~Bastani.
\newblock How far i'll go: Offline goal-conditioned reinforcement learning via
  f-advantage regression.
\newblock In  \citet{neurips22}.

\bibitem[Malan(2021)]{malan-algorithms21a}
K.~Malan.
\newblock A survey of advances in landscape analysis for optimisation.
\newblock \emph{Algorithms}, 2021.

\bibitem[Mao et~al.(2023)Mao, Zhang, Chen, Xu, and Ji]{mao-neurips23a}
Y.~Mao, H.~Zhang, C.~Chen, Y.~Xu, and X.~Ji.
\newblock Supported value regularization for offline reinforcement learning.
\newblock In  \citet{neurips23}.

\bibitem[Mohan et~al.(2023)Mohan, Benjamins, Wienecke, Dockhorn, and
  Lindauer]{mohan-automlconf23a}
A.~Mohan, C.~Benjamins, K.~Wienecke, A.~Dockhorn, and M.~Lindauer.
\newblock Autorl hyperparameter landscapes.
\newblock In A.~Faust, C.~White, F.~Hutter, R.~Garnett, and J.~Gardner,
  editors, \emph{Proceedings of the Second International Conference on
  Automated Machine Learning}. Proceedings of Machine Learning Research, 2023.

\bibitem[Mohan et~al.(2024)Mohan, Zhang, and Lindauer]{mohan-jair24a}
A.~Mohan, A.~Zhang, and M.~Lindauer.
\newblock Structure in deep reinforcement learning: {A} survey and open
  problems.
\newblock \emph{Journal of Artificial Intelligence Research}, 79, 2024.

\bibitem[Mohan et~al.(2026)Mohan, Chuck, Agarwal, and Zhang]{mohan-rlbrew26a}
A.~Mohan, C.~Chuck, S.~Agarwal, and A.~Zhang.
\newblock Is manipulation hard for successor measures?
\newblock In \emph{Third Reinforcement Learning Beyond Rewards Workshop at the
  Reinforcement Learning Conference}, 2026.

\bibitem[Myers et~al.(2025{\natexlab{a}})Myers, Ji, and
  Eysenbach]{myers-iclr25a}
V.~Myers, C.~Ji, and B.~Eysenbach.
\newblock Horizon generalization in reinforcement learning.
\newblock In \emph{The Thirteenth International Conference on Learning
  Representations ({ICLR}'25)} \citet{iclr25}.

\bibitem[Myers et~al.(2025{\natexlab{b}})Myers, Zheng, Eysenbach, and
  Levine]{myers-neurips25a}
V.~Myers, B.~Chunyuan Zheng, B.~Eysenbach, and S.~Levine.
\newblock Offline goal-conditioned reinforcement learning with quasimetric
  representations.
\newblock In  \citet{neurips25}.

\bibitem[Nair et~al.(2020)Nair, Gupta, Dalal, and Levine]{nair-arxiv20a}
A.~Nair, A.~Gupta, M.~Dalal, and S.~Levine.
\newblock Awac: Accelerating online reinforcement learning with offline
  datasets.
\newblock \emph{arXiv preprint arXiv:2006.09359}, 2020.

\bibitem[Nair et~al.(2018)Nair, Pong, Dalal, Bahl, Lin, and
  Levine]{nair-neurips18a}
A.~V. Nair, V.~Pong, M.~Dalal, S.~Bahl, S.~Lin, and S.~Levine.
\newblock Visual reinforcement learning with imagined goals.
\newblock In S.~Bengio, H.~Wallach, H.~Larochelle, K.~Grauman,
  N.~Cesa{-}Bianchi, and R.~Garnett, editors, \emph{Proceedings of the 31st
  International Conference on Advances in Neural Information Processing Systems
  ({N}eur{IPS}'18)}. Curran Associates, 2018.

\bibitem[Nasiriany et~al.(2019)Nasiriany, Pong, Lin, and
  Levine]{nasiriany-neurips19a}
S.~Nasiriany, V.~Pong, S.~Lin, and S.~Levine.
\newblock Planning with goal-conditioned policies.
\newblock In  \citet{neurips19}.

\bibitem[Obando{-}Ceron et~al.(2024)Obando{-}Ceron, Ara{\'{u}}jo, Courville,
  and Castro]{ceron-rlj24a}
J.~Obando{-}Ceron, J.~Ara{\'{u}}jo, A.~Courville, and P.~Castro.
\newblock On the consistency of hyper-parameter selection in value-based deep
  reinforcement learning.
\newblock \emph{Reinforcement Learning Journal}, 2024.

\bibitem[Oh et~al.(2023)Oh, Naumann, Globerson, Saenko, Hardt, and
  Levine]{neurips23}
A.~Oh, T.~Naumann, A.~Globerson, K.~Saenko, M.~Hardt, and S.~Levine, editors.
\newblock \emph{Proceedings of the 37th International Conference on Advances in
  Neural Information Processing Systems ({N}eur{IPS}'23)}, 2023. Curran
  Associates.

\bibitem[Opryshko et~al.(2025)Opryshko, Quan, Voelcker, Du, and
  Gilitschenski]{opryshko-arxiv25a}
E.~Opryshko, J.~Quan, C.~Voelcker, Y.~Du, and I.~Gilitschenski.
\newblock Test-time graph search for goal-conditioned reinforcement learning.
\newblock \emph{CoRR}, abs/2510.07257, 2025.

\bibitem[Paine et~al.(2020)Paine, Paduraru, Michi, G{\"{u}}l{\c{c}}ehre, Zolna,
  Novikov, Wang, and de~Freitas]{paine-arxiv20a}
T.~Paine, C.~Paduraru, A.~Michi, C.~G{\"{u}}l{\c{c}}ehre, K.~Zolna, A.~Novikov,
  Z.~Wang, and N.~de~Freitas.
\newblock Hyperparameter selection for offline reinforcement learning.
\newblock \emph{CoRR}, 2020.
\newblock URL \url{https://arxiv.org/abs/2007.09055}.

\bibitem[Park et~al.(2023)Park, Ghosh, Eysenbach, and Levine]{park-neurips23a}
S.~Park, D.~Ghosh, B.~Eysenbach, and S.~Levine.
\newblock Hiql: Offline goal-conditioned rl with latent states as actions.
\newblock In  \citet{neurips23}.

\bibitem[Park et~al.(2024)Park, Frans, Levine, and Kumar]{park-neurips24a}
S.~Park, K.~Frans, S.~Levine, and A.~Kumar.
\newblock Is value learning really the main bottleneck in offline {RL}?
\newblock In \emph{Proceedings of the 38th International Conference on Advances
  in Neural Information Processing Systems ({N}eur{IPS}'24)} \citet{neurips24}.

\bibitem[Park et~al.(2025{\natexlab{a}})Park, Frans, Eysenbach, and
  Levine]{park-iclr25a}
S.~Park, K.~Frans, B.~Eysenbach, and S.~Levine.
\newblock Ogbench: Benchmarking offline goal-conditioned rl.
\newblock In \emph{The Thirteenth International Conference on Learning
  Representations ({ICLR}'25)} \citet{iclr25}.

\bibitem[Park et~al.(2025{\natexlab{b}})Park, Frans, Mann, Eysenbach, Kumar,
  and Levine]{park-neurips25a}
S.~Park, K.~Frans, D.~Mann, B.~Eysenbach, A.~Kumar, and S.~Levine.
\newblock Horizon reduction makes {RL} scalable.
\newblock In  \citet{neurips25}.

\bibitem[Park et~al.(2025{\natexlab{c}})Park, Li, and Levine]{park-icml25a}
S.~Park, Q.~Li, and S.~Levine.
\newblock Flow {Q}-learning.
\newblock In \emph{Proceedings of the 42nd International Conference on Machine
  Learning (ICML'25)}, 2025{\natexlab{c}}.

\bibitem[Park et~al.(2026)Park, Mann, and Levine]{park-arxiv25a}
S.~Park, D.~Mann, and S.~Levine.
\newblock Dual goal representations.
\newblock In \emph{The Fourteenth International Conference on Learning
  Representations}, 2026.

\bibitem[Parker-Holder et~al.(2022)Parker-Holder, Rajan, Song, Biedenkapp,
  Miao, Eimer, Zhang, Nguyen, Calandra, Faust, Hutter, and
  Lindauer]{parkerholder-jair22a}
J.~Parker-Holder, R.~Rajan, X.~Song, A.~Biedenkapp, Y.~Miao, T.~Eimer,
  B.~Zhang, V.~Nguyen, R.~Calandra, A.~Faust, F.~Hutter, and M.~Lindauer.
\newblock Automated reinforcement learning ({A}uto{RL}): A survey and open
  problems.
\newblock \emph{Journal of Artificial Intelligence Research (JAIR)},
  74:\penalty0 517--568, 2022.

\bibitem[Peng et~al.(2019)Peng, Kumar, Zhang, and Levine]{peng-arxiv19a}
X.~Peng, A.~Kumar, G.~Zhang, and S.~Levine.
\newblock Advantage-weighted regression: Simple and scalable off-policy
  reinforcement learning.
\newblock \emph{arXiv preprint arXiv:1910.00177}, 2019.

\bibitem[Peters and Schaal(2007)]{peters-icml07a}
J.~Peters and S.~Schaal.
\newblock Reinforcement learning by reward-weighted regression for operational
  space control.
\newblock In \emph{icml07}, 2007.

\bibitem[Pitzer and Affenzeller(2012)]{pitzer-fitnesslandscape12a}
E.~Pitzer and M.~Affenzeller.
\newblock A comprehensive survey on fitness landscape analysis.
\newblock \emph{Recent Advances in Intelligent Engineering Systems}, 2012.

\bibitem[Ranzato et~al.(2021)Ranzato, Beygelzimer, Nguyen, Liang, Vaughan, and
  Dauphin]{neurips21}
M.~Ranzato, A.~Beygelzimer, K.~Nguyen, P.~Liang, J.~Vaughan, and Y.~Dauphin,
  editors.
\newblock \emph{Proceedings of the 35th International Conference on Advances in
  Neural Information Processing Systems ({N}eur{IPS}'21)}, 2021. Curran
  Associates.

\bibitem[Schaul et~al.(2015)Schaul, Horgan, Gregor, and Silver]{schaul-icml15a}
T.~Schaul, D.~Horgan, K.~Gregor, and D.~Silver.
\newblock Universal value function approximators.
\newblock In F.~Bach and D.~Blei, editors, \emph{Proceedings of the 32nd
  International Conference on Machine Learning ({ICML}'15)}, volume~37.
  Omnipress, 2015.

\bibitem[Shen et~al.(2026)Shen, Chuck, Feng, and Huang]{shen-arxiv26a}
T.~Shen, C.~Chuck, F.~Feng, and B.~Huang.
\newblock Learning object manipulation from scratch via contrastive
  interaction.
\newblock \emph{arXiv preprint arXiv:2606.11525}, 2026.

\bibitem[Sikchi et~al.(2024)Sikchi, Chitnis, Touati, Geramifard, Zhang, and
  Niekum]{sikchi-iclr24a}
H.~Sikchi, R.~Chitnis, A.~Touati, A.~Geramifard, A.~Zhang, and S.~Niekum.
\newblock {SMORE}: Score models for offline goal-conditioned reinforcement
  learning.
\newblock In \emph{The Twelfth International Conference on Learning
  Representations ({ICLR}'24)}. ICLR, 2024.

\bibitem[Sobol(1967)]{sobol-cmmp67a}
I.~Sobol.
\newblock On the distribution of points in a cube and the approximate
  evaluation of integrals.
\newblock \emph{USSR Computational Mathematics and Mathematical Physics},
  7\penalty0 (4):\penalty0 86--112, 1967.

\bibitem[Venugopal et~al.(2026)Venugopal, Chen, Wu, Zheng, Eysenbach, and
  Schneider]{venugopal-arxiv26a}
A.~Venugopal, J.~Chen, X.~Wu, C.~Zheng, B.~Eysenbach, and J.~Schneider.
\newblock Occupancy reward shaping: Improving credit assignment for offline
  goal-conditioned reinforcement learning.
\newblock In \emph{The Fourteenth International Conference on Learning
  Representations}, 2026.

\bibitem[Wallach et~al.(2019)Wallach, Larochelle, Beygelzimer, d'Alche Buc,
  Fox, and Garnett]{neurips19}
H.~Wallach, H.~Larochelle, A.~Beygelzimer, F.~d'Alche Buc, E.~Fox, and
  R.~Garnett, editors.
\newblock \emph{Proceedings of the 33rd International Conference on Advances in
  Neural Information Processing Systems ({N}eur{IPS}'19)}, 2019. Curran
  Associates.

\bibitem[Wang et~al.(2023)Wang, Torralba, Isola, and Zhang]{wang-icml23a}
T.~Wang, A.~Torralba, P.~Isola, and A.~Zhang.
\newblock Optimal goal-reaching reinforcement learning via quasimetric
  learning.
\newblock In  \citet{icml23}.

\bibitem[Yang et~al.(2022)Yang, Lu, Li, Sun, Fang, Du, Li, Han, and
  Zhang]{yang-iclr22a}
R.~Yang, Y.~Lu, W.~Li, H.~Sun, M.~Fang, Y.~Du, X.~Li, L.~Han, and C.~Zhang.
\newblock Rethinking goal-conditioned supervised learning and its connection to
  offline rl.
\newblock In \emph{The Tenth International Conference on Learning
  Representations ({ICLR}'22)} \citet{iclr22}.

\bibitem[Zheng et~al.(2026)Zheng, Myers, Eysenbach, and Levine]{zheng-arxiv25a}
B.~Zheng, V.~Myers, B.~Eysenbach, and S.~Levine.
\newblock Multistep quasimetric learning for scalable goal-conditioned
  reinforcement learning.
\newblock In \emph{International Conference on Learning Representations
  (ICLR)}, 2026.

\bibitem[Zhou and Kao(2025)]{zhou-neurips25a}
J.~Zhou and J.~Kao.
\newblock Flattening hierarchies with policy bootstrapping.
\newblock In  \citet{neurips25}.

\end{thebibliography}

\newpage
\appendix
\crefalias{section}{appsec}
\crefalias{subsection}{appsubsec}
\crefalias{subsubsection}{appsubsubsec}
\part{Appendix}
{
  \hypersetup{linkcolor=black}
  \parttoc
}

\section{Related Work}
\label{sec:appendix:related}

We organize related work around the three questions underlying our comparison: what each method learns, how it extracts a policy, and how success and sensitivity to tuning are measured.

\paragraph{Offline goal-conditioned reinforcement learning.}
Offline \GCRL{} methods address the same goal-reaching problem through different learning objectives.
They build on goal-conditioned value functions and universal value function approximators~\citep{kaelbling-ijcai93a,schaul-icml15a}, hindsight relabeling~\citep{andrychowicz-neurips17a}, and extensions to visual observations, learned representations, and planning~\citep{nair-neurips18a,ghosh-iclr19a,nasiriany-neurips19a,mohan-jair24a}.
Goal-conditioned supervised learning offers an imitation-based approach, treating relabeled trajectories as demonstrations for the goals they reach~\citep{ghosh-iclr21a,yang-iclr22a}.

In the offline setting, Actionable Models combine goal-conditioned Q-learning with hindsight relabeling~\citep{chebotar-icml21a}, GoFAR derives an objective from state-occupancy matching~\citep{ma-neurips22a}, and SMORe learns unnormalized scores through mixture-distribution matching~\citep{sikchi-iclr24a}.
Related approaches connect goal reaching to other prediction problems: 
C-learning estimates future-state distributions through recursive classification~\citep{eysenbach-iclr21a}, 
while Contrastive RL connects contrastive representation learning to goal-conditioned values~\citep{eysenbach-neurips22a}.
Quasimetric methods represent goal reaching through asymmetric distances~\citep{wang-icml23a,myers-neurips25a,zheng-arxiv25a}, with related work examining generalization to longer horizons~\citep{myers-iclr25a}.
All these objectives produce different estimates for guiding policy learning.
Our comparison examines the advantages supplied by four representative methods to a shared policy extractor.

\paragraph{Improvements to goal learning and policy extraction.}
Recent methods improve goal-reaching performance by changing value learning, policy structure, or test-time planning.
HIQL extracts hierarchical policies from an action-free goal-conditioned value function~\citep{park-neurips23a}.
Option-aware Temporally Abstracted value learning (OTA) improves high-level advantage estimates through temporally abstracted value updates~\citep{ahn-neurips25a}.
Policy bootstrapping trains a flat goal-conditioned policy from subgoal-conditioned policies~\citep{zhou-neurips25a}, while Chain-of-Goals generates sequences of latent subgoals followed by a primitive action within a unified architecture~\citep{choi-icml26a}.

Other approaches reduce the effective learning horizon~\citep{park-neurips25a}, introduce projective quasimetric planning~\citep{kobanda-arxiv25a}, change goal representations~\citep{park-arxiv25a}, derive shaped rewards from learned occupancy models~\citep{venugopal-arxiv26a}, or regularize values using the Eikonal equation~\citep{giammarino-arxiv25a}.
Test-Time Graph Search uses learned distance or cost estimates to select intermediate goals over dataset states for a frozen policy~\citep{opryshko-arxiv25a}.
An improvement in tuned performance leaves open whether high success also persists across a broader range of hyperparameter settings.
This is an additional evaluation question that our protocol could address for these methods.

\paragraph{Policy extraction as an offline RL bottleneck.}
Offline RL methods address errors from unsupported actions through constraints on action selection~\citep{fujimoto-icml19a,kumar-neurips19a} or regularization of value estimates~\citep{kumar-neurips20a,mao-neurips23a}.
Reward-weighted regression, \AWR{}, and AWAC train policies by weighting the likelihood of dataset actions according to estimated returns or advantages~\citep{peters-icml07a,peng-arxiv19a,nair-arxiv20a}.
IQL learns values without querying out-of-distribution actions and extracts a policy through advantage-weighted regression~\citep{kostrikov-iclr22a}.
\citet{park-neurips24a} distinguish value learning, policy extraction, and policy generalization, showing that extraction choices and generalization to evaluation states can substantially limit offline RL performance.

These findings motivate controlling the extraction objective in our comparison.
We evaluate \AlgGCIQL{}, \AlgGCIVL{}, \AlgQRL{}, and \AlgCRL{} with a shared \AWR{} objective.
The methods supply different advantage estimates, and the \AWR{} temperature determines how strongly those estimates affect imitation weights.
We examine these weights alongside the success of the resulting policies.

\paragraph{Benchmarks and evaluation protocols for \GCRL.}
D4RL standardizes fixed-dataset evaluation in offline RL, including normalized-return scores~\citep{fu-arvix20a}.
\OGBench{} benchmarks offline \GCRL{} across locomotion and manipulation, including visual, stochastic, and long-horizon tasks that test capabilities such as trajectory stitching and temporal reasoning~\citep{park-iclr25a}.
\textsc{JaxGCRL} provides complementary infrastructure for efficient self-supervised \GCRL{} experiments with online environment interaction~\citep{bortkiewicz-iclr25a}.

A peak success rate shows that at least one configuration can make a method succeed, but leaves open how much of the training and extraction space leads to usable behavior.
This distinction matters because offline \GCRL{} relies on a policy extractor to obtain a policy from learned value functions.
A single benchmark score therefore mixes three quantities: the quality of the learned signal, the extractor's use of that signal, and the size of the hyperparameter region where the combination works.
Our evaluation separates these quantities using trainability landscapes and post-hoc diagnostics of the learned advantage signal and \AWR{} weight distribution.

\paragraph{Hyperparameter sensitivity and landscapes.}
Deep RL is sensitive to hyperparameters, random seeds, implementation details, and evaluation protocols~\citep{henderson-aaai18a,engstrom-iclr20a,agarwal-neurips21a,parkerholder-jair22a}.
Large-scale studies show that a small number of hyperparameters can dominate performance variation, and that tuning procedures can change algorithm comparisons~\citep{eimer-icml23a,ceron-rlj24a,adkins-neurips24a}.
Hyperparameter interactions can also create narrow viable regions even when each individual knob appears mild~\citep{dierkes-rlj24a}.
These findings support evaluation protocols that report more than the single best configuration.

Landscape analysis studies performance as a function over configurations, including modality, robustness, and the volume of near-optimal regions~\citep{hutter-book19a,pitzer-fitnesslandscape12a,malan-algorithms21a}.
Prior work in automated RL (AutoRL) uses response surfaces to study how good hyperparameters move during online RL training~\citep{mohan-automlconf23a}.
We apply landscape analysis to offline \GCRL{} with fixed datasets and a shared \AWR{} objective, and relate landscape breadth to policy-extraction diagnostics.
Fixing the datasets removes changes in collected experience from the comparison, although value estimates and policies continue to change during training.

\paragraph{Diagnostics of learned goal-conditioned signals.}
Prior work motivates examining the values and advantages used to train policies.
C-learning and Contrastive RL connect goal-conditioned values to future-state discrimination and density-ratio estimation~\citep{eysenbach-iclr21a,eysenbach-neurips22a}.
OTA identifies incorrect high-level advantage signs as a source of poor subgoal selection in long-horizon tasks~\citep{ahn-neurips25a}.
These findings suggest that inspecting learned estimates can provide information beyond the final policy score.

Our diagnostics examine advantage rankings and imitation weights at trained checkpoints.
\FADcap{} compares the advantages assigned to future and random goals for the same transition, using pairwise comparisons and goal rankings.
Weight-concentration diagnostics describe how \AWR{} weights are distributed across dataset transitions.
We evaluate how these measurements relate to policy success and landscape breadth, rather than treating stronger goal discrimination or greater weight concentration as evidence of a better policy on its own.

\section{Hyperparameters}
\label{sec:appendix:hps}

This appendix reports the hyperparameter ranges and defaults used in our experiments.
The trainability landscapes in \Cref{sec:appendix:landscapes} use a controlled two-dimensional Sobol design over learning rate and \AWR{} temperature.

\paragraph{Landscape hyperparameters.}
For landscape construction, we sweep the same two hyperparameters for all four algorithms: learning rate $\eta$ and \AWR{} temperature $\alpha$.
All landscape runs use the \AWR{} actor loss. 
We sample $64=2^6$ configurations with a two-dimensional Sobol design and evaluate each configuration with five training seeds and ten evaluation episodes. 
Unless otherwise stated, all other algorithm parameters are kept at their adapted OGBench defaults.

\begin{table}[ht]
  \centering
  \caption{
    Hyperparameter ranges used for landscape construction. The landscapes are
    restricted to the shared $\eta$--$\alpha$ plane so that trainability can be
    compared directly across \AlgCRL{}, \AlgQRL{}, \AlgGCIQL{}, and \AlgGCIVL{}.
  }
  \vspace{0.5em}
  \label{tab:hp-ranges-landscape}
  \begin{tabular}{llll}
    \toprule
    Hyperparameter & Algorithms & Range & Sampling scale \\
    \midrule
    Learning rate $\eta$
      & \AlgCRL{}, \AlgQRL{}, \AlgGCIQL{}, \AlgGCIVL{}
      & $[10^{-6}, 10^{-2}]$
      & log-uniform \\
    \AWR{} temperature $\alpha$
      & \AlgCRL{}, \AlgQRL{}, \AlgGCIQL{}, \AlgGCIVL{}
      & $[0,30]$
      & uniform \\
    \bottomrule
  \end{tabular}
\end{table}

The Sobol samples are generated in $[0,1]^2$ and then mapped into the hyperparameter domain. For a log-scaled hyperparameter $x$ with lower bound $a$ and upper bound $b$, a Sobol coordinate $u \in [0,1]$ is mapped as
\(
    x(u)
    =
    10^{\log_{10}(a) + u(\log_{10}(b)-\log_{10}(a))}.
\)
For a linearly scaled hyperparameter, we use
\(
    x(u)
    =
    a + u(b-a).
\)
The same sampled configuration set is reused across the four phases of a landscape run. Phase construction, checkpoint resets, and mobility analysis are described separately in \Cref{sec:appendix:landscapes}.

\paragraph{Default configurations.}
Default values are adapted from the OGBench implementations and modified to use the \AWR{} actor loss. 
These defaults are used whenever a parameter is not swept.
For the landscape experiments, this means that only $\eta$ and $\alpha$ vary; all remaining entries in \Cref{tab:hps:defaults} are fixed.

\begin{table}[ht]
  \centering
  \caption{
    Default hyperparameter values for the four algorithms. Parameters marked
    with -- are not used by the corresponding algorithm.
  }
  \vspace{0.5em}
  \label{tab:hps:defaults}
  \begin{tabular}{lllll}
    \toprule
    Hyperparameter & \AlgCRL{} & \AlgQRL{} & \AlgGCIQL{} & \AlgGCIVL{} \\
    \midrule
    Learning rate $\eta$
      & $3\cdot10^{-4}$ & $3\cdot10^{-4}$ & $3\cdot10^{-4}$ & $3\cdot10^{-4}$ \\
    Batch size
      & $1024$ & $1024$ & $1024$ & $1024$ \\
    Discount $\gamma$
      & $0.99$ & $0.99$ & $0.99$ & $0.99$ \\
    Latent dimension
      & $512$ & $512$ & -- & -- \\
    Constant actor std.
      & true & true & true & true \\
    $\epsilon$
      & -- & $0.05$ & -- & -- \\
    Target-network rate $\tau$
      & -- & -- & $0.005$ & $0.005$ \\
    Expectile
      & -- & -- & $0.9$ & $0.9$ \\
    \AWR{} temperature $\alpha$
      & $3.0$ & $3.0$ & $3.0$ & $3.0$ \\
    Value trajectory-goal probability
      & $1.0$ & $0.0$ & $0.5$ & $0.5$ \\
    Value current-goal probability
      & $0.0$ & $0.0$ & $0.2$ & $0.2$ \\
    Value random-goal probability
      & $0.0$ & $1.0$ & $0.3$ & $0.3$ \\
    Value geometric sampling
      & true & true & true & true \\
    Actor trajectory-goal probability
      & $1.0$ & $1.0$ & $1.0$ & $1.0$ \\
    Actor current-goal probability
      & $0.0$ & $0.0$ & $0.0$ & $0.0$ \\
    Actor random-goal probability
      & $0.0$ & $0.0$ & $0.0$ & $0.0$ \\
    Actor geometric sampling
      & false & false & false & false \\
    Actor loss
      & \AWR{} & \AWR{} & \AWR{} & \AWR{} \\
    Advantage normalization
      & false & false & false & false \\
    \bottomrule
  \end{tabular}
\end{table}

\paragraph{Diagnostic hyperparameters.}
Some quantities used in the analysis are not training hyperparameters.
In particular, training clips \AWR{} weights at $w_{\max}=100$ for all methods, whereas the sweep over $w_{\max}$ in \Cref{app:wmax_sensitivity} only recomputes normalized weights, effective sample size, and \TopFivePctMass{} post hoc, without retraining.
This sweep is therefore reported with the robustness diagnostics rather than in the training hyperparameter tables.

\section{Hyperparameter Landscape Construction and Analysis}
\label{sec:appendix:landscapes}

\subsection{Phased training and phase selection}
\label{sec:appendix:phases}

\begin{figure}[ht]
  \includegraphics[width=\textwidth]{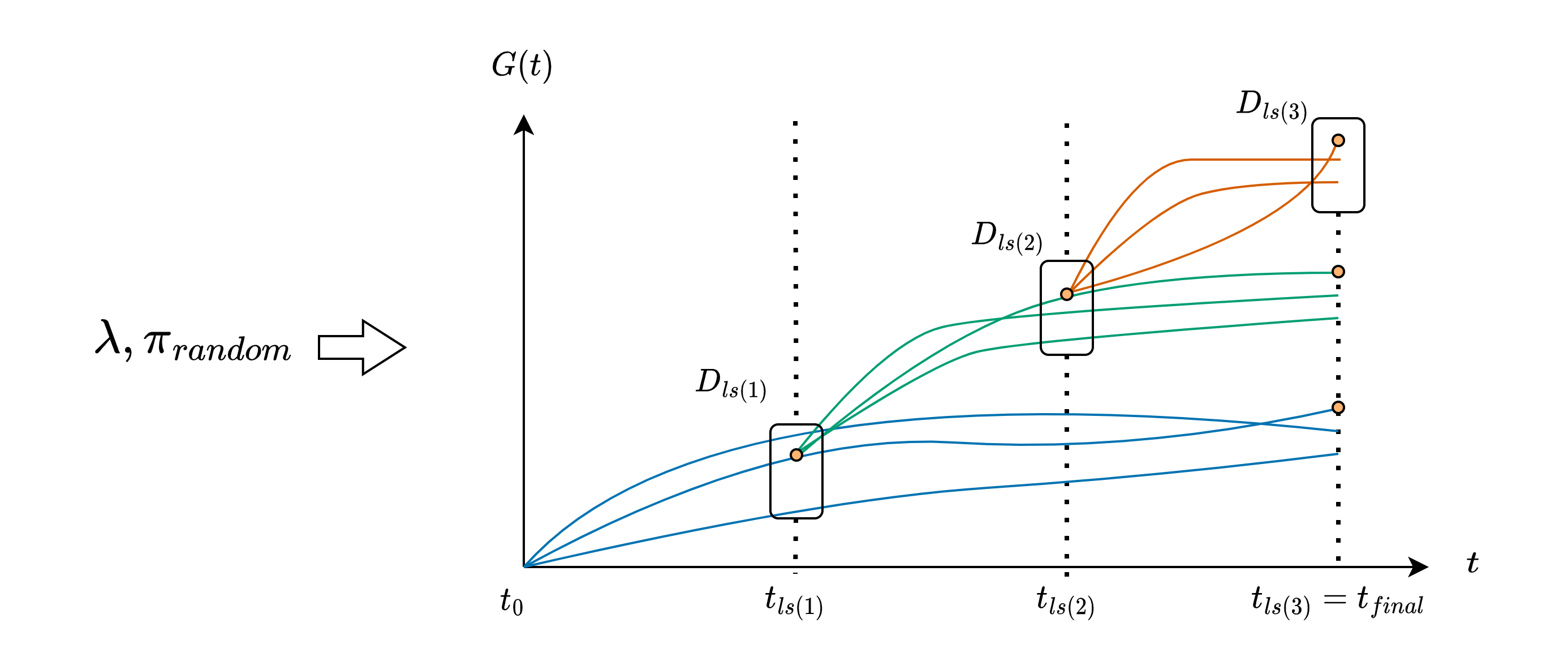}
  \caption{
    \textbf{Phased training~\citep{mohan-automlconf23a}.}
    At $t_0$, all sampled configurations start training and are evaluated at $t_{\mathrm{ls}}(1)$.
    In the second phase, all configurations restart from the checkpoint of the best configuration of phase one and are re-evaluated at $t_{\mathrm{ls}}(2)$.
    We select the best configuration at the end of each phase, $t_{\mathrm{ls}}(i)$.
    }
  \label{fig:approach:phased-training}
\end{figure}

We collect landscape data during training with the phased pipeline of \citet{mohan-automlconf23a} (\Cref{fig:approach:phased-training}).
All configurations in $C$ are trained from $t_0$ to $t_{\mathrm{ls}}(1)$ and evaluated there.
Each subsequent phase restarts all configurations from the checkpoint of the best configuration of the previous phase (\Cref{sec:appendix:landscape-datasets}).
This gives every configuration the same starting point and implicitly schedules the hyperparameters.
The process is repeated until all phases are completed.
Unlike \citet{mohan-automlconf23a}, we select the checkpoint configuration by its performance at the end of each phase, not after training it until the end of the last phase, because the latter worsens performance in our setting.

Phase boundaries should lie neither after an algorithm has mostly converged nor so early that it is far from its potential.
For each algorithm and dataset, we therefore train a reference agent for $10^6$ gradient steps and record its success rate. The agent uses the OGBench default hyperparameters~\citep{park-iclr25a}, except that the policy extractor is \AWR{} for all four methods.
We evaluate more frequently early in training, where performance changes most; the evaluation steps (in thousands) are
$\{1, 2.5, 5, 7.5, 10, 20, 30, 40, 50, 75, 100, 150, 200, 300, 400, 500, 600, 700, 800, 900, 1000\}$,
and the training curve linearly interpolates between them.
We define $100\%$ training progress as the first step at which success reaches $95\%$ of the maximum success observed in the reference run.
The phase boundaries $t_{\mathrm{ls}}(i)$ are set at $25\%$, $50\%$, $75\%$, and $100\%$ of this progress (\Cref{tab:convergence:phases}).

\begin{table}[ht]
  \centering

\begin{tabular}{llrrrr}
\toprule
Algorithm & Environment & \multicolumn{4}{c}{Phase Steps} \\
\cmidrule(l){3-6}
 & & Phase 1 & Phase 2 & Phase 3 & Phase 4 \\
\midrule
\multirow[t]{4}{*}{CRL} & antmaze-medium & 39772 & 79545 & 119318 & 159090 \\
 & antmaze-large & 231625 & 463250 & 694875 & 926500 \\
 & cube & 78879 & 157758 & 236637 & 315517 \\
 & scene & 13250 & 26500 & 39750 & 53000 \\
\midrule
\multirow[t]{4}{*}{GCIQL} & antmaze-medium & 205069 & 410138 & 615208 & 820277 \\
 & antmaze-large & 156574 & 313148 & 469722 & 626296 \\
 & cube & 237395 & 474791 & 712187 & 949583 \\
 & scene & 211428 & 422857 & 634285 & 845714 \\
\midrule
\multirow[t]{4}{*}{GCIVL} & antmaze-medium & 216448 & 432896 & 649345 & 865793 \\
 & antmaze-large & 10741 & 21482 & 32223 & 42964 \\
 & cube & 11722 & 23444 & 35166 & 46888 \\
 & scene & 142250 & 284500 & 426750 & 569000 \\
\midrule
\multirow[t]{4}{*}{QRL} & antmaze-medium & 77687 & 155375 & 233062 & 310750 \\
 & antmaze-large & 100555 & 201111 & 301666 & 402222 \\
 & cube & 202250 & 404500 & 606750 & 809000 \\
 & scene & 20937 & 41875 & 62812 & 83750 \\
\bottomrule
\end{tabular}

  \vspace{0.5em}
  \caption{
    \textbf{Phase boundaries.}
    Gradient steps at the end of each phase for each algorithm and environment.
    Phase~4 corresponds to $100\%$ training progress of a reference run with the OGBench default hyperparameters, except for the policy extractor, which is \AWR{} for all methods.
  }
  \label{tab:convergence:phases}
\end{table}

The default configuration may be suboptimal for a given algorithm and dataset, so the phase boundaries are approximate.
This is acceptable because the phases only need to place the landscapes at roughly comparable stages of training.
Because convergence behavior varies across algorithms and datasets, phase lengths differ substantially.
Phase indices therefore mark comparable stages of learning within each algorithm and dataset, and phase lengths say little about relative convergence speed.

\subsection{Hyperparameter spaces and configuration sampling}
\label{sec:appendix:hyperparameters}

The search space $\Lambda=\Lambda_\eta\times\Lambda_\alpha$ contains the learning rate and the \AWR{} temperature (\Cref{tab:hp-ranges-landscape}).
Both dimensions are bounded; the learning rate is sampled log-uniformly and the temperature uniformly.

For each algorithm, we sample a fixed configuration set
$C = \{\lambda^{(1)}, \dots, \lambda^{(N)}\} \subseteq \Lambda$ prior to training.
The same configuration set is reused across all phases to enable
direct comparison of landscape geometry.
Configuration sampling is performed once and kept fixed throughout the study.

\subsection{Landscape datasets and checkpoint resets}
\label{sec:appendix:landscape-datasets}

For each phase $i$, all configurations $\lambda \in C$ are trained starting from a shared reference checkpoint, corresponding to the configuration with the best IQM in phase $i-1$ under the same algorithm and data regime.
This reset rule ensures that landscape measurements reflect the \emph{local} effect of hyperparameters within each phase, starting from a shared state. 
We compare this reset protocol against straight-through (uninterrupted) training for \AlgGCIQL{} and \AlgQRL{} on \EnvAntMazeM{} in \Cref{app:straight-vs-phased}.
At the first evaluation step the two protocols rank configurations almost identically (Spearman $0.94$ for \AlgGCIQL{}, $0.98$ for \AlgQRL{}). 
Afterward they diverge for \AlgGCIQL{} ($\rho \approx -0.2$), while \AlgQRL{}'s rankings stay positively correlated ($0.36$ to $0.51$).
However, the pattern of \AlgQRL{} being broader than \AlgGCIQL{} persists under straight-through training, demonstrating that it holds without the resets.

Evaluating each configuration at $t_{\mathrm{ls}}(i)$ yields a landscape dataset
\[
D_{\mathrm{ls}}^{(i)} = \{(\lambda^{(j)}, f_i(\lambda^{(j)}))\}_{j=1}^{N},
\]
where $f_i(\lambda)$ denotes the aggregated evaluation metric over random seeds.
Unless otherwise stated, the evaluation metrics correspond to the success rate averaged
over $K$ evaluation episodes.

\subsection{Landscape modeling and visualization}
\label{sec:appendix:landscape-modeling}

To visualize the two-dimensional hyperparameter landscapes, we follow
\cite{mohan-automlconf23a} and fit independent Gaussian process regressors
(IGPRs) to the observed configuration--performance pairs.
Each IGPR is trained on the two swept hyperparameters, $\eta$ and $\alpha$.

We emphasize that IGPRs are used \emph{only} for visualization and qualitative
inspection of landscape shape.
All quantitative metrics reported in the main paper are computed directly on the
evaluated configurations without reliance on surrogate predictions.

\subsection{Reproducibility details}
\label{sec:appendix:reproducibility}

All experiments are conducted with fixed random seeds per configuration and multiple evaluation rollouts.
Compute budgets, seed counts, and implementation details are reported alongside hyperparameter tables.
The code for landscape construction, analysis, and visualization is available at \url{https://github.com/LUH-AI/GCRL-Landscapes}.

\section{Extended Trainability Landscapes (Mobility Plots)}
\label{app:mobility_plots}

In this section, we provide extended trainability landscapes (mobility plots) for all evaluated environments: \EnvAntMazeM{}, \EnvAntMazeL{}, \EnvCube{}, and \EnvScene{}. For each environment, we show the response surfaces indicating the hyperparameter configurations that achieve at least 80\% of the best observed normalized return (top-20\% regions), matching the threshold used in the main text.

These visualizations track how the optimal regions in the learning-rate ($\eta$) and \AWR{} temperature ($\alpha$) plane move and evolve across the four training phases.

\subsection{\EnvAntMazeM{}}

\noindent\begin{minipage}{\textwidth}
  \captionsetup{type=figure, justification=raggedright, singlelinecheck=false, font=small}
  \centering
  \begin{minipage}[c]{0.66\textwidth}
    \centering
    \begin{tabular}{@{}cc@{}}
      \includegraphics[width=0.48\linewidth]{figures/awr-mobility/antmaze-medium/80/mobility-gciql-antmaze-medium-navigate-v0.pdf}
      &
      \includegraphics[width=0.48\linewidth]{figures/awr-mobility/antmaze-medium/80/mobility-gcivl-antmaze-medium-navigate-v0.pdf}
      \\
      \footnotesize (a) \AlgGCIQL{} & \footnotesize (b) \AlgGCIVL{} \\[4pt]
      \includegraphics[width=0.48\linewidth]{figures/awr-mobility/antmaze-medium/80/mobility-crl-antmaze-medium-navigate-v0.pdf}
      &
      \includegraphics[width=0.48\linewidth]{figures/awr-mobility/antmaze-medium/80/mobility-qrl-antmaze-medium-navigate-v0.pdf}
      \\
      \footnotesize (c) \AlgCRL{} & \footnotesize (d) \AlgQRL{}
    \end{tabular}
  \end{minipage}\hfill
  \begin{minipage}[c]{0.32\textwidth}
    \caption{\textbf{\EnvAntMazeM{} (top-20\% regions).}
    Trainability landscapes in the $\eta$--$\alpha$ plane. \AlgCRL{} and \AlgQRL{} have broad, stable top-20\% regions. \AlgGCIVL{} has a smaller region. \AlgGCIQL{} has narrower regions that move more across phases.
    }
    \label{fig:mobility_antmaze_medium_80}
  \end{minipage}
\end{minipage}

\subsection{\EnvAntMazeL{}}

\noindent\begin{minipage}{\textwidth}
  \captionsetup{type=figure, justification=raggedright, singlelinecheck=false, font=small}
  \centering
  \begin{minipage}[c]{0.66\textwidth}
    \centering
    \begin{tabular}{@{}cc@{}}
      \includegraphics[width=0.48\linewidth]{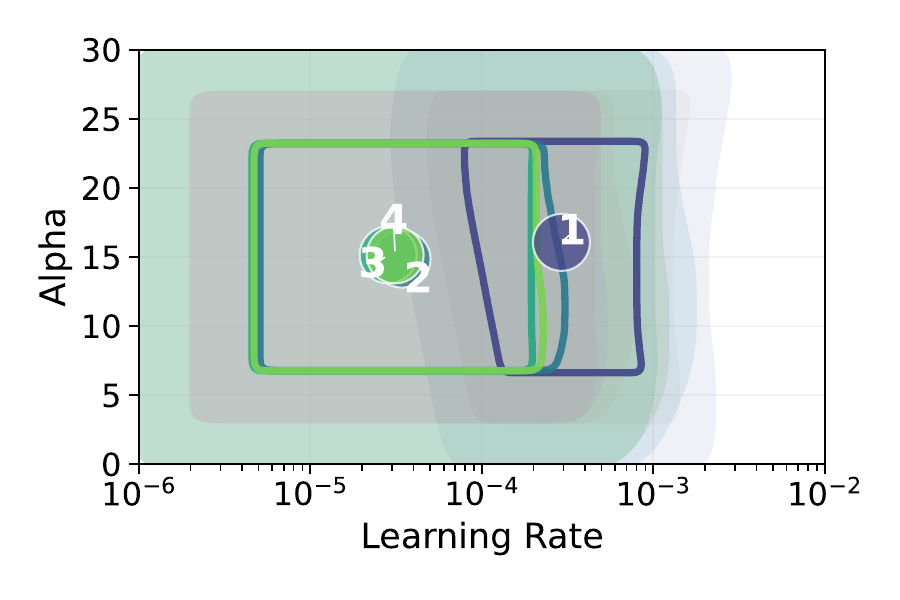}
      &
      \includegraphics[width=0.48\linewidth]{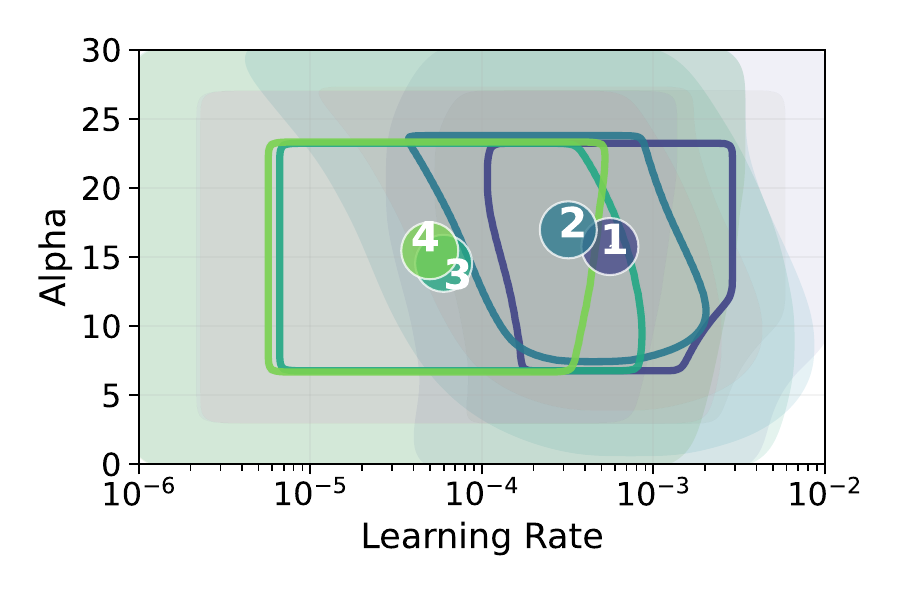}
      \\
      \footnotesize (a) \AlgGCIQL{} & \footnotesize (b) \AlgGCIVL{} \\[4pt]
      \includegraphics[width=0.48\linewidth]{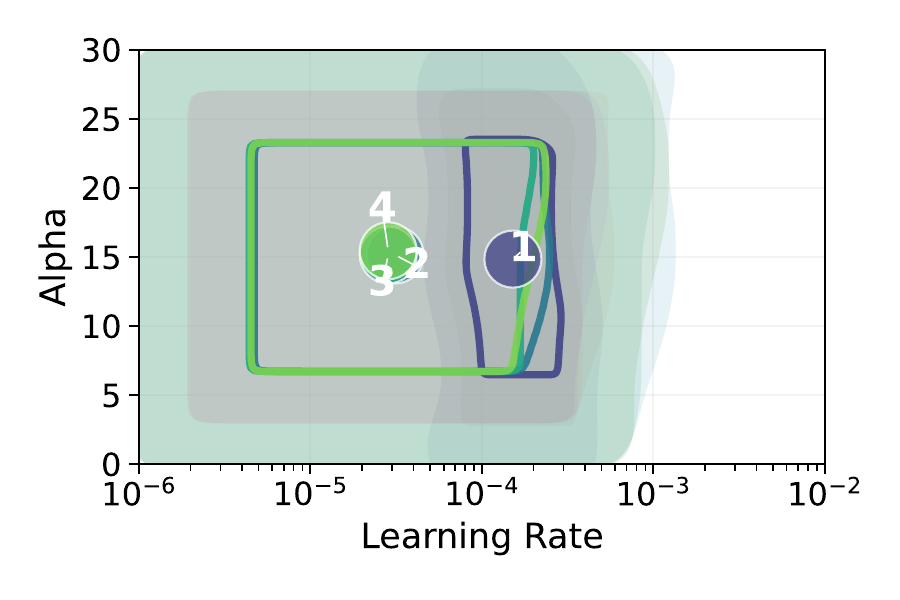}
      &
      \includegraphics[width=0.48\linewidth]{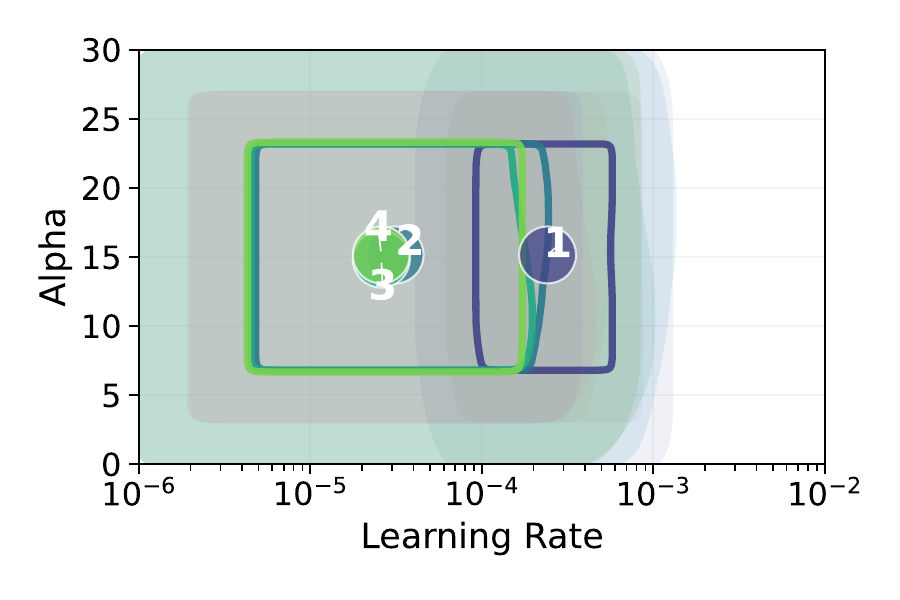}
      \\
      \footnotesize (c) \AlgCRL{} & \footnotesize (d) \AlgQRL{}
    \end{tabular}
  \end{minipage}\hfill
  \begin{minipage}[c]{0.32\textwidth}
    \caption{\textbf{\EnvAntMazeL{} (top-20\% regions).}
    On the longer-horizon task, landscapes generally become narrower. \AlgQRL{} keeps the broadest top-20\% region, while \AlgGCIVL{}'s region shrinks compared to \EnvAntMazeM{}.
    }
    \label{fig:mobility_antmaze_large_80}
  \end{minipage}
\end{minipage}

\subsection{\EnvCube{}}

\noindent\begin{minipage}{\textwidth}
  \captionsetup{type=figure, justification=raggedright, singlelinecheck=false, font=small}
  \centering
  \begin{minipage}[c]{0.66\textwidth}
    \centering
    \begin{tabular}{@{}cc@{}}
      \includegraphics[width=0.48\linewidth]{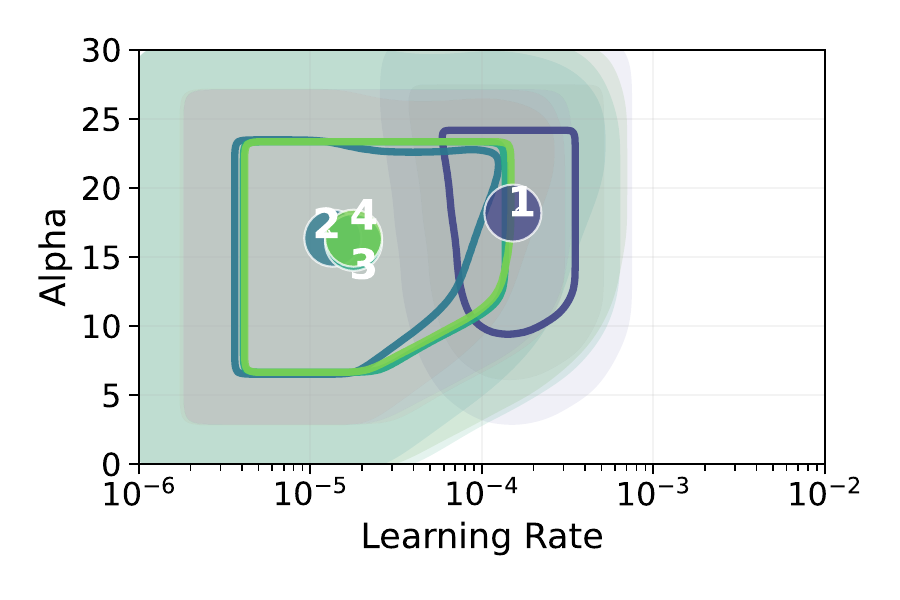}
      &
      \includegraphics[width=0.48\linewidth]{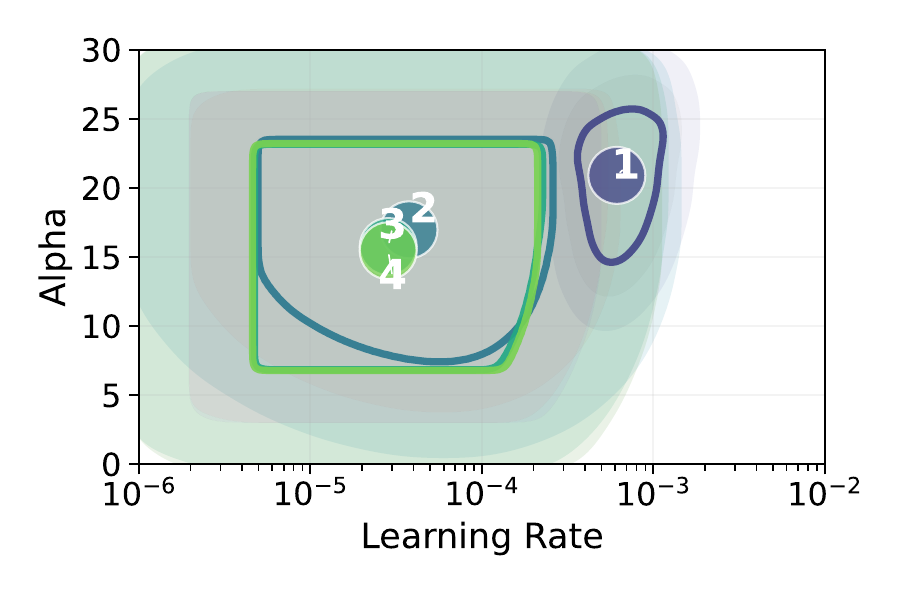}
      \\
      \footnotesize (a) \AlgGCIQL{} & \footnotesize (b) \AlgGCIVL{} \\[4pt]
      \includegraphics[width=0.48\linewidth]{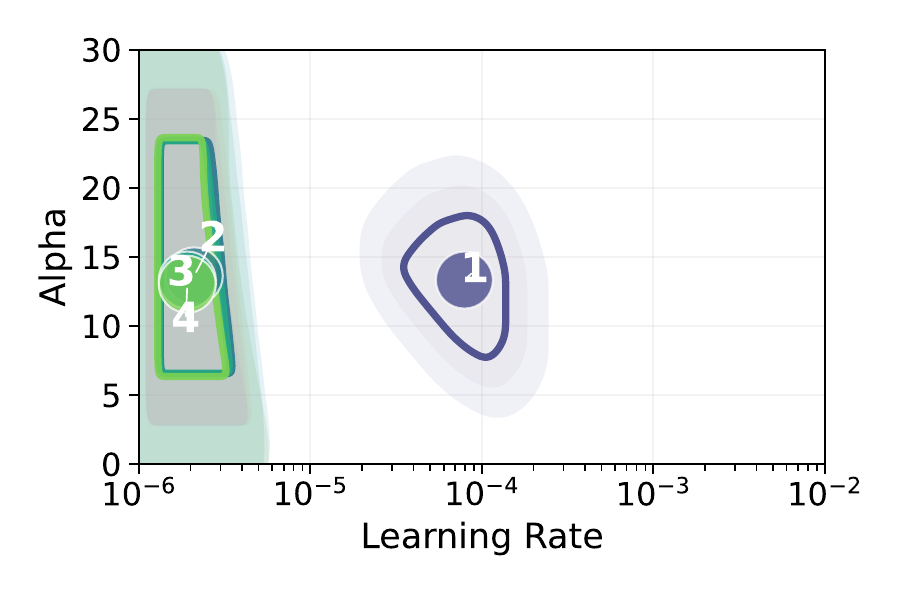}
      &
      \includegraphics[width=0.48\linewidth]{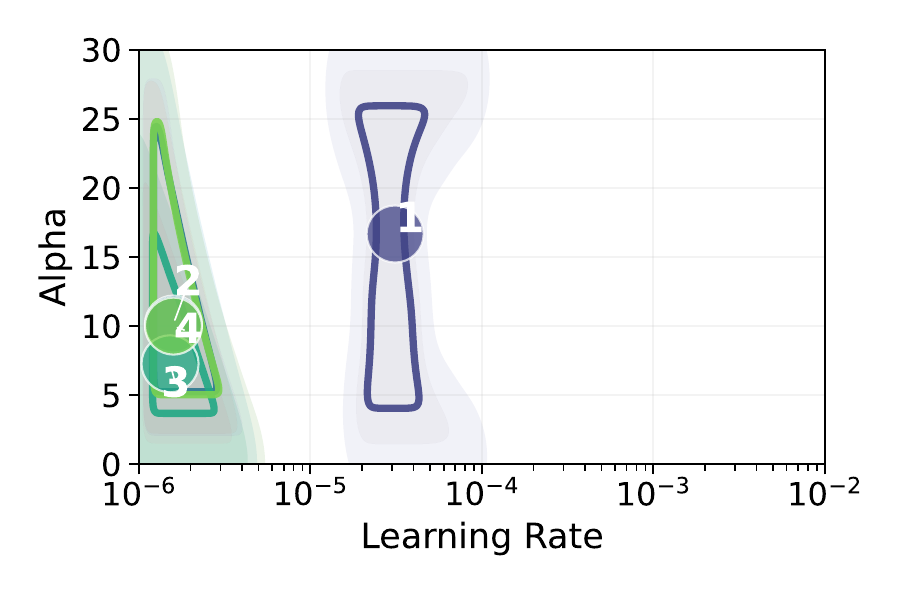}
      \\
      \footnotesize (c) \AlgCRL{} & \footnotesize (d) \AlgQRL{}
    \end{tabular}
  \end{minipage}\hfill
  \begin{minipage}[c]{0.32\textwidth}
    \caption{\textbf{\EnvCube{} (top-20\% regions).}
    \AlgGCIVL{} has a broad, stable top-20\% region and high success.
    }
    \label{fig:mobility_cube_80}
  \end{minipage}
\end{minipage}

\subsection{\EnvScene{}}

\noindent\begin{minipage}{\textwidth}
  \captionsetup{type=figure, justification=raggedright, singlelinecheck=false, font=small}
  \centering
  \begin{minipage}[c]{0.66\textwidth}
    \centering
    \begin{tabular}{@{}cc@{}}
      \includegraphics[width=0.48\linewidth]{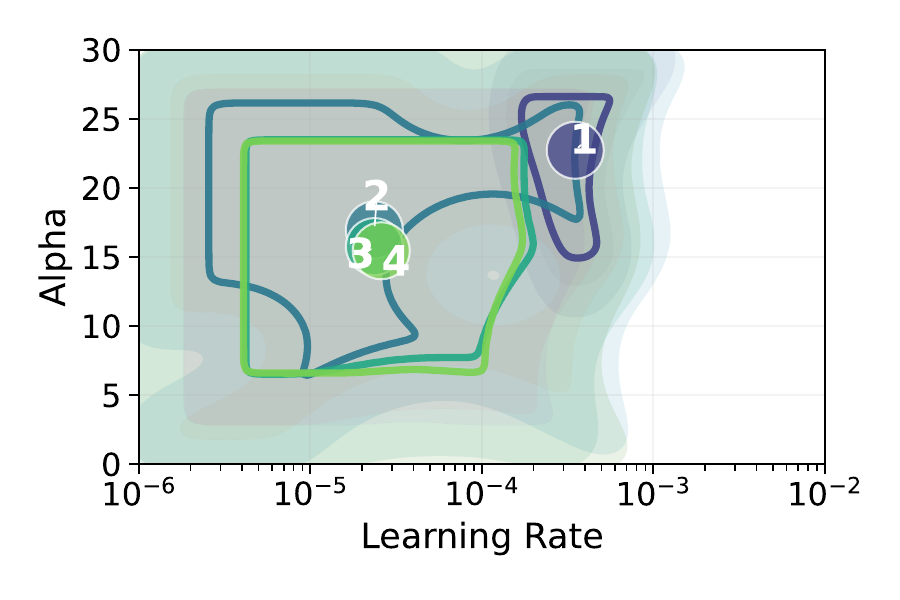}
      &
      \includegraphics[width=0.48\linewidth]{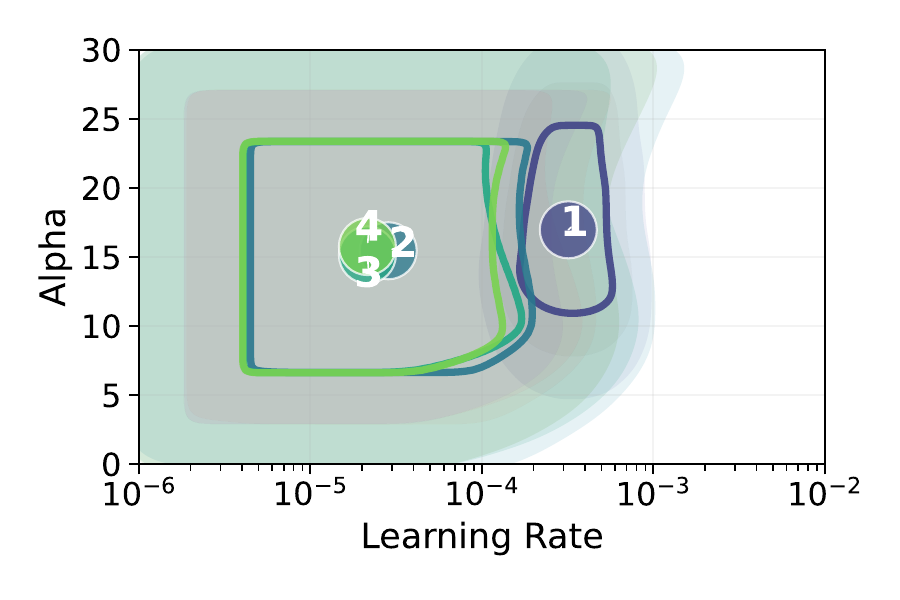}
      \\
      \footnotesize (a) \AlgGCIQL{} & \footnotesize (b) \AlgGCIVL{} \\[4pt]
      \includegraphics[width=0.48\linewidth]{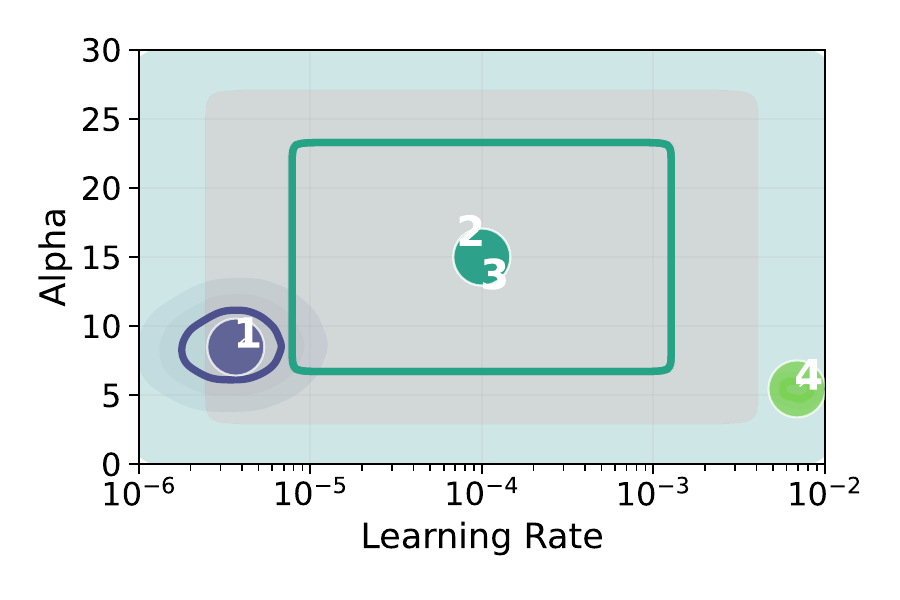}
      &
      \includegraphics[width=0.48\linewidth]{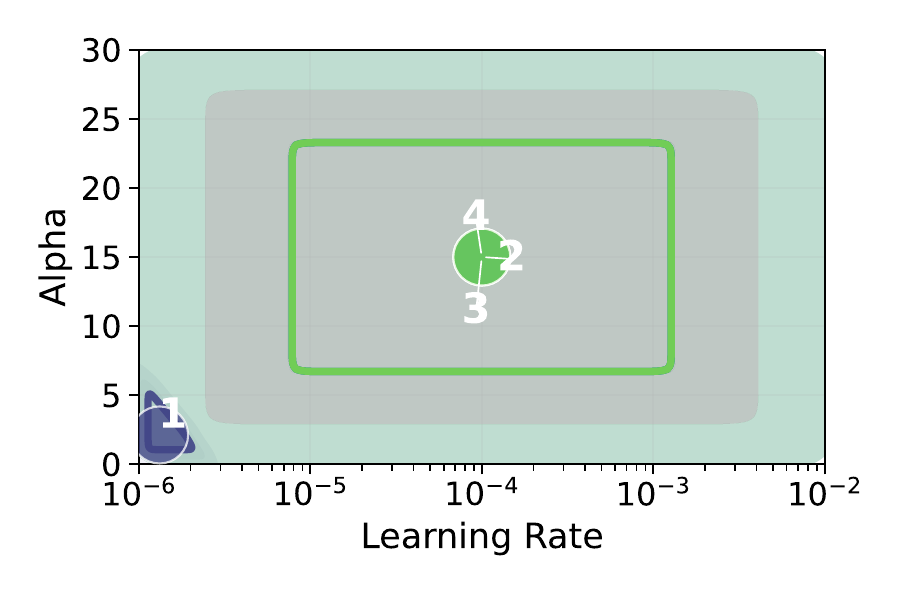}
      \\
      \footnotesize (c) \AlgCRL{} & \footnotesize (d) \AlgQRL{}
    \end{tabular}
  \end{minipage}\hfill
  \begin{minipage}[c]{0.32\textwidth}
    \caption{\textbf{\EnvScene{} (top-20\% regions).}
    \AlgGCIQL{} and \AlgGCIVL{} both reach high success with differently shaped landscapes. \AlgGCIQL{} reaches high success despite a fragmented top-20\% region.
    }
    \label{fig:mobility_scene_80}
  \end{minipage}
\end{minipage}

\subsection{Normalized advantage mobility}
\label{app:norm-adv-mobility}

\noindent\begin{minipage}{\textwidth}
  \captionsetup{type=figure, justification=raggedright, singlelinecheck=false, font=small}
  \centering
  \begin{minipage}[c]{0.66\textwidth}
    \centering
    \begin{tabular}{@{}cc@{}}
      \includegraphics[width=0.48\linewidth]{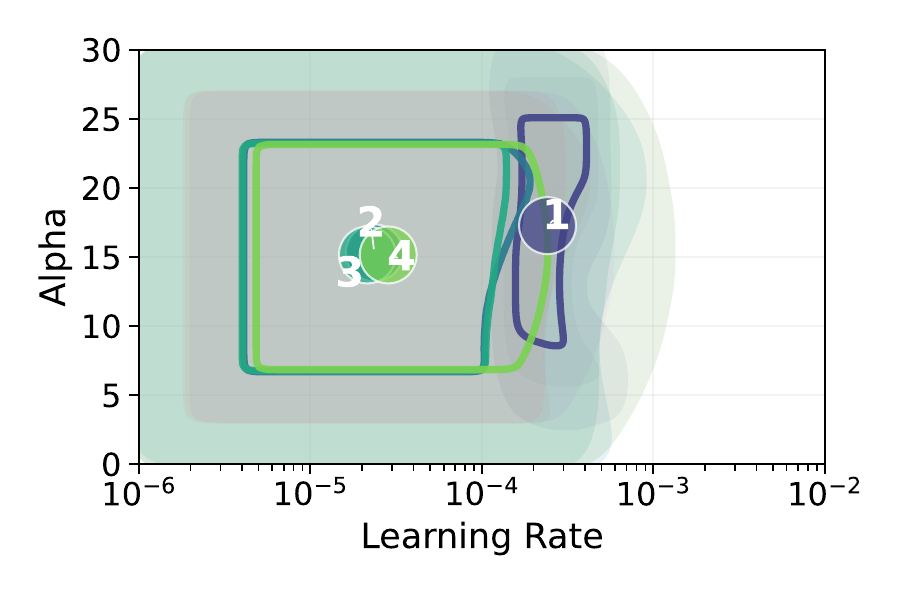}
      &
      \includegraphics[width=0.48\linewidth]{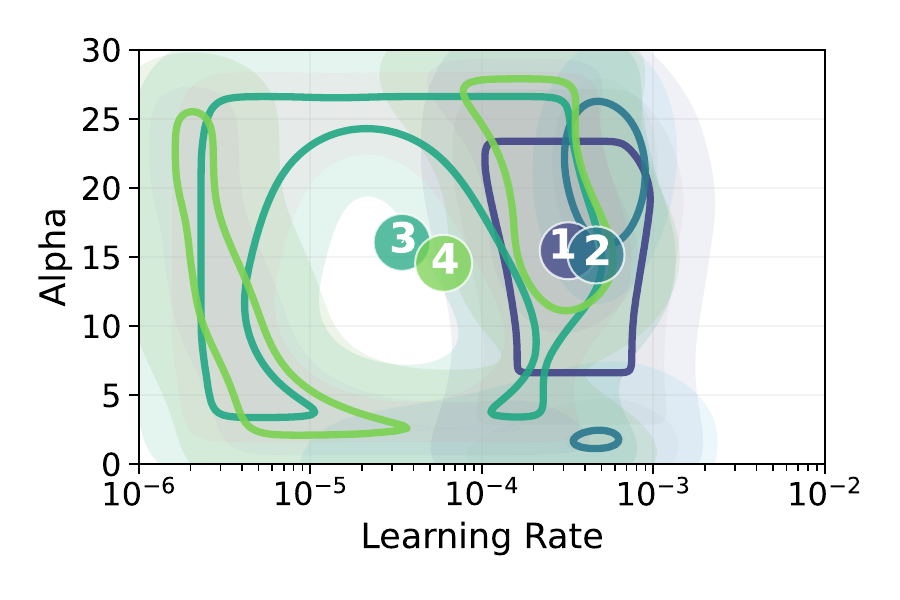}
      \\
      \footnotesize (a) \AlgCRL{} & \footnotesize (b) \AlgGCIQL{} \\[4pt]
      \includegraphics[width=0.48\linewidth]{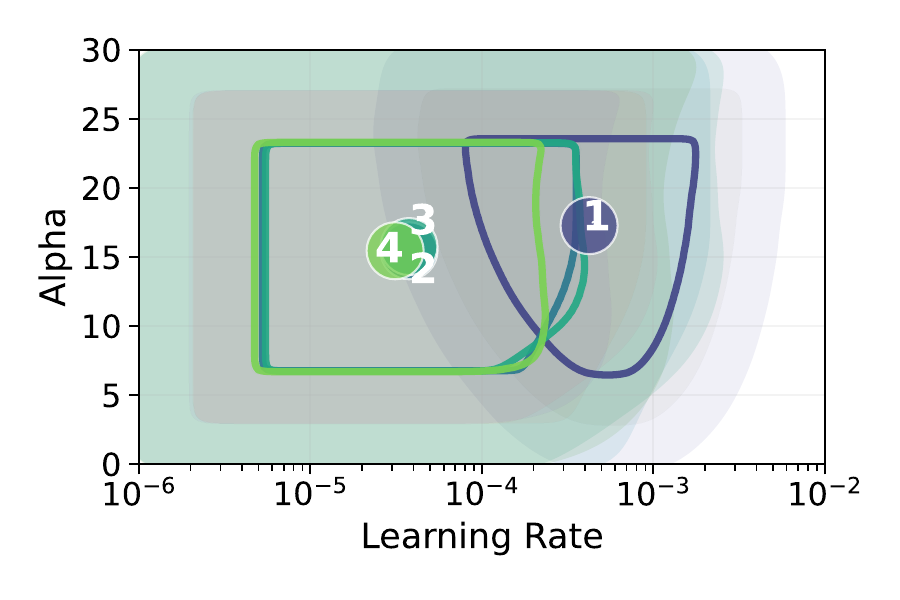}
      &
      \includegraphics[width=0.48\linewidth]{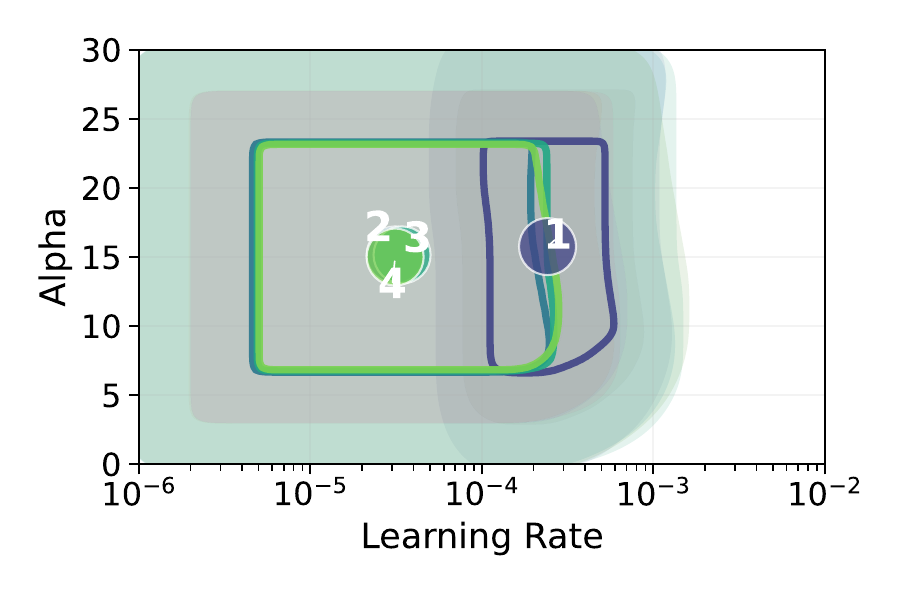}
      \\
      \footnotesize (c) \AlgGCIVL{} & \footnotesize (d) \AlgQRL{}
    \end{tabular}
  \end{minipage}\hfill
  \begin{minipage}[c]{0.32\textwidth}
    \caption{\textbf{\EnvAntMazeM{} (top-20\% regions, normalized advantage).}}
    \label{fig:mobility_norm_adv_antmaze_medium_80}
  \end{minipage}
\end{minipage}

\section{Additional Q-vs-Distance Plots}
\label{app:q-vs-distance}

\Cref{fig:qvd:antmaze-large} shows the same analysis as \cref{fig:qvd:antmaze-medium} for \EnvAntMazeL{}.
\Cref{fig:qvd:cube} and \cref{fig:qvd:scene} apply the same diagnostic to the manipulation environments.

\noindent\begin{minipage}{\textwidth}
  \captionsetup{type=figure}
  \centering

  \subcaptionbox{\AlgCRL{}\label{fig:qvd:antmaze-large:crl}}%
    {\includegraphics[width=0.45\textwidth]{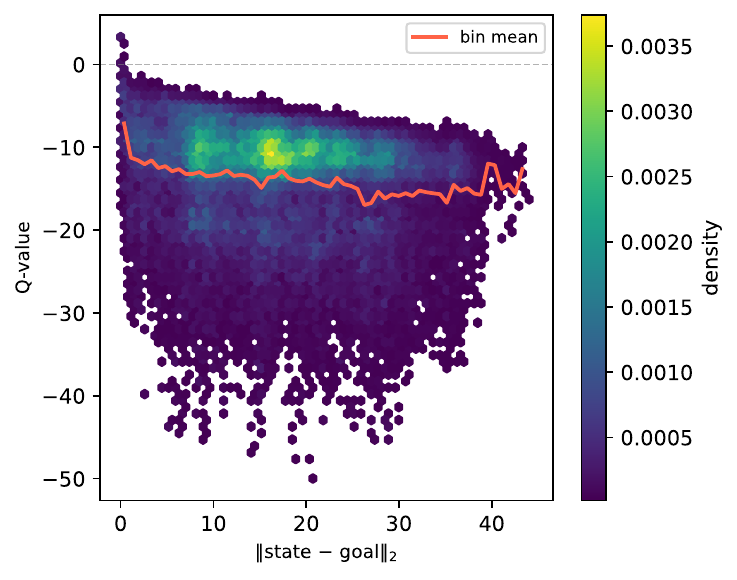}}
  \subcaptionbox{\AlgGCIQL{}\label{fig:qvd:antmaze-large:gciql}}%
    {\includegraphics[width=0.45\textwidth]{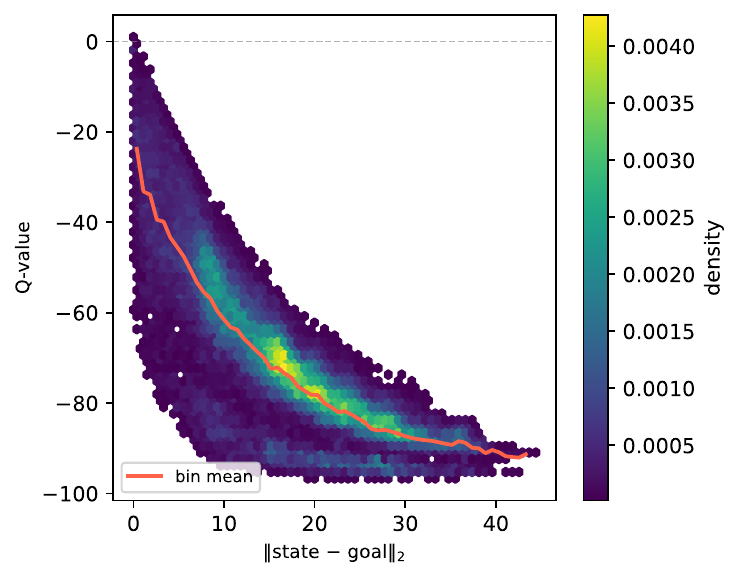}}

  \smallskip

  \subcaptionbox{\AlgGCIVL{}\label{fig:qvd:antmaze-large:gcivl}}%
    {\includegraphics[width=0.45\textwidth]{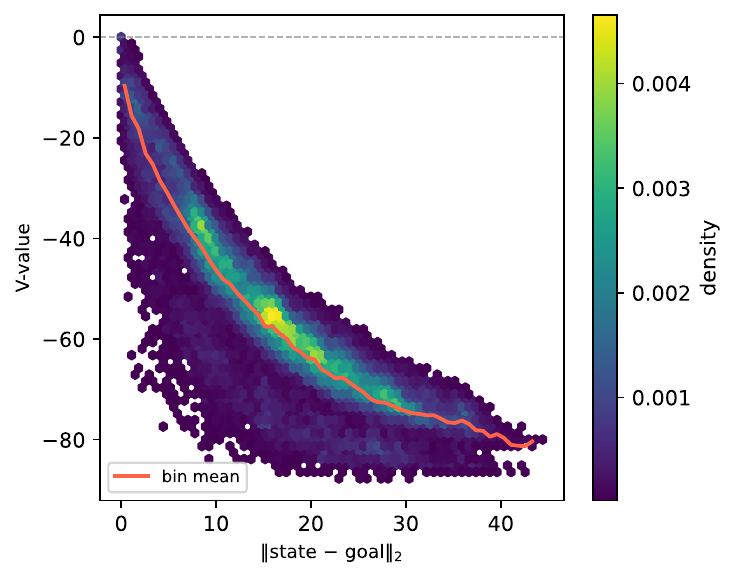}}
  \subcaptionbox{\AlgQRL{}\label{fig:qvd:antmaze-large:qrl}}%
    {\includegraphics[width=0.45\textwidth]{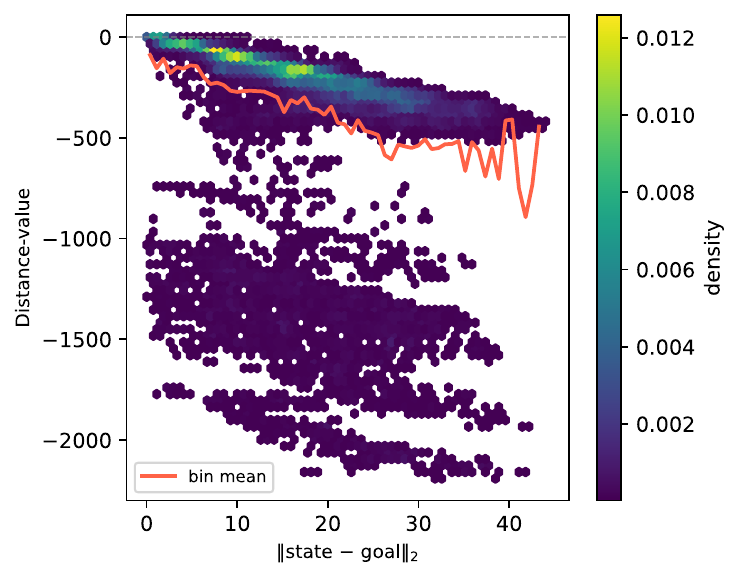}}

  \caption{\textbf{Score geometry across goal distance on \EnvAntMazeL{}.}
    Hexbin density from the $256 \times 256$ cross-product of states and goals from a validation batch, indexed by agent $xy$-distance to the goal.
    A distance-aligned score produces a negative trend.
    This diagnostic is coarse: Euclidean $xy$-distance ignores maze walls and the rest of the robot state.}
  \label{fig:qvd:antmaze-large}%
\end{minipage}

\noindent\begin{minipage}{\textwidth}
  \captionsetup{type=figure}
  \centering

  \subcaptionbox{\AlgCRL{}\label{fig:qvd:cube:crl}}%
    {\includegraphics[width=0.45\textwidth]{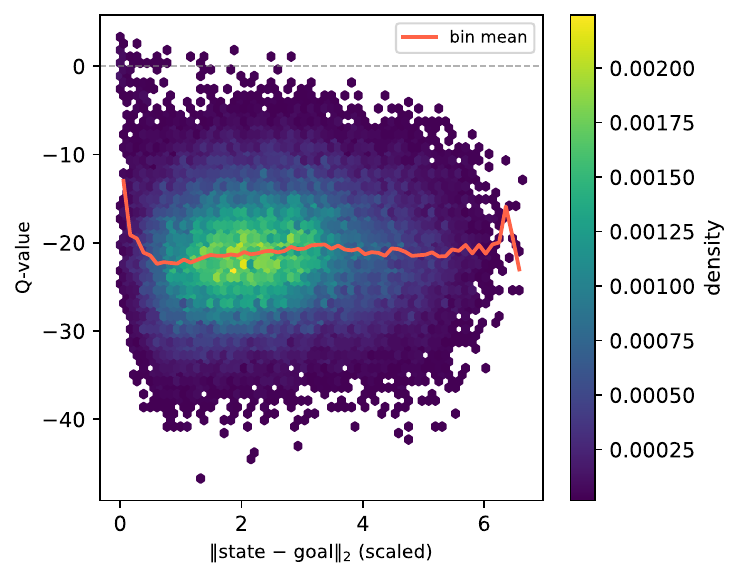}}
  \subcaptionbox{\AlgGCIQL{}\label{fig:qvd:cube:gciql}}%
    {\includegraphics[width=0.45\textwidth]{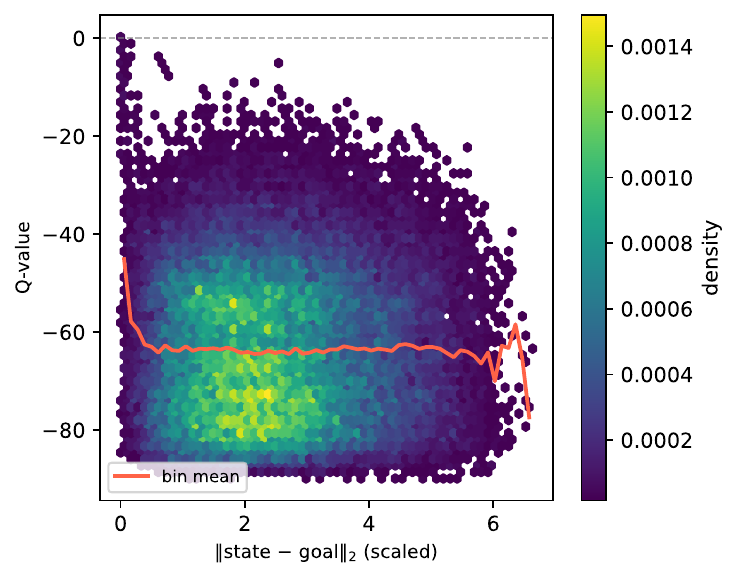}}

  \smallskip

  \subcaptionbox{\AlgGCIVL{}\label{fig:qvd:cube:gcivl}}%
    {\includegraphics[width=0.45\textwidth]{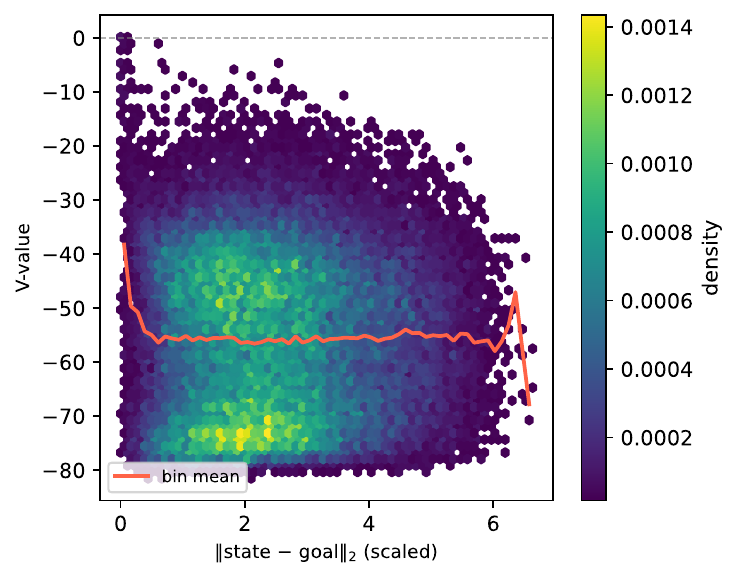}}
  \subcaptionbox{\AlgQRL{}\label{fig:qvd:cube:qrl}}%
    {\includegraphics[width=0.45\textwidth]{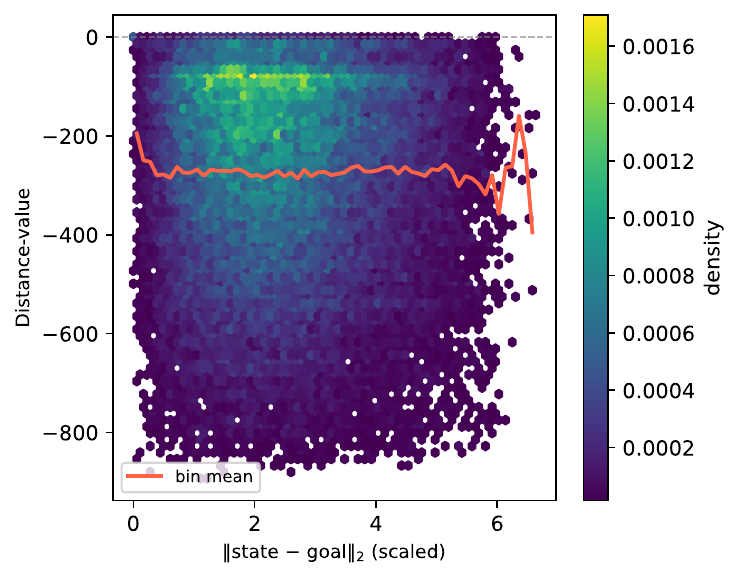}}

  \caption{\textbf{Score geometry across goal distance on \EnvCube{}.}
    Hexbin density from the $256 \times 256$ cross-product of states and goals from a validation batch, indexed by the Euclidean distance between the cube position and its goal position.
    A distance-aligned score produces a negative trend.
    The distance ignores the arm and gripper state.}
  \label{fig:qvd:cube}%
\end{minipage}

\noindent\begin{minipage}{\textwidth}
  \captionsetup{type=figure}
  \centering

  \subcaptionbox{\AlgCRL{}\label{fig:qvd:scene:crl}}%
    {\includegraphics[width=0.45\textwidth]{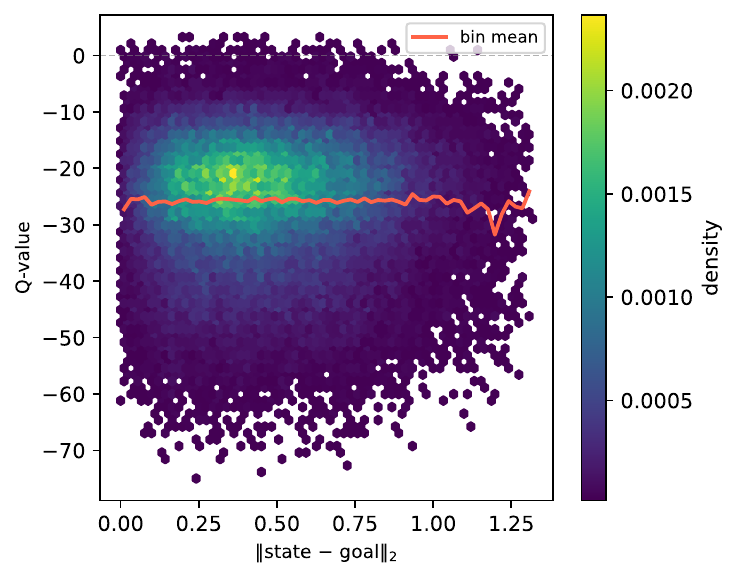}}
  \subcaptionbox{\AlgGCIQL{}\label{fig:qvd:scene:gciql}}%
    {\includegraphics[width=0.45\textwidth]{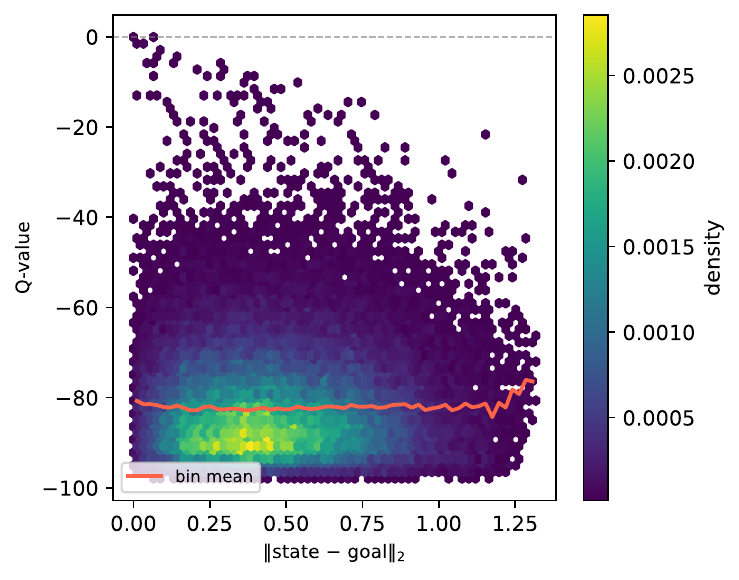}}

  \smallskip

  \subcaptionbox{\AlgGCIVL{}\label{fig:qvd:scene:gcivl}}%
    {\includegraphics[width=0.45\textwidth]{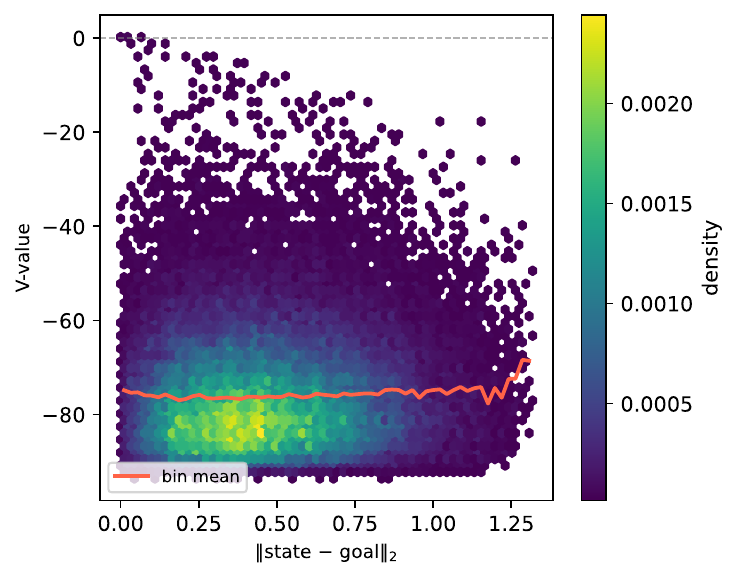}}
  \subcaptionbox{\AlgQRL{}\label{fig:qvd:scene:qrl}}%
    {\includegraphics[width=0.45\textwidth]{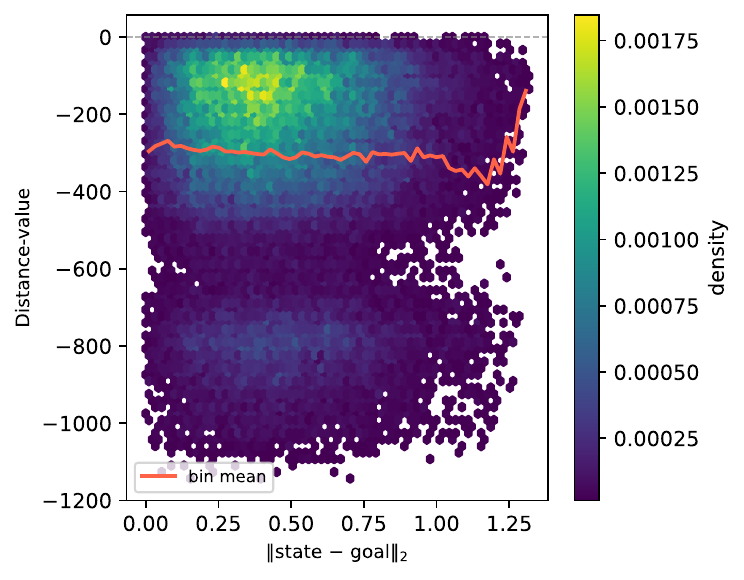}}

  \caption{\textbf{Score geometry across goal distance on \EnvScene{}.}
    Hexbin density from the $256 \times 256$ cross-product of states and goals from a validation batch, indexed by the Euclidean distance between the object positions and their goal positions.
    A distance-aligned score produces a negative trend.
    The distance ignores the arm and gripper state.}
  \label{fig:qvd:scene}%
\end{minipage}

\section{Robustness and Sensitivity Analyses}
\label{app:robustness}

This appendix collects checks that support the interpretation in \cref{sec:experiments}. 
They test whether the conclusions depend on one diagnostic table, one configuration sample, one clipping threshold, one seed
realization, or one advantage scaling choice. 
The checks do not change the main claim: downstream success, landscape breadth, and policy-extraction diagnostics measure different parts of the offline \GCRL{} training pipeline.

\paragraph{Notation used throughout this appendix.}
We follow the symbols introduced in \cref{sec:method}: \(C\subseteq\Lambda\) is the sampled configuration set with \(|C|=64\), each
\(\lambda\in C\) carries a learning rate \(\eta(\lambda)\) and \AWR{} temperature \(\alpha(\lambda)\), and \(\bar S_t(\lambda)\) is the interquartile
mean (IQM) of binary success over seeds at phase \(t\in\{1,2,3,4\}\). The four phases use a
common seed set with \(K=5\) seeds; we write \(S_t(\lambda,k)\) for the
single-seed success and \(T=4\) for the final phase. Each
(algorithm, environment) cell is fixed implicitly; we add explicit indices
only when comparisons cross algorithms or environments.

\AWR{} weights and effective sample size follow the form in \cref{sec:method}:
\[
  w_\alpha(s,a,g)
  \;=\;
  \min\bigl\{\exp(\alpha\,A_{\mathrm{alg}}(s,a,g)),\,w_{\max}\bigr\},
  \qquad w_{\max}=100,
\]
\[
  \ESS
  \;=\;
  \frac{\bigl(\sum_{i=1}^{B}w_i\bigr)^2}{B\,\sum_{i=1}^{B}w_i^{\,2}},
\]
where \(B\) is the diagnostic batch size (\(B=256\) throughout this
appendix). Where multiple independent diagnostic batches are drawn, we index
them by \(m=1,\ldots,M\) with \(M=10\).

\subsection{Diagnostic-success scatter plots}
\label{app:diagnostic_scatter}

\begin{figure*}[htb]
\centering
\includegraphics[width=\textwidth]{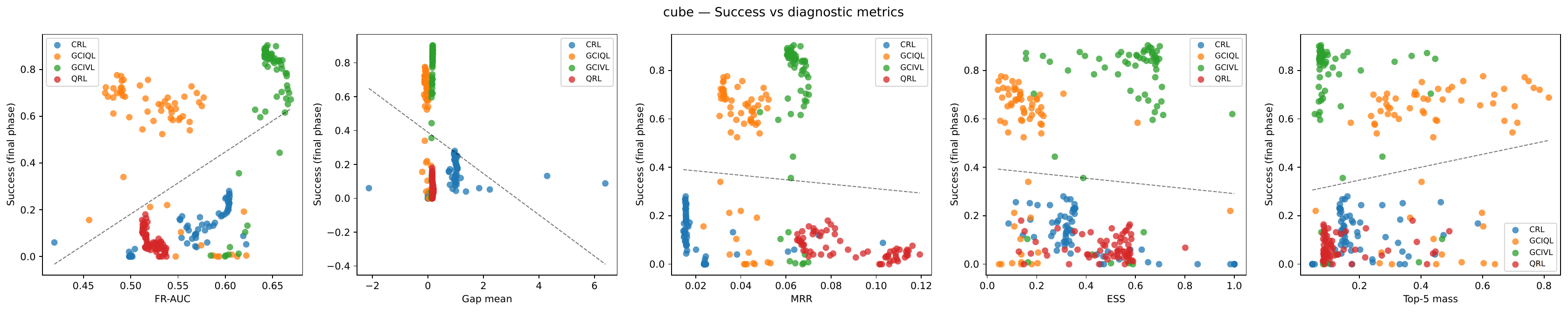}%
\caption{\textbf{\EnvCube{}: success versus diagnostics at the configuration level.}
Each point is a configuration, colored by method. \FRAUC{} has the strongest
pooled correlation with success, but the method clusters show that no
diagnostic alone explains downstream performance. Dashed lines show pooled
linear trends and should not be interpreted as method-internal causal
effects.}
\label{fig:diagnostic_scatter_cube}
\end{figure*}

\begin{figure*}[htb]
\centering
\includegraphics[width=\textwidth]{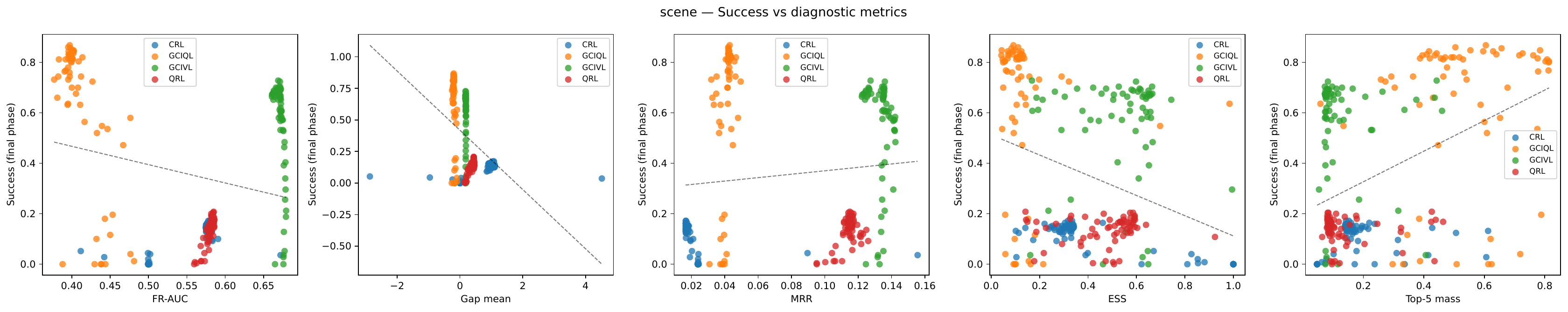}%
\caption{\textbf{\EnvScene{}: success versus diagnostics at the configuration level.}
\AlgGCIQL{} achieves high success despite poor future-association diagnostics,
while \AlgQRL{} has strong ranking metrics but weak downstream success.
Goal-ranking diagnostics therefore describe the learned signal and predict
policy success poorly. Dashed lines show pooled linear trends across methods
and carry no causal meaning within a method.}
\label{fig:diagnostic_scatter_scene}
\end{figure*}

\paragraph{What we compute.}
For each configuration \(\lambda\in C\) at the final phase \(T\), we
aggregate each diagnostic into a single per-configuration value and pair it
with the IQM success \(\bar S_T(\lambda)\). For a given diagnostic
\(D\in\{\FRAUC,\,\mathrm{Gap},\,\MRR,\,\ESS,\,\text{\TopFivePctMass{}}\}\) and its
per-batch value \(D(\lambda,k,T,m)\) on the \(m\)th fixed-seed validation
batch for seed \(k\) and phase \(T\), the per-configuration aggregate is
\[
  \widetilde D(\lambda)
  \;=\;
  \frac{1}{K}\sum_{k=1}^{K}\;\frac{1}{M}\sum_{m=1}^{M}\,
  D(\lambda,k,T,m).
\]
The horizontal axes of
\cref{fig:diagnostic_scatter_cube,fig:diagnostic_scatter_scene}
correspond to the five diagnostic columns of \cref{tab:compact_diagnostics};
the vertical axis is \(\bar S_T(\lambda)\). Each point is one configuration,
colored by algorithm. Pooled trends are summarized with the Pearson
correlation coefficient \(r\) on the pooled
\((\widetilde D(\lambda),\bar S_T(\lambda))\) sample across algorithms.

The configuration-level trends match the main manipulation result.
On \EnvCube, the strongest pooled trend is \FRAUC{} with success
\((r=0.46)\), while gap, \MRR, \ESS, and \TopFivePctMass{} are weak or negative.
On \EnvScene, \TopFivePctMass{} \((r=0.43)\) and \ESS{} \((r=-0.33)\) are more
correlated with success than \FRAUC, while gap is negatively correlated
\((r=-0.41)\). This supports the main interpretation: diagnostics describe
the learned signal and extractor behavior, and downstream evaluation
remains necessary.

\subsection{Label sensitivity: filtering non-distance-reducing transitions}
\label{app:anchor_filter}

\FRAUC{}, Gap, and \MRR{} treat the future goal sampled from the same
trajectory as the positive label for each transition. This label can be
misleading: for transitions whose logged action does not move toward
that future goal, treating it as a positive is questionable. We recompute
all three diagnostics after filtering to transitions whose logged action
reduces the task-relevant distance (agent \(xy\) in AntMaze, object position
in \EnvCube{}/\EnvScene{}) to the future goal, using the same 10 fixed validation batches, final-phase
checkpoints, and \(255\) random candidate goals as the main
diagnostic table.

\begin{table}[ht]
\centering
\small
\setlength{\tabcolsep}{4pt}
\resizebox{\textwidth}{!}{%
\begin{tabular}{llrrrrrrr}
\toprule
Environment & Algorithm & Retained & \multicolumn{2}{c}{\FRAUC{}} & \multicolumn{2}{c}{Gap} & \multicolumn{2}{c}{\MRR{}} \\
 & & frac. & all & kept & all & kept & all & kept \\
\midrule
\EnvAntMazeM{} & \AlgCRL{}   & 0.629 & 0.530 & 0.530 & 0.169  & 0.177 & 0.022 & 0.022 \\
               & \AlgGCIQL{} & 0.629 & 0.471 & 0.492 & -0.102 & -0.044 & 0.045 & 0.045 \\
               & \AlgGCIVL{} & 0.629 & 0.564 & 0.606 & 0.143  & 0.254 & 0.078 & 0.090 \\
               & \AlgQRL{}   & 0.629 & 0.546 & 0.640 & 0.358  & 0.693 & 0.072 & 0.094 \\
\midrule
\EnvAntMazeL{} & \AlgCRL{}   & 0.671 & 0.527 & 0.532 & 0.174  & 0.191 & 0.019 & 0.019 \\
               & \AlgGCIQL{} & 0.671 & 0.472 & 0.495 & -0.101 & -0.044 & 0.042 & 0.044 \\
               & \AlgGCIVL{} & 0.671 & 0.575 & 0.626 & 0.136  & 0.249 & 0.084 & 0.096 \\
               & \AlgQRL{}   & 0.671 & 0.582 & 0.660 & 0.680  & 1.072 & 0.080 & 0.100 \\
\midrule
\EnvCube{}     & \AlgCRL{}   & 0.311 & 0.562 & 0.570 & 0.875  & 0.937 & 0.022 & 0.023 \\
               & \AlgGCIQL{} & 0.311 & 0.531 & 0.516 & -0.039 & -0.146 & 0.040 & 0.043 \\
               & \AlgGCIVL{} & 0.311 & 0.644 & 0.663 & 0.160  & 0.185 & 0.064 & 0.072 \\
               & \AlgQRL{}   & 0.311 & 0.523 & 0.547 & 0.167  & 0.253 & 0.092 & 0.116 \\
\midrule
\EnvScene{}    & \AlgCRL{}   & 0.260 & 0.555 & 0.565 & 0.695  & 0.789 & 0.022 & 0.022 \\
               & \AlgGCIQL{} & 0.260 & 0.411 & 0.406 & -0.192 & -0.174 & 0.041 & 0.041 \\
               & \AlgGCIVL{} & 0.260 & 0.672 & 0.718 & 0.185  & 0.341 & 0.134 & 0.154 \\
               & \AlgQRL{}   & 0.260 & 0.580 & 0.632 & 0.379  & 0.972 & 0.115 & 0.147 \\
\bottomrule
\end{tabular}%
}
\vspace{0.5em}
\caption{\textbf{Diagnostics after filtering non-distance-reducing transitions.}
``all'' is the unfiltered value from \cref{tab:compact_diagnostics}; ``kept''
uses only the retained transitions. Retained fraction is per environment
(transition selection depends only on the data).}
\label{tab:anchor_filter}
\end{table}

Retained fraction is much lower in the manipulation tasks (\(26\)--\(31\,\%\))
than in AntMaze (\(63\)--\(67\,\%\)): in \EnvCube{}/\EnvScene{}, most
single-step logged actions do not reduce the object-position distance to a
sampled future goal from the same trajectory. Single-step positional
progress is therefore a weak proxy for task progress in manipulation.

Of the \(48\) environment/algorithm/metric cells in \cref{tab:anchor_filter},
filtering raises \(39\), leaves \(6\) unchanged, and lowers \(3\) (all
\AlgGCIQL{}: \EnvCube{} \FRAUC{} and Gap, \EnvScene{} \FRAUC{}). The
qualitative separations in
\cref{tab:compact_diagnostics} mostly survive: of the \(12\)
(environment, metric) rankings across the four algorithms, \(5\) are
identical before and after filtering, and in \(11\) of \(12\) the
lowest-scoring algorithm is unchanged. The one exception is \EnvCube{}
\FRAUC{}, where \AlgQRL{} (\(0.523\)) and \AlgGCIQL{} (\(0.531\)) are within
\(0.01\) unfiltered and swap under filtering (\(0.547\) vs.\ \(0.516\)).
Every remaining rank change swaps two adjacent positions, mostly between
closely valued algorithms (e.g.\ \AlgGCIVL{} and \AlgQRL{} trade the top
\FRAUC{}/\MRR{} rank on \EnvAntMazeM{}). The largest change is \EnvScene{}
Gap, where \AlgQRL{} overtakes \AlgCRL{} for the top rank
(\(0.379\to0.972\) vs.\ \(0.695\to0.789\)); \AlgQRL{}'s diagnostics
benefit most from filtering overall.

\subsection{Full phase-resolved AntMaze concentration table}
\label{app:antmaze_ess_full}

The main text reports the final-phase \AWR{} concentration pattern on
\EnvAntMazeM. \Cref{tab:antmaze_ess_full} reports the full phase-resolved
table.

\paragraph{What we compute.}
For each configuration \(\lambda\in C\), seed \(k\), and phase \(t\) on
\EnvAntMazeM, we read the per-step advantage array used by the \AWR{}
actor at the final evaluation step of phase \(t\) from that run's training
log. Call this vector \(A_t(\lambda,k)\in\mathbb{R}^{B}\) (with \(B=256\)). The
clipped \AWR{} weights are
\[
  w_t(\lambda,k)_i
  \;=\;
  \min\!\bigl\{\exp\bigl(\alpha(\lambda)\,A_t(\lambda,k)_i\bigr),\,w_{\max}\bigr\},
\]
and the per-(configuration, seed, phase) effective sample size is
\[
  \ESS_t(\lambda,k)
  \;=\;
  \frac{\bigl(\sum_{i=1}^{B} w_t(\lambda,k)_i\bigr)^2}{B\,\sum_{i=1}^{B} w_t(\lambda,k)_i^{\,2}}.
\]
Each cell at \((a,t)\) of \cref{tab:antmaze_ess_full} reports the mean and
standard deviation of \(\ESS_t(\lambda,k)\) over the \(64\times 5=320\)
\((\lambda,k)\) pairs.

For the correlation column we form the per-phase normalized return
\(\widetilde R_t(\lambda,k)=R_t(\lambda,k)/\max_{\lambda'}R_t(\lambda',k)\),
where \(R_t(\lambda,k)=\Rdist\) is the mean normalized goal-distance return
of that cell, and report the Pearson correlation coefficient
\[
  r^{(a,t)}
  \;=\;
  \mathrm{corr}\bigl(\,
    \{\ESS_t(\lambda,k)\}_{\lambda,k},\;
    \{\widetilde R_t(\lambda,k)\}_{\lambda,k}
  \bigr)
\]
with its two-sided \(p\)-value.

\begin{table}[ht]
  \centering
  \small
  \setlength{\tabcolsep}{4pt}
  \resizebox{\textwidth}{!}{%
  \begin{tabular}{lccc ccc ccc ccc}
    \toprule
    & \multicolumn{3}{c}{Phase 1} & \multicolumn{3}{c}{Phase 2} & \multicolumn{3}{c}{Phase 3} & \multicolumn{3}{c}{Phase 4} \\
    \cmidrule(lr){2-4} \cmidrule(lr){5-7} \cmidrule(lr){8-10} \cmidrule(lr){11-13}
    Algorithm & \ESS{} & Corr. & \(p\) & \ESS{} & Corr. & \(p\) & \ESS{} & Corr. & \(p\) & \ESS{} & Corr. & \(p\) \\
    \midrule
    \AlgCRL{}   & \(0.38\pm0.34\) & \(-0.43\) & \(<0.01\) & \(0.37\pm0.31\) & \(-0.89\) & \(<0.01\) & \(0.30\pm0.24\) & \(-0.83\) & \(<0.01\) & \(0.32\pm0.26\) & \(-0.86\) & \(<0.01\) \\
    \AlgGCIQL{} & \(0.16\pm0.14\) &  \(0.27\) & \(<0.01\) & \(0.15\pm0.11\) &  \(0.10\) & \(0.08\)   & \(0.15\pm0.12\) &  \(0.06\) & \(0.30\)   & \(0.14\pm0.12\) &  \(0.10\) & \(0.07\) \\
    \AlgGCIVL{} & \(0.46\pm0.13\) &  \(0.14\) & \(0.01\)  & \(0.49\pm0.14\) &  \(0.26\) & \(<0.01\)  & \(0.49\pm0.14\) & \(-0.01\) & \(0.89\)   & \(0.49\pm0.14\) &  \(0.02\) & \(0.74\) \\
    \AlgQRL{}   & \(0.55\pm0.10\) &  \(0.14\) & \(0.01\)  & \(0.57\pm0.11\) &  \(0.00\) & \(0.97\)   & \(0.57\pm0.11\) &  \(0.05\) & \(0.39\)   & \(0.57\pm0.11\) &  \(0.05\) & \(0.38\) \\
    \bottomrule
  \end{tabular}%
  }
  \vspace{0.5em}
  \caption{\textbf{Phase-resolved \AWR{} concentration on \EnvAntMazeM.}
  \ESS{} measures the concentration of \AWR{} weights. Lower \ESS{} is associated
  with higher return for \AlgCRL{} but not for \AlgGCIQL{}, consistent with
  concentration being useful only when the selected samples are reliable.}
  \label{tab:antmaze_ess_full}
\end{table}

\subsection{Lagged and partial concentration--success relationship}
\label{app:lagged_ess_antmaze}

\paragraph{What we compute.}
The previous subsection correlates \ESS{} with the normalized return
\(\widetilde R_t\) within the same phase. Here we ask two related
questions on \EnvAntMazeM, using success \(S_t\) as the outcome: does
concentration at phase \(t\) predict success one phase later, and does the
same-phase \ESS{}--success relationship hold after controlling for the
learning rate? Using
the per-configuration \ESS{} values \(\ESS_t(\lambda,k)\) defined above
(\(64\times5=320\) \((\lambda,k)\) pairs per phase), we compute the lagged
Pearson correlation
\[
  r^{\mathrm{lag}}_t
  = \mathrm{corr}\bigl(\{\ESS_t(\lambda,k)\},\{S_{t+1}(\lambda,k)\}\bigr),
  \qquad t\in\{1,2,3\},
\]
and the partial Pearson correlation of \(\ESS_t\) and \(S_t\) controlling
for \(\log\eta(\lambda)\), evaluated at the final phase \(t=4\).

\begin{table}[ht]
\centering
\small
\setlength{\tabcolsep}{5pt}
\begin{tabular}{lrrrr}
\toprule
& \multicolumn{3}{c}{Lagged \(r(\ESS_t, S_{t+1})\)} & Partial \(r\) \\
\cmidrule(lr){2-4}
Algorithm & \(t{=}1\) & \(t{=}2\) & \(t{=}3\) & phase 4, controls \(\log\eta\) \\
\midrule
\AlgCRL{}   & \(-0.785\) & \(-0.791\) & \(-0.762\) & \(-0.805\) \\
\AlgGCIQL{} & \(0.324\)  & \(0.213\)  & \(0.118\)  & \(0.448\)  \\
\AlgGCIVL{} & \(0.273\)  & \(0.113\)  & \(0.101\)  & \(0.123\)  \\
\AlgQRL{}   & \(0.094\)  & \(0.042\)  & \(0.049\)  & \(-0.047\) \\
\bottomrule
\end{tabular}
\vspace{0.5em}
\caption{\textbf{Lagged and learning-rate-adjusted \ESS{}--success
correlations on \EnvAntMazeM.} Left columns: Pearson correlation between
\ESS{} at phase \(t\) and success at phase \(t{+}1\), over the \(320\)
\((\lambda,k)\) pairs. Right column: partial correlation of \ESS{} and
success at the final phase, controlling for \(\log\eta(\lambda)\).}
\label{tab:lagged_ess_antmaze}
\end{table}

\AlgCRL{} shows a strong, stable negative lagged relationship
(\(r\approx-0.78\)) from phase 1 onward, and the same-phase relationship
remains strong after controlling for the learning rate at the final phase
(\(-0.805\)): concentrated \AWR{} weights in one phase are followed by
higher success in the next. \AlgGCIVL{} and \AlgQRL{} show only weak
relationships (\(|r|<0.3\)). \AlgGCIQL{} is weakly positive in the
lagged correlations (\(0.12\)--\(0.32\)) but moderately positive once the
learning rate is controlled for (\(0.448\)), the opposite sign to
\AlgCRL{}. These
are descriptive correlations over configurations and seeds; they do not
support a causal claim that concentration drives later success.

\subsection{Seed variance}
\label{app:seed_variance}

\paragraph{What we compute.}
For each configuration \(\lambda\in C\) at the final phase \(T\) we compute
the unbiased seed variance
\[
  \sigma^2(\lambda)
  \;=\;
  \frac{1}{K-1}
  \sum_{k=1}^{K}\bigl(S_T(\lambda,k)-\bar S_T(\lambda)\bigr)^2.
\]
The per-(algorithm, environment) summary is the mean over configurations,
\[
  \nu^{(a,e)}
  \;=\;
  \frac{1}{|C|}\sum_{\lambda\in C}\sigma^2(\lambda).
\]
\Cref{tab:seed_variance_summary} reports \(\min_a\nu^{(a,e)}\),
\(\max_a\nu^{(a,e)}\), and the mean over the four algorithms,
\(\frac{1}{4}\sum_a\nu^{(a,e)}\), for each environment \(e\). Seed variance
is small across the board, with mean values between \(0.001\) and
\(0.006\); this indicates that the main landscape-breadth differences are
not driven by extreme seed instability.

\begin{table}[ht]
\centering
\small
\setlength{\tabcolsep}{5pt}
\begin{tabular}{lrrr}
\toprule
Environment & Min mean seed var. & Max mean seed var. & Mean seed var. \\
\midrule
\EnvAntMazeL{} & 0.0015 & 0.0034 & 0.0025 \\
\EnvAntMazeM{} & 0.0035 & 0.0041 & 0.0038 \\
\EnvCube{}     & 0.0018 & 0.0056 & 0.0043 \\
\EnvScene{}    & 0.0012 & 0.0035 & 0.0026 \\
\bottomrule
\end{tabular}
\vspace{0.5em}
\caption{\textbf{Range of final-phase seed variance across method--environment pairs.}
Values are computed from per-configuration success over seeds. The small
scale indicates that landscape differences are not dominated by seed noise.}
\label{tab:seed_variance_summary}
\end{table}

Seed variance is not always negatively related to performance.
For \AlgCRL{} and \AlgQRL{}, more successful configurations are often more
variable. For \AlgGCIVL{} on \EnvAntMazeM{} and \EnvScene, higher-success
configurations are less variable. This reinforces the need to report both
performance and landscape breadth.

\subsection{Phase mobility}
\label{app:phase_mobility}

The top-configuration mobility plots in the main text show that useful
regions move across phases. We quantify this in two ways: (i) Jaccard overlap
between top-10\,\% configuration sets in adjacent phases, and (ii) centroid
drift of those sets in normalized search coordinates.

\paragraph{What we compute.}
Per (algorithm, environment) and phase \(t\), define the top-10\,\% set
\[
  T_t
  \;=\;
  \operatorname*{arg\,top-}_{\lambda\in C}{}^{q}\;\bar S_t(\lambda)
  \quad\text{with}\quad
  q=\lceil 0.10\cdot|C|\rceil.
\]
The Jaccard overlap between adjacent phases is
\[
  J_t
  \;=\;
  \frac{|T_t\cap T_{t+1}|}{|T_t\cup T_{t+1}|}.
\]
For centroid drift, each configuration is mapped to Sobol-like coordinates
\((u_\eta(\lambda),u_\alpha(\lambda))\in[0,1]^2\) by min--max scaling
\[
  u_\eta(\lambda)
  =
  \frac{\log\eta(\lambda)-\min_{\lambda'}\log\eta(\lambda')}{\max_{\lambda'}\log\eta(\lambda')-\min_{\lambda'}\log\eta(\lambda')},
\]
\[
  u_\alpha(\lambda)
  =
  \frac{\alpha(\lambda)-\min_{\lambda'}\alpha(\lambda')}{\max_{\lambda'}\alpha(\lambda')-\min_{\lambda'}\alpha(\lambda')},
\]
and the centroid drift between phases \(t\) and \(t+1\) is
\[
  d_t
  \;=\;
  \Bigl\|\,
    \tfrac{1}{|T_t|}\sum_{\lambda\in T_t}\!\bigl(u_\eta(\lambda),u_\alpha(\lambda)\bigr)
    \;-\;
    \tfrac{1}{|T_{t+1}|}\sum_{\lambda\in T_{t+1}}\!\bigl(u_\eta(\lambda),u_\alpha(\lambda)\bigr)
  \,\Bigr\|_{2}.
\]
\Cref{tab:phase_mobility} reports the mean of \(J_t\) and \(d_t\) over the
three phase transitions \((1{\to}2,\,2{\to}3,\,3{\to}4)\), then the average
across the four algorithms per environment.

\begin{table}[ht]
\centering
\small
\setlength{\tabcolsep}{4pt}
\resizebox{\textwidth}{!}{%
\begin{tabular}{lrrcr}
\toprule
Environment & Mean Jaccard overlap & Mean centroid drift & Clears null? & Spearman \\
\midrule
\EnvAntMazeL{} & 0.075 & 0.178 & No  & 0.495 \\
\EnvAntMazeM{} & 0.241 & 0.210 & Yes & 0.57  \\
\EnvCube{}     & 0.169 & 0.208 & Yes & 0.587 \\
\EnvScene{}    & 0.173 & 0.235 & Yes & 0.516 \\
\bottomrule
\end{tabular}%
}s
\vspace{0.5em}
\caption{\textbf{Phase mobility of top-10\% configurations.}
Low Jaccard overlap indicates substantial turnover in the best configuration set across phases. 
Centroid drift measures movement of the top-set location in normalized search coordinates. 
The ``Clears null?'' column compares the observed overlap against a 200{,}000-permutation null on the mean top-7 Jaccard overlap over the \(12\) (algorithm, adjacent-phase) cells per environment (with \(|C|=64\), the top-10\,\% set contains 7 configurations). 
The null mean is \(0.062\) with 95\,\% band \([0.026,0.103]\), identical across environments because each has \(12\) cells; ``clears null'' means the observed overlap lies above that band. The Spearman column is the mean over the same \(12\) cells of the rank correlation of all \(64\) configurations between adjacent phases.}
\label{tab:phase_mobility}
\end{table}

The low overlaps show that phase-resolved analysis adds information beyond the final landscape: the identity of high-performing configurations changes during training. \EnvAntMazeL{} is the most mobile
environment by this measure, while \EnvAntMazeM{} is relatively more stable.

Three of the four environments show top-set overlap clearly above the permutation chance level. 
The exception is \EnvAntMazeL{}, whose observed overlap falls inside the null band (\(p \approx 0.22\)); its mean adjacent-phase Spearman correlation (\(0.495\), positive in \(11\) of \(12\) cells) is nevertheless clearly nonzero. 
The two quantities answer different questions there: the identity of the top set cannot be distinguished from chance-level turnover, while the quality ordering of the full configuration set is preserved.

\subsection{Subsampling stability of landscape statistics}
\label{app:subsample_stability}

\paragraph{What we compute.}
Let \(\mu^{(a)}=\frac{1}{|C|}\sum_{\lambda\in C}\bar S_T(\lambda)\) denote
the full-grid mean success of algorithm \(a\) at the final phase. For each
ordered pair of algorithms \((a_1,a_2)\) the full-grid ranking declares
\(a_1\succ a_2\) iff \(\mu^{(a_1)}>\mu^{(a_2)}\).

For each \(q\in\{32,48\}\) we draw \(N_{\mathrm{sub}}=1000\) random
subsamples \(C^{(b)}\subset C\) of size \(q\) without replacement (a common
index set across algorithms and a fixed random seed). The
subsample mean is
\(\hat\mu^{(a,b)}=\frac{1}{q}\sum_{\lambda\in C^{(b)}}\bar S_T(\lambda)\),
and we report the agreement rate
\[
  \mathrm{Agreement}_{q}(a_1,a_2)
  \;=\;
  \frac{1}{N_{\mathrm{sub}}}
  \sum_{b=1}^{N_{\mathrm{sub}}}
  \mathbf{1}\!\left[\,\hat\mu^{(a_1,b)}>\hat\mu^{(a_2,b)}\,\right]\!\cdot 100\%
\]
for each close pair (those whose full-grid means are close enough that some
subsamples reverse the ordering). Pairs that agree in \(100\,\%\) of
subsamples are omitted.

\begin{table}[ht]
\centering
\small
\setlength{\tabcolsep}{4pt}
\begin{tabular}{llrr}
\toprule
Environment & Close pair & Agreement at 32 configs & Agreement at 48 configs \\
\midrule
\EnvAntMazeL{} & \AlgQRL{} \(>\) \AlgCRL{}   & 74.8\% & 86.1\% \\
\EnvAntMazeM{} & \AlgCRL{} \(>\) \AlgQRL{}   & 53.8\% & 55.6\% \\
\bottomrule
\end{tabular}
\vspace{0.5em}
\caption{\textbf{Subsampling stability for close method pairs.}
Most pairwise method orderings are stable in 100\,\% of subsamples. The
table shows the few close comparisons where ordering is genuinely
uncertain.}
\label{tab:subsample_stability_close_pairs}
\end{table}

The result is consistent with the main landscape table. Method orderings
that are far apart remain stable under subsampling. The unstable cases are
close in the full sweep, so the instability reflects genuine near-ties.

\subsection{Sensitivity of \AWR{} concentration to the clipping threshold}
\label{app:wmax_sensitivity}

\paragraph{What we compute.}
Let \(A^{+}_{T}(\lambda,k,m)\in\mathbb{R}^{B}\) denote the diagonal of the
\(B\times B\) cross-goal advantage matrix for cell \((\lambda,k,T)\) and
validation batch \(m\) (\(B=256\), \(M=10\) batches).
For each configuration we
concatenate diagonals across seeds and batches into a single vector
\(A^{+}(\lambda)\in\mathbb{R}^{B\,K\,M}\), then for each clipping ceiling
\(w_{\max}\in\{20,50,100,200\}\) recompute weights and concentration:
\[
  w(\lambda;w_{\max})_i
  =
  \min\!\bigl\{\exp\bigl(\alpha(\lambda)\,A^{+}(\lambda)_i\bigr),\,w_{\max}\bigr\},
\]
\[
  \ESS(\lambda;w_{\max})
  =
  \frac{\bigl(\sum_i w(\lambda;w_{\max})_i\bigr)^2}{|A^{+}(\lambda)|\,\sum_i w(\lambda;w_{\max})_i^{\,2}},
\]
\[
  \mathrm{Top}\text{-}5(\lambda;w_{\max})
  =
  \frac{\sum_{i\in\mathrm{top}\,5\%}w(\lambda;w_{\max})_i}{\sum_i w(\lambda;w_{\max})_i}.
\]
Each cell at \((e,a,w_{\max})\) is the mean of these per-configuration
values over \(C\). \Cref{tab:wmax_sensitivity} shows the two endpoints
\(w_{\max}\in\{20,200\}\) side by side; the values at \(w_{\max}=50,100\)
interpolate between them.

\begin{table}[ht]
\centering
\small
\setlength{\tabcolsep}{4pt}
\resizebox{\textwidth}{!}{%
\begin{tabular}{llrrrr}
\toprule
Environment & Algorithm & \ESS{} \(w_{\max}=20\) & \ESS{} \(w_{\max}=200\) & \TopFivePctMass{} \(w_{\max}=20\) & \TopFivePctMass{} \(w_{\max}=200\) \\
\midrule
\EnvAntMazeM{} & \AlgCRL{}   & 0.311 & 0.244 & 0.195 & 0.274 \\
               & \AlgGCIQL{} & 0.177 & 0.121 & 0.356 & 0.525 \\
               & \AlgGCIVL{} & 0.553 & 0.433 & 0.109 & 0.172 \\
               & \AlgQRL{}   & 0.601 & 0.504 & 0.093 & 0.123 \\
\midrule
\EnvAntMazeL{} & \AlgCRL{}   & 0.315 & 0.261 & 0.191 & 0.261 \\
               & \AlgGCIQL{} & 0.184 & 0.119 & 0.357 & 0.541 \\
               & \AlgGCIVL{} & 0.546 & 0.423 & 0.111 & 0.175 \\
               & \AlgQRL{}   & 0.655 & 0.568 & 0.084 & 0.108 \\
\midrule
\EnvCube{}     & \AlgCRL{}   & 0.342 & 0.300 & 0.186 & 0.224 \\
               & \AlgGCIQL{} & 0.209 & 0.127 & 0.325 & 0.516 \\
               & \AlgGCIVL{} & 0.655 & 0.512 & 0.094 & 0.154 \\
               & \AlgQRL{}   & 0.557 & 0.417 & 0.108 & 0.174 \\
\midrule
\EnvScene{}    & \AlgCRL{}   & 0.332 & 0.299 & 0.209 & 0.253 \\
               & \AlgGCIQL{} & 0.179 & 0.100 & 0.393 & 0.612 \\
               & \AlgGCIVL{} & 0.638 & 0.465 & 0.098 & 0.167 \\
               & \AlgQRL{}   & 0.561 & 0.421 & 0.108 & 0.174 \\
\bottomrule
\end{tabular}%
}
\vspace{0.5em}
\caption{\textbf{Sensitivity of \AWR{} concentration to \(w_{\max}\).}
We recompute \ESS{} and \TopFivePctMass{} for
\(w_{\max}\in\{20,50,100,200\}\) without retraining. The table shows the
two endpoints. The qualitative ordering is stable: \AlgGCIQL{} remains the
most concentrated method, while \AlgGCIVL{} and \AlgQRL{} remain more
diffuse.}
\label{tab:wmax_sensitivity}
\end{table}

The main qualitative result is stable across clipping thresholds.
Increasing \(w_{\max}\) reduces \ESS{} and increases \TopFivePctMass{}, as
expected, but \AlgGCIQL{} remains the most concentrated method in every
environment. \AlgGCIQL{}'s concentration pattern therefore holds across
clipping values.

\subsection{Advantage normalization sensitivity}
\label{app:adv_norm_sensitivity}

\paragraph{What we compute.}
Starting from the same per-configuration diagonal advantage vector
\(A^{+}(\lambda)\) used in the \(w_{\max}\) sweep, we apply four
normalization maps \(\phi\) and recompute concentration. The temperature
used to exponentiate is the trained \(\alpha(\lambda)\) for raw advantages
and \(1\) for the three rescaling schemes (which absorb the magnitude):
\[
  \phi_{\mathrm{raw}}\bigl(A^{+}\bigr)_i = A^{+}_i,\;\alpha_{\mathrm{eff}}=\alpha(\lambda),
  \quad
  \phi_{z}\bigl(A^{+}\bigr)_i = \frac{A^{+}_i-\mu}{\sigma},\;\alpha_{\mathrm{eff}}=1,
\]
\[
  \phi_{\mathrm{rank}}\bigl(A^{+}\bigr)_i
  = \frac{\operatorname{rank}_i(A^{+})}{|A^{+}|-1}\in[0,1],\;\alpha_{\mathrm{eff}}=1,
\]
\[
  \phi_{\mathrm{clipstd}}\bigl(A^{+}\bigr)_i
  =
  \frac{\operatorname{clip}(A^{+}_i,\mu-3\sigma,\mu+3\sigma)-\tilde\mu}{\tilde\sigma},
  \;\alpha_{\mathrm{eff}}=1,
\]
where \(\mu,\sigma\) are the mean and standard deviation of \(A^{+}\),
and \(\tilde\mu,\tilde\sigma\) are computed after the clip. Weights and
\ESS{} are then computed exactly as in \cref{app:wmax_sensitivity}, with
the trained \(w_{\max}=100\) clip. The cell at \((e,a,\phi)\) is the mean
\ESS{} over configurations.

For non-raw schemes this is a diagnostic-only probe: changing \(\phi\)
defines a different policy extractor and would, under retraining, also
change the trained advantages.

\begin{table}[ht]
\centering
\small
\setlength{\tabcolsep}{4pt}
\resizebox{\textwidth}{!}{%
\begin{tabular}{llrrrr}
\toprule
Environment & Algorithm & Raw \ESS{} & Z-score \ESS{} & Clipped-std \ESS{} & Rank \ESS{} \\
\midrule
\EnvAntMazeM{} & \AlgCRL{}   & 0.260 & 0.196 & 0.342 & 0.924 \\
\EnvAntMazeM{} & \AlgGCIQL{} & 0.133 & 0.208 & 0.289 & 0.924 \\
\EnvAntMazeM{} & \AlgGCIVL{} & 0.465 & 0.210 & 0.259 & 0.924 \\
\EnvAntMazeM{} & \AlgQRL{}   & 0.532 & 0.278 & 0.537 & 0.924 \\
\midrule
\EnvCube{}     & \AlgCRL{}   & 0.311 & 0.353 & 0.414 & 0.924 \\
\EnvCube{}     & \AlgGCIQL{} & 0.145 & 0.221 & 0.225 & 0.924 \\
\EnvCube{}     & \AlgGCIVL{} & 0.553 & 0.250 & 0.230 & 0.924 \\
\EnvCube{}     & \AlgQRL{}   & 0.454 & 0.089 & 0.211 & 0.924 \\
\midrule
\EnvScene{}    & \AlgCRL{}   & 0.308 & 0.361 & 0.432 & 0.924 \\
\EnvScene{}    & \AlgGCIQL{} & 0.118 & 0.296 & 0.241 & 0.924 \\
\EnvScene{}    & \AlgGCIVL{} & 0.512 & 0.165 & 0.184 & 0.924 \\
\EnvScene{}    & \AlgQRL{}   & 0.458 & 0.139 & 0.123 & 0.924 \\
\bottomrule
\end{tabular}%
}
\vspace{0.5em}
\caption{\textbf{Diagnostic-only advantage normalization sensitivity.}
Raw advantages reproduce the main concentration pattern: \AlgGCIQL{} has
low \ESS{}. Z-scoring and clipped standardization change absolute
concentration and sometimes change the ordering. Rank normalization
collapses the weight distribution by construction. Advantage
normalization therefore defines a different extractor.}
\label{tab:adv_norm_sensitivity}
\end{table}

This analysis supports a narrower claim than the \(w_{\max}\) sweep. The
raw \AlgGCIQL{} concentration pattern is strong, and it is robust to
clipping-ceiling changes. However, normalization changes the effective
extractor and can alter concentration. Future work should test
rank-normalized or standardized \AWR{} with retraining.

\subsection{Sensitivity over the sampled search space}
\label{app:sampled_space_sensitivity}

The main text interprets sensitivity only as a descriptive check. Here we
include both a log-linear \(R^2\) decomposition and random-forest
partial-dependence ranges. The random forest only provides a
nonlinear descriptive view of the sampled landscape; its numbers describe
sensitivity within our search ranges.

\paragraph{What we compute --- \(R^2\) decomposition.}
For each (algorithm, environment) at the final phase \(T\) we form the data
matrix \(\bigl\{\bigl(\log\eta(\lambda),\log\alpha(\lambda),\bar S_T(\lambda)\bigr)\bigr\}_{\lambda\in C}\)
and standardize the predictors:
\(\ell_\lambda=z(\log\eta(\lambda))\), \(a_\lambda=z(\log\alpha(\lambda))\).
We then fit two ordinary-least-squares regressions and read off the
coefficients of determination
\[
  R^2_{\eta} = R^2(\bar S_T\sim\ell),
  \qquad
  R^2_{\alpha} = R^2(\bar S_T\sim a).
\]
Bootstrap margins are obtained by resampling configurations with
replacement \(N_{\mathrm{boot}}=2000\) times and reporting the half-width of
the \([2.5\,\%,\,97.5\,\%]\) percentile interval as the \(\pm\) value next
to each \(R^2\).

\paragraph{What we compute --- partial-dependence ranges.}
For each (algorithm, environment) we fit a
random-forest regressor with \(500\) trees, \(\sqrt{p}\) features
per split, bootstrap sampling, and a fixed random seed, on the same
\((\log\eta,\log\alpha)\to\bar S_T\) data. Let \(f\) be the fitted
regressor. The partial-dependence function in \(\eta\) is
\[
  \mathrm{PD}_\eta(v)
  =
  \frac{1}{|C|}
  \sum_{\lambda\in C}
  f\bigl(v,\,\log\alpha(\lambda)\bigr),
\]
evaluated on a uniform grid of \(40\) points spanning the observed range of
\(\log\eta\); the partial-dependence range is
\[
  \mathrm{PD\,range}_\eta
  =
  \max_v\mathrm{PD}_\eta(v)-\min_v\mathrm{PD}_\eta(v),
\]
and likewise for \(\alpha\). These ranges are point estimates from the
random forest --- no bootstrap is reported on them.

\begin{table}[ht]
\centering
\small
\setlength{\tabcolsep}{3.5pt}
\resizebox{\textwidth}{!}{%
\begin{tabular}{llrrrr}
\toprule
Environment & Algorithm & \(R^2_{\eta}\) & \(R^2_{\alpha}\) & PD range \(\eta\) & PD range \(\alpha\) \\
\midrule
\EnvAntMazeM{} & \AlgCRL{}   & \(0.50 \pm 0.12\) & \(0.03 \pm 0.07\) & 0.702 & 0.160 \\
               & \AlgGCIQL{} & \(0.65 \pm 0.15\) & \(0.07 \pm 0.12\) & 0.431 & 0.241 \\
               & \AlgGCIVL{} & \(0.40 \pm 0.17\) & \(0.04 \pm 0.11\) & 0.504 & 0.139 \\
               & \AlgQRL{}   & \(0.54 \pm 0.12\) & \(0.02 \pm 0.06\) & 0.614 & 0.128 \\
\midrule
\EnvAntMazeL{} & \AlgCRL{}   & \(0.66 \pm 0.11\) & \(0.02 \pm 0.05\) & 0.431 & 0.107 \\
               & \AlgGCIQL{} & \(0.61 \pm 0.12\) & \(0.02 \pm 0.05\) & 0.188 & 0.032 \\
               & \AlgGCIVL{} & \(0.27 \pm 0.19\) & \(0.03 \pm 0.07\) & 0.100 & 0.038 \\
               & \AlgQRL{}   & \(0.68 \pm 0.09\) & \(0.01 \pm 0.03\) & 0.421 & 0.062 \\
\midrule
\EnvCube{}     & \AlgCRL{}   & \(0.87 \pm 0.07\) & \(0.02 \pm 0.05\) & 0.226 & 0.030 \\
               & \AlgGCIQL{} & \(0.67 \pm 0.12\) & \(0.03 \pm 0.07\) & 0.655 & 0.164 \\
               & \AlgGCIVL{} & \(0.55 \pm 0.13\) & \(0.03 \pm 0.07\) & 0.753 & 0.158 \\
               & \AlgQRL{}   & \(0.79 \pm 0.08\) & \(0.02 \pm 0.05\) & 0.120 & 0.019 \\
\midrule
\EnvScene{}    & \AlgCRL{}   & \(0.68 \pm 0.12\) & \(0.01 \pm 0.04\) & 0.138 & 0.017 \\
               & \AlgGCIQL{} & \(0.66 \pm 0.09\) & \(0.01 \pm 0.04\) & 0.736 & 0.102 \\
               & \AlgGCIVL{} & \(0.67 \pm 0.08\) & \(0.03 \pm 0.06\) & 0.582 & 0.079 \\
               & \AlgQRL{}   & \(0.64 \pm 0.12\) & \(0.01 \pm 0.03\) & 0.149 & 0.023 \\
\bottomrule
\end{tabular}%
}
\vspace{0.5em}
\caption{\textbf{Sensitivity within the sampled search space.}
The \(R^2\) columns come from a log-linear model of final-phase success,
with bootstrap margins of error. The partial-dependence ranges come from a
random-forest model over \(\log\eta\) and \(\log\alpha\). Because the
learning-rate and \(\alpha\) ranges differ, these values describe sensitivity
within our sampled search space only.}
\label{tab:sensitivity_summary}
\end{table}

Across environments, the sampled-space variation along learning rate is
larger than the variation along \AWR{} temperature. This rules out the
simplest explanation that the observed \EnvCube{} and \EnvScene{} behavior
is only an \AWR{} temperature issue. Claims about intrinsic importance would
require comparable search ranges.

\subsection{Landscape visualization details}
\label{app:visualization_details}

For landscape figures, the main text shows top-20\,\% regions because they
visualize the location and mobility of useful configurations without
collapsing to only a few points. The mobility plots in \cref{app:mobility_plots}
use the same top-20\,\% threshold for consistency with the main text.
These plots are relative to each method's own performance. They should therefore
be read together with the success and landscape-breadth table in
\cref{tab:final_landscape_ci}. Independent Gaussian process regressor
surfaces are used only for visualization. All reported landscape masses,
correlations, and diagnostic tables are computed from raw configuration
evaluations.

\subsection{\texorpdfstring{TD horizon ($n$-step returns)}{TD horizon (n-step returns)}}
\label{app:nstep}

\AlgGCIVL{} and \AlgGCIQL{} train their critics with one-step TD targets.
To test whether the landscape shapes depend on this choice, we replace the
target with an \(n\)-step return, which sums \(n\) logged rewards before
bootstrapping, \(\sum_{k=0}^{n-1}\gamma^k r_{t+k} + \gamma^n V(s_{t+n}, g)\).
Longer horizons propagate value faster but add variance and
behavior-policy bias. Everything else is unchanged: the same \(64\)
\((\eta,\alpha)\) configurations, the same \AWR{} extractor, and the same
phase schedule. We test \(n\in\{3,5\}\) for \AlgGCIVL{} on \EnvCube{} and
\EnvAntMazeL{}, and \(n=3\) for \AlgGCIQL{} on \EnvAntMazeM{}, with three
seeds per configuration, and compare against the main-study runs
(\(n=1\), five seeds) at the final phase. Because the \(n>1\) runs have
three seeds, their per-configuration success is a plain seed mean, whereas
\(n=1\) uses the five-seed IQM \(\bar S_T\) (\cref{sec:method}); small
differences, particularly in \(\rho_{0.9}\), should be read with this in mind.

\paragraph{\AlgGCIVL{}.}
On \EnvCube{}, longer horizons lower both peak success
(\(0.91\to0.71\to0.66\)) and mean success (\(0.70\to0.49\to0.45\);
\cref{tab:nstep_gcivl_breadth}). The broad moderate-success region
persists: \(86\)--\(89\,\%\) of configurations reach success \(\ge0.25\) at
every horizon. The near-peak region, however, shrinks sharply
(\(\rho_{0.9}\): \(0.56\to0.09\to0.20\)). On \EnvAntMazeL{}, all horizons
stay near the floor (peak \(\le0.23\)), so this environment is
uninformative about breadth.

\begin{table}[ht]
\centering
\small
\begin{tabular}{lrrrrr}
\toprule
Environment & \(n\) & Peak & Mean & \(\rho_{\mathrm{abs}}(0.25)\) & \(\rho_{0.9}\) \\
\midrule
\EnvCube{}     & 1 & 0.91 & 0.70 & 0.88 & 0.56 \\
\EnvCube{}     & 3 & 0.71 & 0.49 & 0.89 & 0.09 \\
\EnvCube{}     & 5 & 0.66 & 0.45 & 0.86 & 0.20 \\
\midrule
\EnvAntMazeL{} & 1 & 0.17 & 0.09 & 0.00 & 0.02 \\
\EnvAntMazeL{} & 3 & 0.17 & 0.08 & 0.00 & 0.02 \\
\EnvAntMazeL{} & 5 & 0.23 & 0.14 & 0.00 & 0.03 \\
\bottomrule
\end{tabular}
\vspace{0.5em}
\caption{\textbf{\AlgGCIVL{} breadth under \(n\)-step TD targets.} Final
phase, matched \((\eta,\alpha)\) configurations. The \(n{=}3,5\) experimental conditions use
the mean over three seeds; the \(n{=}1\) baseline uses the five-seed IQM.}
\label{tab:nstep_gcivl_breadth}
\end{table}

\paragraph{\AlgGCIQL{}.}
On \EnvAntMazeM{}, \(n=3\) improves \AlgGCIQL{} on every statistic
(\cref{tab:nstep_gciql_breadth}): peak success rises from \(0.70\) to
\(0.79\), mean success from \(0.34\) to \(0.59\), and the near-peak region
widens from \(3\,\%\) to \(20\,\%\) of configurations. \AlgGCIQL{}'s narrow
peak in the main study is therefore partly caused by the one-step
target.

\begin{table}[h]
\centering
\small
\begin{tabular}{lrrrrr}
\toprule
\(n\) & Configs & Seeds & Max & Mean & \(\rho_{0.9}\) \\
\midrule
1 & 64 & 5 & 0.700 & 0.336 & 0.031 \\
3 & 64 & 3 & 0.787 & 0.592 & 0.203 \\
\bottomrule
\end{tabular}
\vspace{0.5em}
\caption{\textbf{\AlgGCIQL{} \(n\)-step breadth, \EnvAntMazeM{}, final
phase.} \(n=3\) both raises peak success and broadens the near-peak region
(\(\rho_{0.9}\)) relative to \(n=1\) at the same \(64\)-configuration
budget.}
\label{tab:nstep_gciql_breadth}
\end{table}

\paragraph{Diagnostics.}
We recompute the advantage diagnostics on the same fixed validation
batches. For \AlgGCIVL{}, longer horizons raise \FRAUC{}, \MRR{}, and Gap
in both environments (e.g.\ \EnvCube{} Gap \(0.16\to0.43\to0.74\)) even
though \EnvCube{} success falls, and per-configuration diagnostic changes
are not correlated with success changes (all \(95\,\%\) CIs include zero).
For \AlgGCIQL{}, average diagnostics barely change
(\FRAUC{} \(0.471\to0.477\); the \(n=1\) values are the five-seed values of
\cref{tab:compact_diagnostics}), but per-configuration changes in \MRR{} do
track changes in success (\(\rho=+0.50\), \(95\,\%\) CI \([+0.31,+0.66]\)).
For \AlgGCIVL{}, better \FAD{} thus does not by itself
imply better extracted policies; for \AlgGCIQL{}, the two change together
across configurations.

\paragraph{Scope.}
The main study uses the standard one-step targets of the reference
implementations. The \(n\)-step results show that the target horizon is a
further design choice that can shape the landscape. Since each method was tested
on different environments, we read them as per-setting observations.

\subsection{Training protocol: straight-through vs.\ checkpoint-conditioned}
\label{app:straight-vs-phased}

The main study starts each phase from the checkpoint of the best
configuration in the previous phase
(\cref{sec:appendix:landscape-datasets}). Since every configuration
inherits a good starting point, this checkpoint conditioning could in
principle make landscapes look broader than they are. We test whether the
breadth differences between methods depend on it. Checkpoint conditioning
raises absolute breadth and shifts where the best configurations lie,
while the contrast between methods remains.

We train each configuration of \AlgGCIQL{} and \AlgQRL{} on
\EnvAntMazeM{} uninterrupted for the full budget (straight-through) and
read out success at the same four evaluation steps as the main study,
using \(3\) seeds per configuration for both protocols.
\AlgQRL{} is broader than \AlgGCIQL{} under both protocols
(\cref{tab:a1_breadth}), both near the peak (\(\rho_{0.9}\): \(0.359\)
vs.\ \(0.047\) checkpoint-conditioned; \(0.141\) vs.\ \(0.016\)
straight-through) and at the absolute \(0.25\) threshold (\(0.906\) vs.\
\(0.812\); \(0.656\) vs.\ \(0.422\)). Absolute breadth is lower for both
methods under straight-through training, so breadth values should
only be compared between methods within the same protocol.

\begin{table}[h]
\centering
\small
\begin{tabular}{llrrr}
\toprule
Algorithm & Protocol & \(\rho_{0.9}\) & \(\rho_{0.8}\) & \(\rho_{\mathrm{abs}}(0.25)\) \\
\midrule
\AlgGCIQL{} & checkpoint-conditioned & 0.047 (3/64)  & 0.047 (3/64)  & 0.812 \\
\AlgGCIQL{} & straight-through        & 0.016 (1/64)  & 0.016 (1/64)  & 0.422 \\
\AlgQRL{}   & checkpoint-conditioned & 0.359 (23/64) & 0.703 (45/64) & 0.906 \\
\AlgQRL{}   & straight-through        & 0.141 (9/64)  & 0.281 (18/64) & 0.656 \\
\bottomrule
\end{tabular}
\vspace{0.5em}
\caption{\textbf{Breadth under the two training protocols, \EnvAntMazeM{}.}
Parenthesized counts give the number of configurations above the
threshold. Breadth is comparable between methods within a protocol;
absolute success rates are not directly equivalent across protocols.}
\label{tab:a1_breadth}
\end{table}

\begin{figure}[htb]
\centering
\begin{tabular}{@{}cccc@{}}
\includegraphics[width=0.23\linewidth]{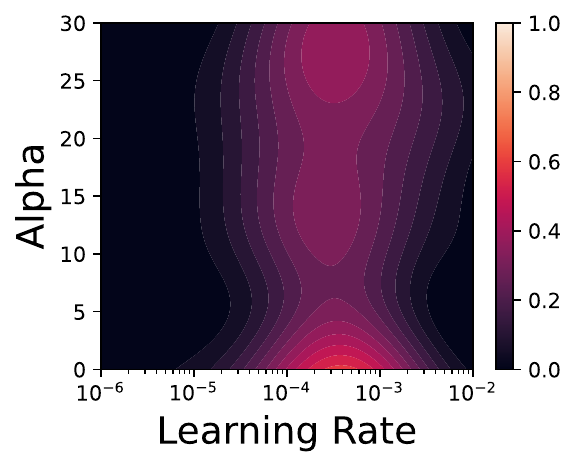} &
\includegraphics[width=0.23\linewidth]{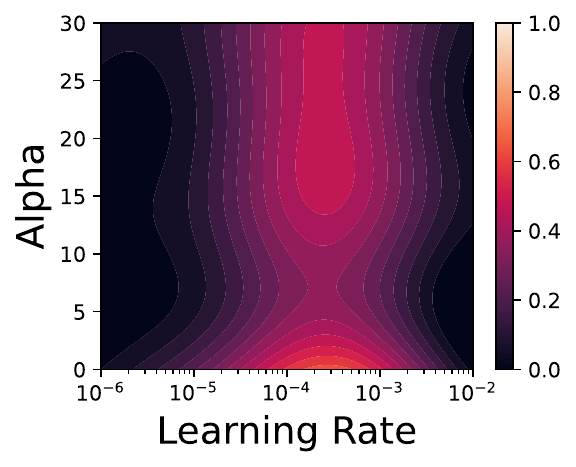} &
\includegraphics[width=0.23\linewidth]{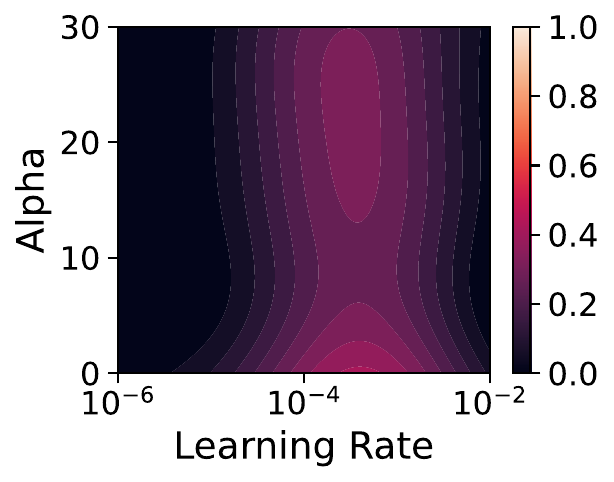} &
\includegraphics[width=0.23\linewidth]{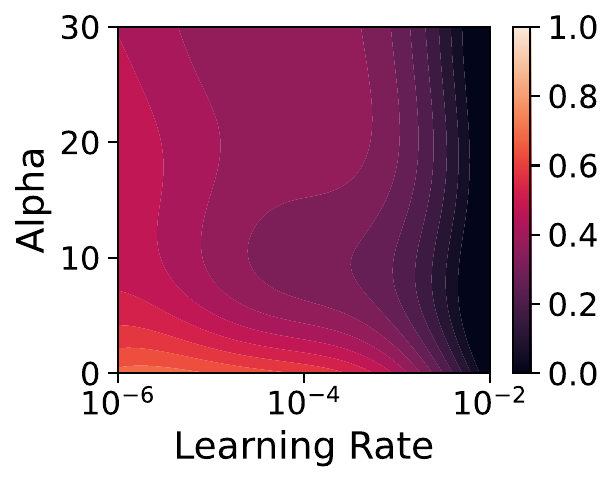} \\
\footnotesize (a) straight, step 1 & \footnotesize (b) straight, final & \footnotesize (c) phased, step 1 & \footnotesize (d) phased, final \\[4pt]
\multicolumn{2}{c}{\includegraphics[width=0.46\linewidth]{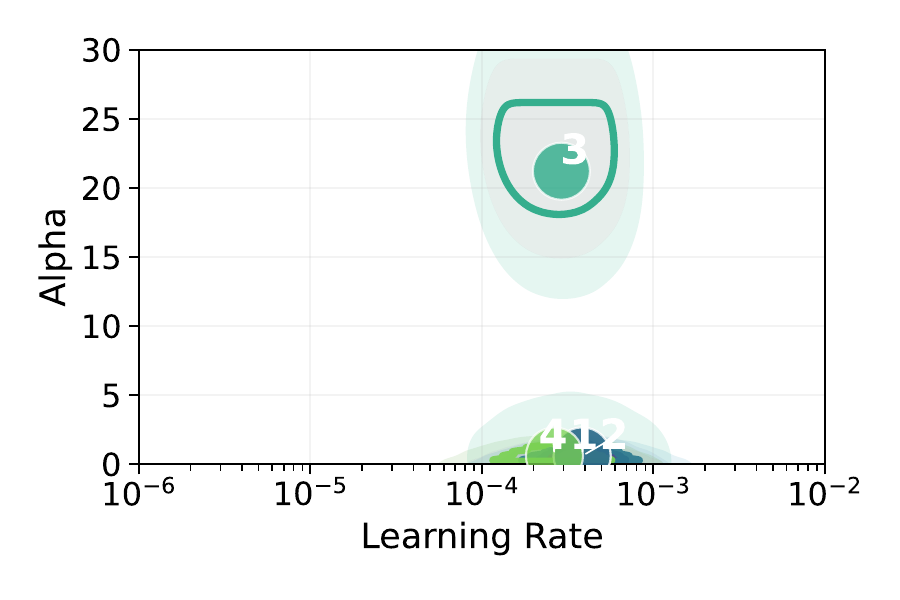}} &
\multicolumn{2}{c}{\includegraphics[width=0.46\linewidth]{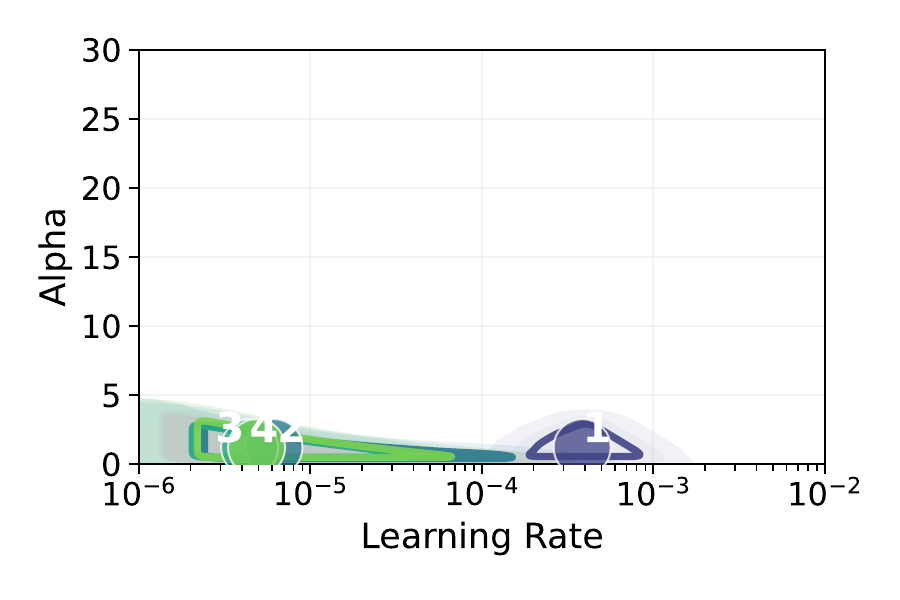}} \\
\multicolumn{2}{c}{\footnotesize (e) straight mobility (top-20\,\%, 4 steps)} & \multicolumn{2}{c}{\footnotesize (f) phased mobility (top-20\,\%, 4 steps)}
\end{tabular}
\caption{\textbf{\AlgGCIQL{}, straight-through vs.\ checkpoint-conditioned,
\EnvAntMazeM{}.} Landscapes at step 1 and final step under each protocol,
and the top-20\,\% mobility footprint across all four steps.}
\label{fig:a1_gciql}
\end{figure}

\begin{figure}[htb]
\centering
\begin{tabular}{@{}cccc@{}}
\includegraphics[width=0.23\linewidth]{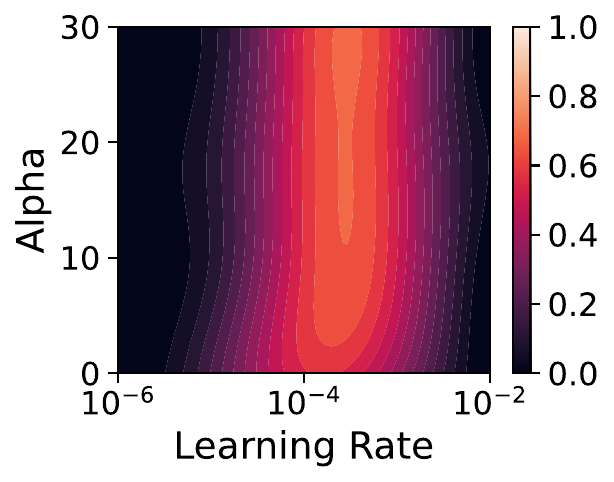} &
\includegraphics[width=0.23\linewidth]{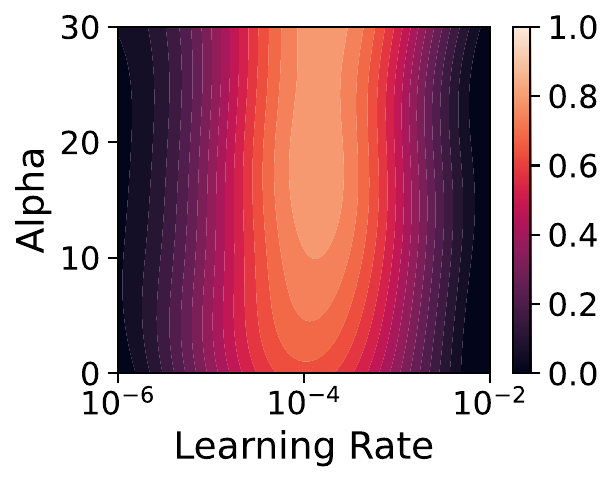} &
\includegraphics[width=0.23\linewidth]{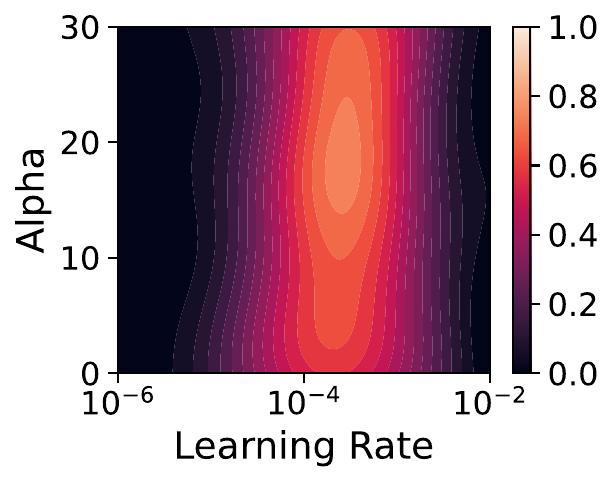} &
\includegraphics[width=0.23\linewidth]{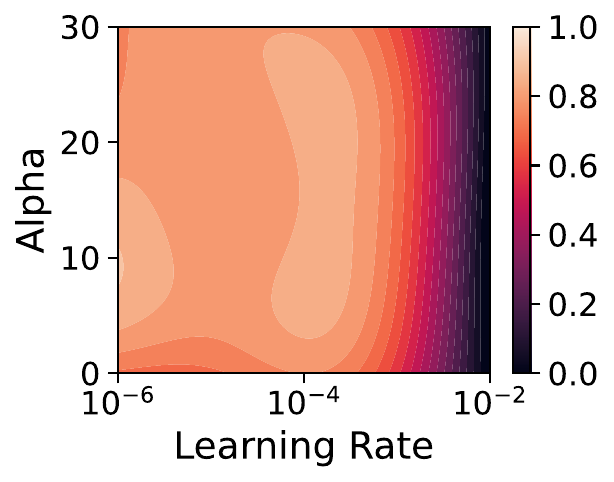} \\
\footnotesize (a) straight, step 1 & \footnotesize (b) straight, final & \footnotesize (c) phased, step 1 & \footnotesize (d) phased, final \\[4pt]
\multicolumn{2}{c}{\includegraphics[width=0.46\linewidth]{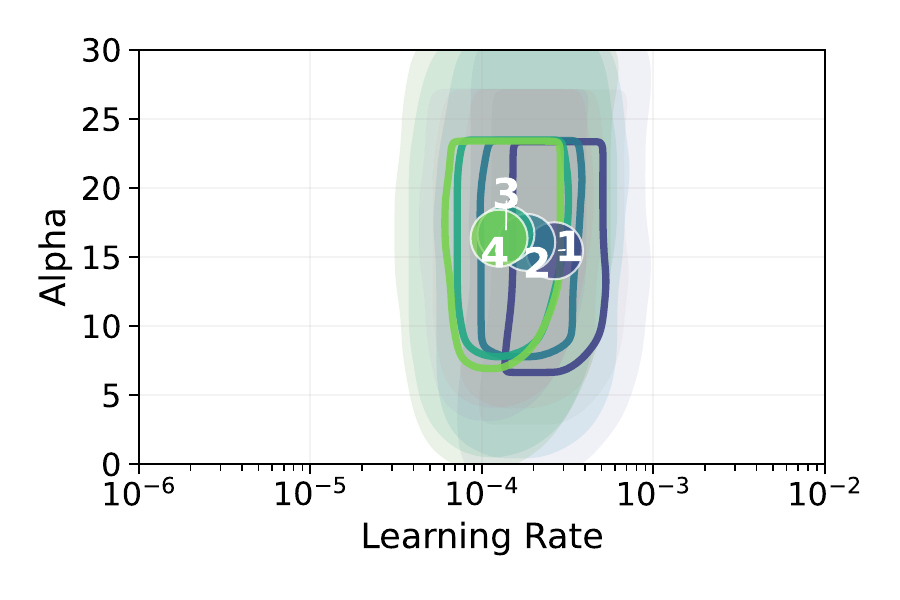}} &
\multicolumn{2}{c}{\includegraphics[width=0.46\linewidth]{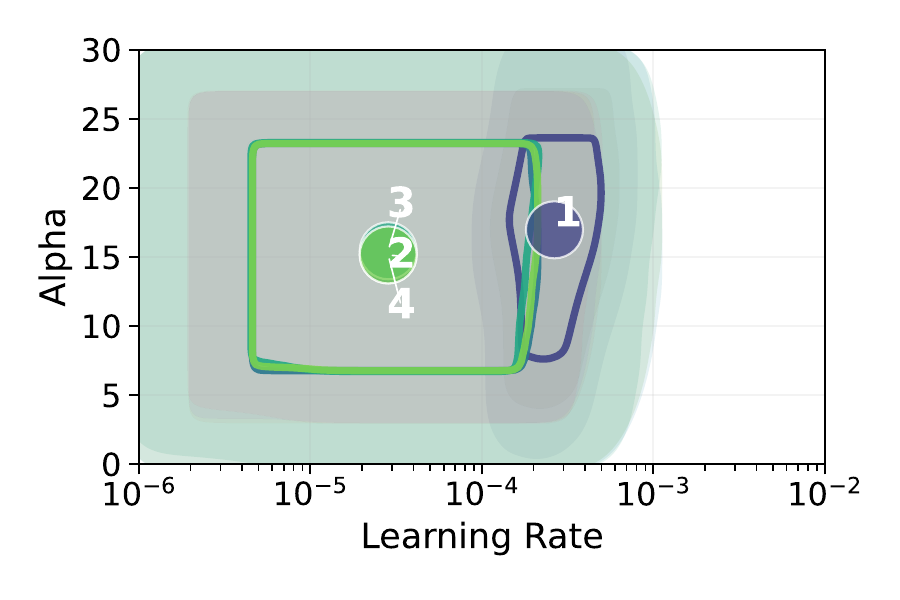}} \\
\multicolumn{2}{c}{\footnotesize (e) straight mobility (top-20\,\%, 4 steps)} & \multicolumn{2}{c}{\footnotesize (f) phased mobility (top-20\,\%, 4 steps)}
\end{tabular}
\caption{\textbf{\AlgQRL{}, straight-through vs.\ checkpoint-conditioned,
\EnvAntMazeM{}.} Landscapes at step 1 and final step under each protocol,
and the top-20\,\% mobility footprint across all four steps.}
\label{fig:a1_qrl}
\end{figure}

\Cref{fig:a1_gciql,fig:a1_qrl} show the phase-1 and final-step landscapes
under both protocols, together with the mobility footprint across all four
evaluation steps. At step 1, both protocols place the top-20\,\% region
in a similar part of the \((\eta,\alpha)\) plane. For \AlgGCIQL{}, the
protocols then diverge: under straight-through training the region stays
near \(\eta\approx3\times10^{-4}\), \(\alpha\approx0\) (apart from an
outlier at step 3 with \(\alpha\approx22\)), while under
checkpoint-conditioned training it moves to \(\eta\approx3\times10^{-6}\).
For \AlgQRL{}, the straight-through region stays tightly around
\(\eta\approx1\)--\(3\times10^{-4}\), while the checkpoint-conditioned
region shifts to lower learning rates and widens after step 1; the two
regions overlap at all steps.

All landscapes in this subsection use \(3\) seeds per configuration
(vs.\ \(5\) in the main study) and are therefore noisier.

\section{Generalization Beyond the Shared-\AWR{}, Single-Dataset Setup}
\label{app:generalization}

The main study evaluates all four algorithms under one dataset composition (navigate) and one flat extractor (\AWR{}), per \cref{sec:method}. 
This appendix tests how far the resulting patterns extend: first to a harder, stitching-required dataset composition, and then to alternative policy-extraction mechanisms in place of \AWR{}.

\subsection{Dataset composition: stitch vs.\ navigate}
\label{app:stitch-vs-navigate}

We compare the main navigate-dataset experiment on \EnvAntMazeM{} against a stitch-composition dataset, for all four algorithms, at each algorithm's own final-phase budget (\(3\) seeds, capped to match the stitch experimental conditions). 
\Cref{tab:a8_summary} reports peak and mean success under each dataset together with the Spearman correlation across the \(64\) paired configurations. 
The stitch runs were trained without phase resets, so the comparison changes both dataset and protocol; since the protocol alone reorders \AlgGCIQL{}'s configurations (\cref{app:straight-vs-phased}), its near-zero correlation (\(-0.12\)) should not be attributed to the dataset alone.

\begin{table}[h]
\centering
\small
\begin{tabular}{lrrrl}
\toprule
Algorithm & Navigate (peak/mean) & Stitch (peak/mean) & Spearman & 95\,\% CI \\
\midrule
\AlgGCIVL{} & 0.740 / 0.494 & 0.480 / 0.246 & \(+0.587\) & \([+0.373,+0.738]\) \\
\AlgQRL{}   & 0.867 / 0.672 & 0.567 / 0.303 & \(+0.439\) & \([+0.208,+0.641]\) \\
\AlgCRL{}   & 0.847 / 0.666 & 0.507 / 0.195 & \(+0.270\) & \([-0.005,+0.516]\) \\
\AlgGCIQL{} & 0.700 / 0.336 & 0.473 / 0.122 & \(-0.117\) & \([-0.415,+0.181]\) \\
\bottomrule
\end{tabular}
\vspace{0.5em}
\caption{\textbf{Navigate vs.\ stitch dataset, \EnvAntMazeM{}.} Every algorithm loses roughly a third of its navigate-dataset peak on the stitch-composition dataset. Only \AlgGCIVL{} and \AlgQRL{} have Spearman CIs that exclude zero; \AlgCRL{}'s CI spans zero (positive but not distinguishable from no ranking agreement); \AlgGCIQL{} is the lone algorithm with a negative point estimate. Navigate columns use three seeds (capped to match the stitch experimental conditions), so navigate peaks here equal or slightly exceed the five-seed peaks of the main results table.} \label{tab:a8_summary}
\end{table}

\begin{figure}[htb]
\centering
\begin{tabular}{@{}cc@{}}
\includegraphics[width=0.45\linewidth,height=0.19\textheight,keepaspectratio]{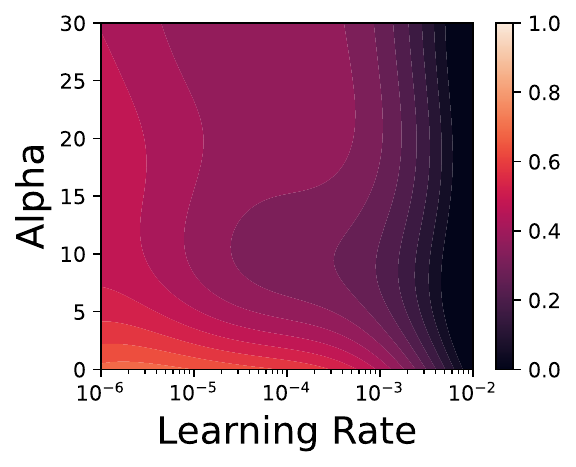} &
\includegraphics[width=0.45\linewidth,height=0.19\textheight,keepaspectratio]{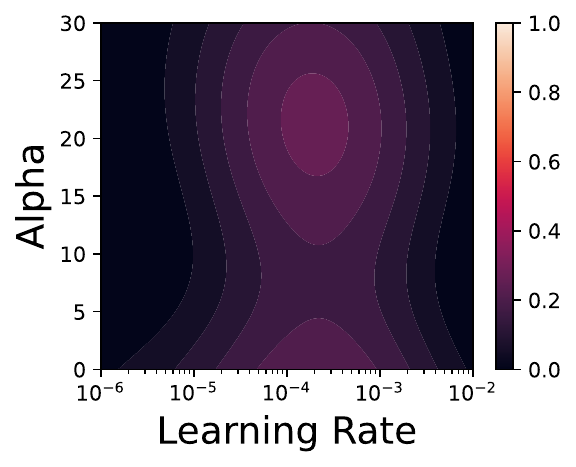} \\
\multicolumn{2}{c}{\footnotesize (a) \AlgGCIQL{}: navigate (left), stitch (right)} \\[2pt]
\includegraphics[width=0.45\linewidth,height=0.19\textheight,keepaspectratio]{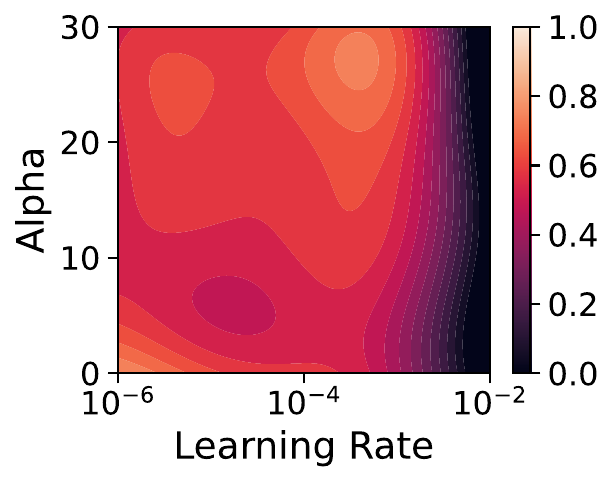} &
\includegraphics[width=0.45\linewidth,height=0.19\textheight,keepaspectratio]{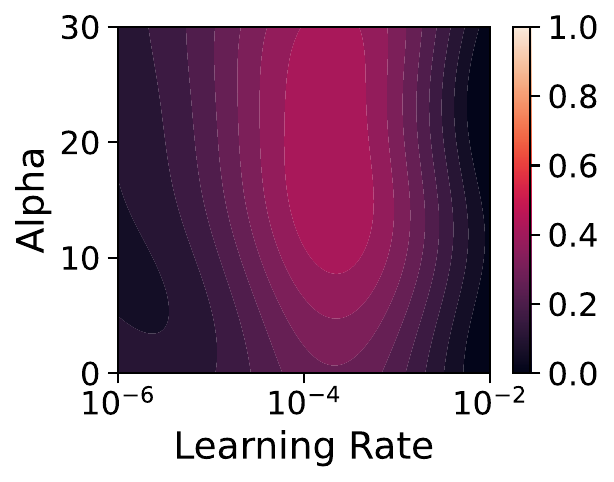} \\
\multicolumn{2}{c}{\footnotesize (b) \AlgGCIVL{}: navigate (left), stitch (right)} \\[2pt]
\includegraphics[width=0.45\linewidth,height=0.19\textheight,keepaspectratio]{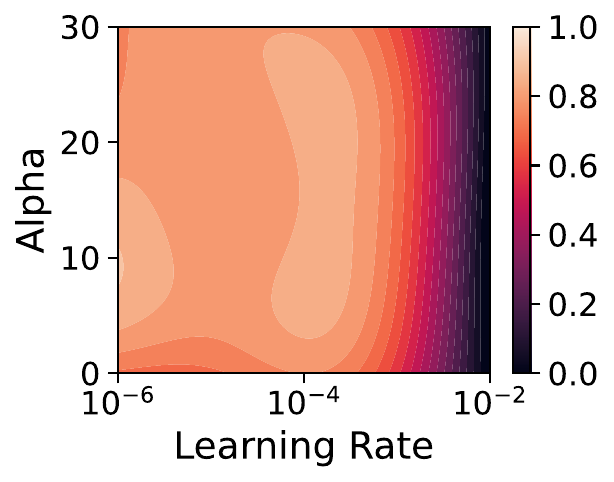} &
\includegraphics[width=0.45\linewidth,height=0.19\textheight,keepaspectratio]{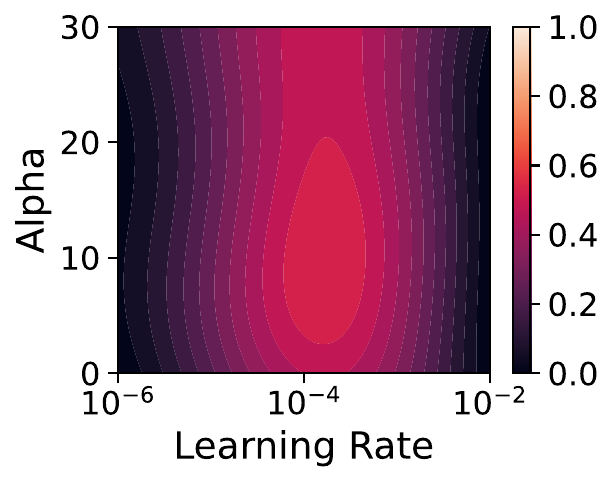} \\
\multicolumn{2}{c}{\footnotesize (c) \AlgQRL{}: navigate (left), stitch (right)} \\[2pt]
\includegraphics[width=0.45\linewidth,height=0.19\textheight,keepaspectratio]{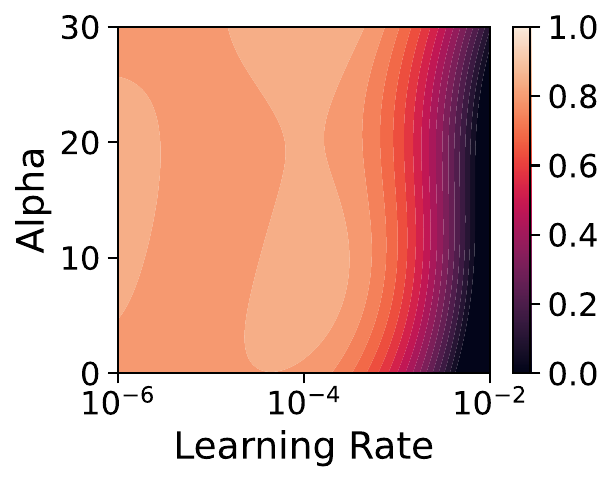} &
\includegraphics[width=0.45\linewidth,height=0.19\textheight,keepaspectratio]{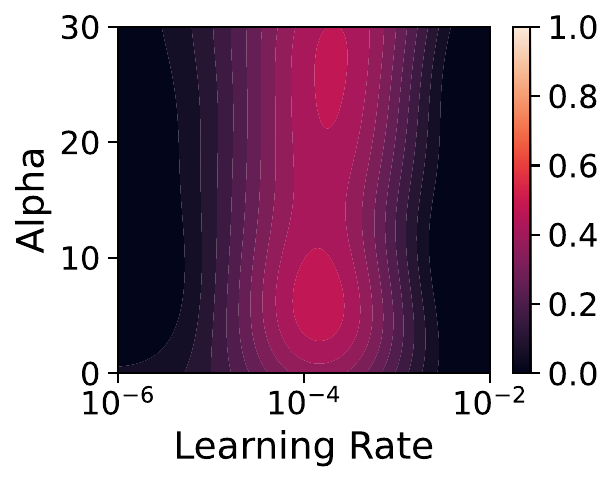} \\
\multicolumn{2}{c}{\footnotesize (d) \AlgCRL{}: navigate (left), stitch (right)}
\end{tabular}
\caption{\textbf{Navigate vs.\ stitch landscapes, \EnvAntMazeM{}.} Each row
is one algorithm's final-phase landscape under the navigate dataset (left)
and the stitch-composition dataset (right).}
\label{fig:a8_landscapes}
\end{figure}

\begin{figure}[htb]
\centering
\begin{tabular}{@{}cc@{}}
\includegraphics[width=0.45\linewidth]{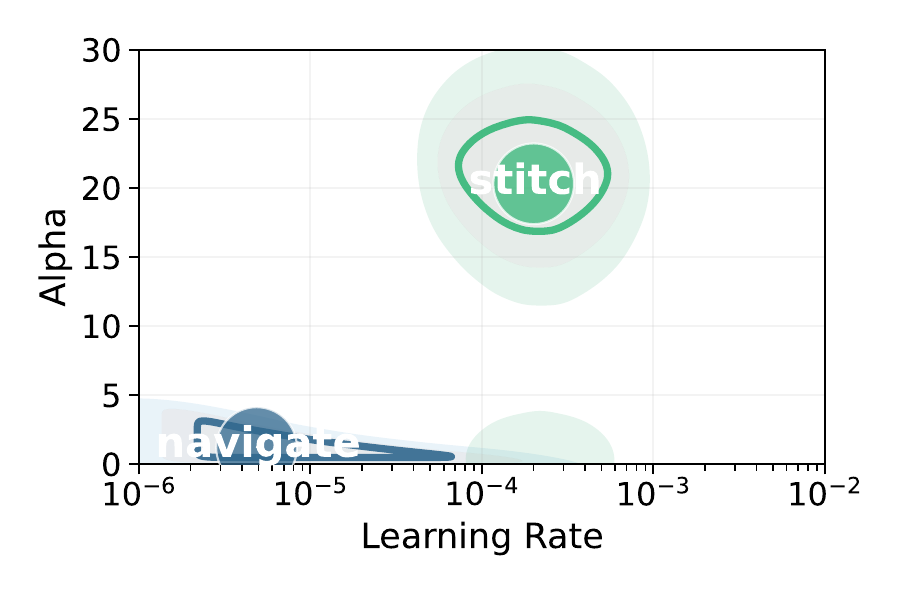} &
\includegraphics[width=0.45\linewidth]{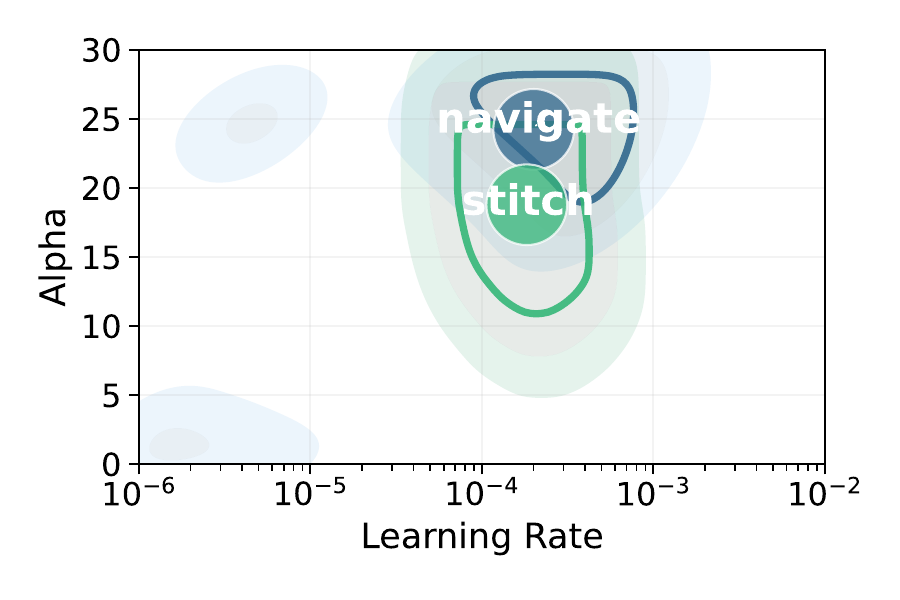} \\
\footnotesize (a) \AlgGCIQL{} & \footnotesize (b) \AlgGCIVL{} \\[4pt]
\includegraphics[width=0.45\linewidth]{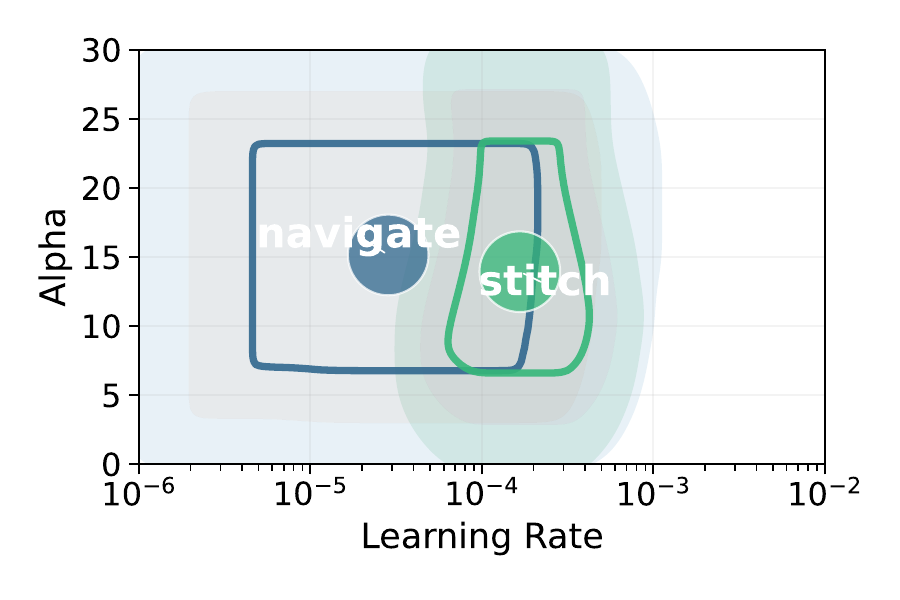} &
\includegraphics[width=0.45\linewidth]{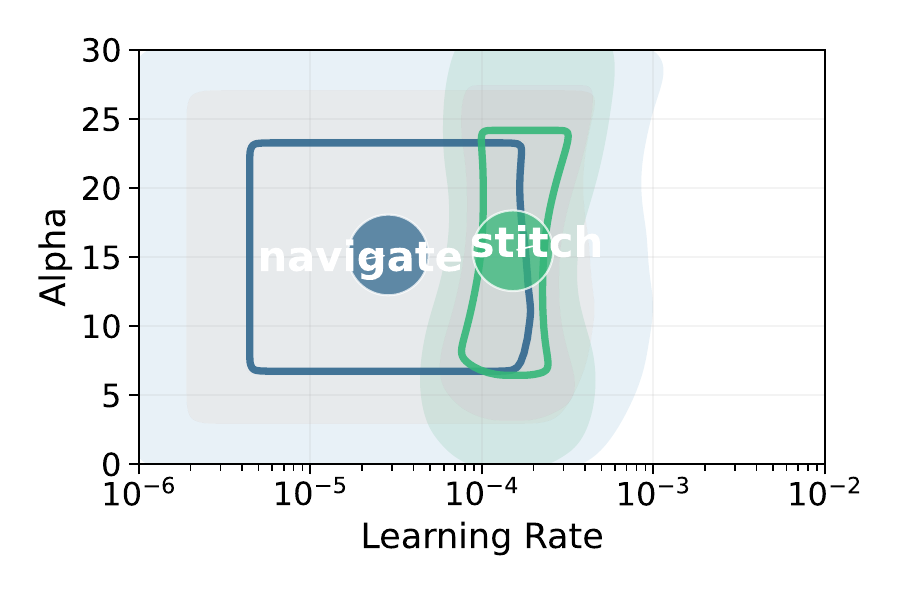} \\
\footnotesize (c) \AlgQRL{} & \footnotesize (d) \AlgCRL{}
\end{tabular}
\caption{\textbf{Navigate\(\to\)stitch mobility footprint, \EnvAntMazeM{}.}
Two-layer top-20\,\% mobility KDE showing how the top-20\,\% region moves when
the dataset changes from navigate to stitch composition.}
\label{fig:a8_mobility}
\end{figure}

\Cref{fig:a8_landscapes} shows the navigate and stitch landscapes for all four algorithms, and \cref{fig:a8_mobility} shows the corresponding two-layer (navigate\(\to\)stitch) mobility footprint per algorithm. 
Both datasets are evaluated at the final budget only, so we report only the single-transition mobility footprint and make no phase-mobility claim for this ablation.

\subsection{Alternative policy extractors}
\label{app:extractor-ablations}

The main study fixes \AWR{} as the shared policy extractor for all four
algorithms. Here we test whether the landscape patterns reported in
\cref{sec:experiments} are specific to \AWR{}, using four alternative
extraction mechanisms on \EnvAntMazeM{}, three of which we also run on
\EnvCube{}.

\paragraph{DDPG+BC extraction, \EnvCube{} (\AlgGCIQL{}, \AlgCRL{},
\AlgQRL{}).}
We replace the shared \AWR{} extractor with a DDPG+BC
actor~\citep{fujimoto-neurips21a} (deterministic policy trained against the learned critic with a behavioral
cloning regularizer) on \EnvCube{}, using independent Sobol draws over the
learning rate and a BC coefficient per algorithm (\(5\) seeds; the BC
coefficient plays the role of \(\alpha\) but is not directly comparable to
the \AWR{} temperature). Because the two grids are independently sampled,
per-configuration values are not paired across extractors, so we report
success rates and breadth summaries only, with no Spearman correlation between
them.
\Cref{tab:a3_summary} reports each algorithm's peak/mean success under
both extractors together with the training-budget ratio
(DDPG+BC final-phase steps \(\div\) \AWR{} final-phase steps), which flags
whether an apparent gain could instead be a budget confound.

\begin{table}[h]
\centering
\small
\begin{tabular}{lrrrl}
\toprule
Algorithm & DDPG+BC (peak/mean) & \AWR{} (peak/mean) & Budget ratio & Verdict \\
\midrule
\AlgGCIQL{} & 0.920 / 0.493 & 0.773 / 0.527 & \(0.86\times\) & conservative \\
\AlgCRL{}   & 0.353 / 0.129 & 0.280 / 0.118 & \(3.15\times\) & confounded \\
\AlgQRL{}   & 0.120 / 0.040 & 0.180 / 0.066 & \(1.11\times\) & conservative \\
\bottomrule
\end{tabular}
\vspace{0.5em}
\caption{\textbf{DDPG+BC vs.\ \AWR{} extraction, \EnvCube{}.} Budget ratio is DDPG+BC over \AWR{} final-phase training steps. 
\emph{conservative} means the budget difference works against the observed direction: 
\AlgGCIQL{}'s higher DDPG+BC peak uses \(14\%\) fewer steps (this argument does not apply to its lower mean), and \AlgQRL{}'s lower DDPG+BC scores use more steps. 
\emph{confounded} means \AlgCRL{}'s DDPG+BC runs use over \(3\times\) the \AWR{} budget; at \(0.79\times\) the budget its peak advantage remains (\(0.41\) vs.\ \(0.28\)) but its mean advantage does not.
Configurations are independent draws on each side, so only distribution-level comparisons apply.}
\label{tab:a3_summary}
\end{table}

The \EnvCube{} ordering of the main results (\AlgGCIQL{} ahead of
\AlgCRL{} ahead of \AlgQRL{}) also persists under DDPG+BC: \AlgGCIQL{}
combines a high peak (\(0.920\)) with substantial absolute breadth
(\(\rho_{\mathrm{abs}}(0.25)=0.766\), \(\rho_{\mathrm{abs}}(0.5)=0.609\))
while remaining narrow near its own peak (\(\rho_{0.9}=0.094\)), whereas
\AlgCRL{}'s peak (\(0.353\)) stays below the \(0.5\) threshold
(\(\rho_{\mathrm{abs}}(0.25)=0.203\)) and \AlgQRL{}'s peak (\(0.120\))
stays below both thresholds.

\begin{figure}[htb]
\centering
\begin{tabular}{@{}cc@{}}
\includegraphics[width=0.45\linewidth]{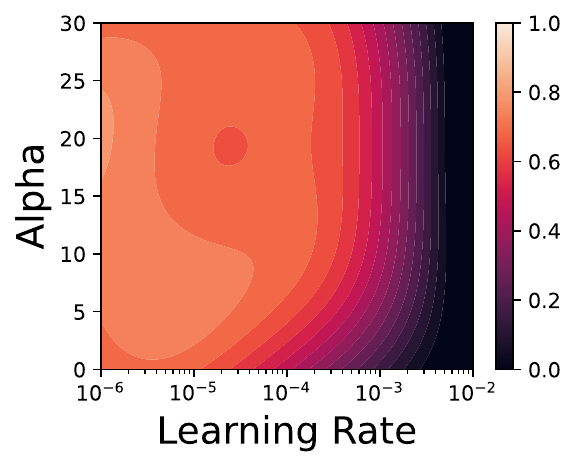} &
\includegraphics[width=0.45\linewidth]{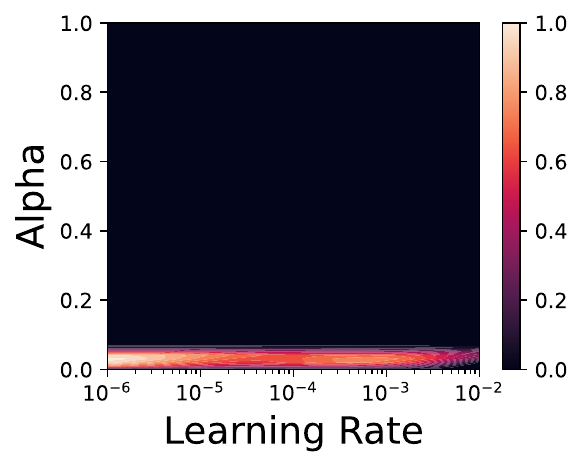} \\
\multicolumn{2}{c}{\footnotesize (a) \AlgGCIQL{}: \AWR{} (left), DDPG+BC (right)} \\[4pt]
\includegraphics[width=0.45\linewidth]{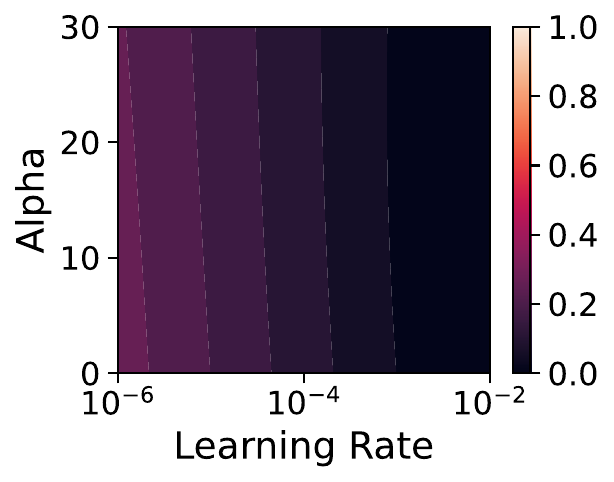} &
\includegraphics[width=0.45\linewidth]{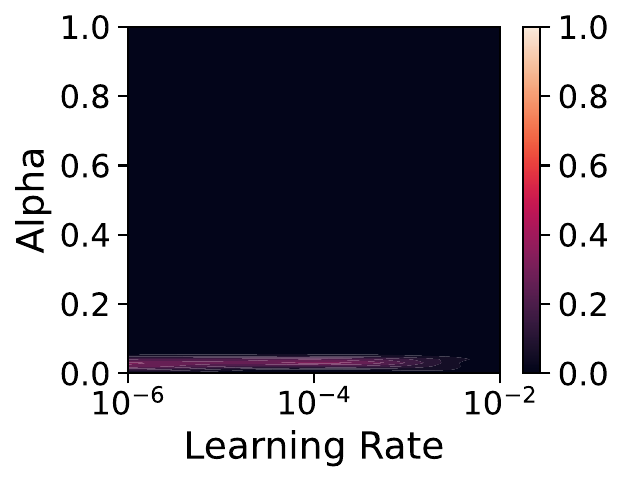} \\
\multicolumn{2}{c}{\footnotesize (b) \AlgCRL{}: \AWR{} (left), DDPG+BC (right)} \\[4pt]
\includegraphics[width=0.45\linewidth]{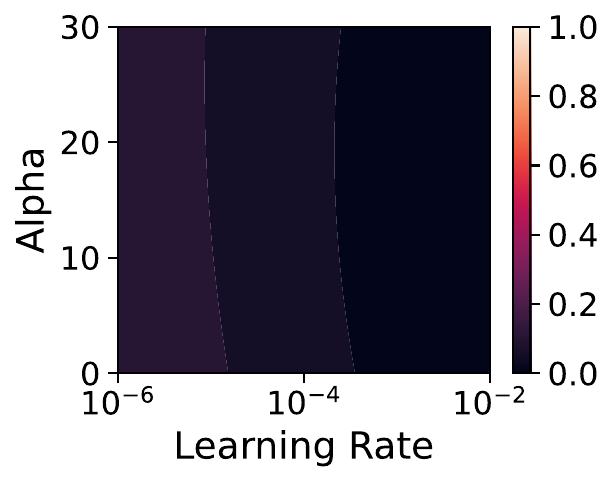} &
\includegraphics[width=0.45\linewidth]{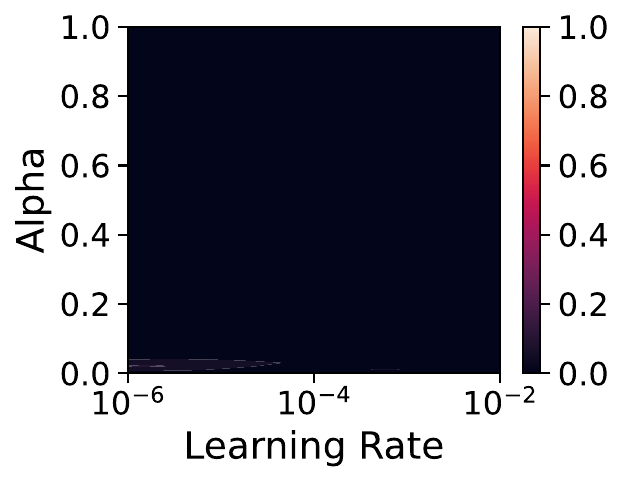} \\
\multicolumn{2}{c}{\footnotesize (c) \AlgQRL{}: \AWR{} (left), DDPG+BC (right)}
\end{tabular}
\caption{\textbf{\AWR{} vs.\ DDPG+BC landscapes, \EnvCube{}.} Independent
Sobol draws per extractor; compare breadth class and success rates only.}
\label{fig:a3_landscapes}
\end{figure}

\paragraph{DDPG+BC extraction, \EnvAntMazeM{} (\AlgGCIQL{}, \AlgCRL{},
\AlgQRL{}).}
OGBench provides DDPG+BC variants for \AlgGCIQL{}, \AlgCRL{}, and
\AlgQRL{} but not for \AlgGCIVL{}, whose value-only architecture lacks an
action-conditioned critic for the DDPG+BC actor; \AlgQRL{}'s variant
additionally introduces a latent dynamics model for action gradients.
\Cref{tab:a4_summary} compares the two extractors on \EnvAntMazeM{}.
Near the peak, the breadth class carries over for two of the three
algorithms: \AlgGCIQL{} remains narrow under both extractors (\(\rho_{0.9}\)
\(0.03\to0.17\)), \AlgCRL{} remains broad (\(0.66\to0.81\)), and
\AlgQRL{} flips from broad to narrow (\(0.59\to0.02\)). \AlgQRL{}'s critic
has no action input, so its DDPG+BC actor receives action gradients only
through the learned latent dynamics model; the flip likely reflects this
indirect action gradient. At the absolute
\(0.25\) threshold, all three stay broad under both extractors
(\(\rho_{\mathrm{abs}}(0.25)\) between \(0.80\) and \(0.92\)). These comparisons are qualitative:
grids are independently sampled, and the actor-update budgets differ
(\(1.0\times\), \(2.1\times\), and \(3.5\times\) the \AWR{} budgets for
\AlgGCIQL{}, \AlgQRL{}, and \AlgCRL{}); \AlgGCIQL{}'s budget ratio is
effectively matched (\(1.011\times\), verdict \emph{clean}), so its
peak/mean gain (\(0.70/0.336\to0.82/0.557\)) is attributable to the
extractor.

\begin{table}[h]
\centering
\small
\setlength{\tabcolsep}{4pt}
\begin{tabular}{llrrrr}
\toprule
Algorithm & Extractor & Peak & Mean & \(\rho_{0.9}\) & \(\rho_{\mathrm{abs}}(0.25)\) \\
\midrule
\AlgGCIQL{} & \AWR{}     & 0.70 & 0.336 & 0.031 & 0.828 \\
\AlgGCIQL{} & DDPG+BC    & 0.82 & 0.557 & 0.172 & 0.906 \\
\midrule
\AlgCRL{}   & \AWR{}     & 0.83 & 0.669 & 0.656 & 0.891 \\
\AlgCRL{}   & DDPG+BC    & 0.99 & 0.872 & 0.812 & 0.922 \\
\midrule
\AlgQRL{}   & \AWR{}     & 0.84 & 0.663 & 0.594 & 0.906 \\
\AlgQRL{}   & DDPG+BC    & 0.58 & 0.332 & 0.016 & 0.797 \\
\bottomrule
\end{tabular}
\vspace{0.5em}
\caption{\textbf{\AWR{} vs.\ DDPG+BC extraction, \EnvAntMazeM{}.} Final
phase, independent Sobol draws per extractor (five seeds).
Actor-update budget ratios relative to \AWR{}: \AlgGCIQL{}
\(1.011\times\) (clean), \AlgQRL{} \(2.112\times\) (conservative),
\AlgCRL{} \(3.482\times\) (confounded --- do not attribute its gain to
the extractor alone).}
\label{tab:a4_summary}
\end{table}

\paragraph{Critic-guided reranking of frozen \AWR{} checkpoints
(\AlgGCIQL{}, \AlgCRL{}; \EnvAntMazeM{} and \EnvCube{}).}
Rather than retraining under a new extractor, this ablation reranks the
frozen \AWR{}-trained action candidates at each checkpoint using the
learned critic and measures whether the induced ordering agrees with the
original \AWR{} success ordering. \Cref{tab:a6_summary} reports the
Spearman correlation between reranked and original orderings on
\EnvAntMazeM{}; on \EnvCube{}, the reranked policy is at floor (max
success \(0.02\) for \AlgCRL{}, \(0.06\) for \AlgGCIQL{}), so no
conclusion can be drawn there.

\begin{table}[h]
\centering
\small
\begin{tabular}{lrl}
\toprule
Algorithm & Spearman \(\rho\) & 95\,\% CI \\
\midrule
\AlgCRL{}   & 0.549 & \([0.322, 0.735]\) \\
\AlgGCIQL{} & 0.607 & \([0.400, 0.760]\) \\
\bottomrule
\end{tabular}
\vspace{0.5em}
\caption{\textbf{Critic-guided reranking of frozen \AWR{} checkpoints,
\EnvAntMazeM{}.} The \AWR{} ordering largely survives reranking by the
learned critic. Correlations pool all checkpoints, so \(\rho\) also
reflects differences between phases. This is a reduced
2-seed/5-episode protocol; its own
test--retest reliability caps \AlgCRL{}'s achievable correlation at
\(0.539\) and \AlgGCIQL{}'s at \(0.795\). \AlgCRL{}'s \(\rho=0.549\) is at its
ceiling (\(0.539\)), so this protocol cannot resolve how strong its
association is. \AlgGCIQL{}'s \(0.607\) lies below its ceiling
(\(0.795\)), which indicates a partial match.}
\label{tab:a6_summary}
\end{table}

\paragraph{BC-proposal reranking without \AWR{} (SfBC-style).}
This ablation, in the style of SfBC~\citep{chen-iclr23a}, removes \AWR{}
entirely: candidate actions are proposed by a
goal-conditioned behavioral-cloning (GCBC) policy and selected by the
learned critic, with no \AWR{} weighting at any stage. On \EnvAntMazeM{},
the reranked ordering (pooled over all checkpoints) again correlates
positively with the original
\AWR{} ordering for both algorithms (\AlgCRL{} \(\rho=+0.305\),
\(95\,\%\) CI \([+0.031,+0.542]\); \AlgGCIQL{} \(\rho=+0.401\),
\(95\,\%\) CI \([+0.141,+0.616]\)); both CIs exclude zero, but
\AlgCRL{}'s lower bound is barely above zero, so its match is weak. On \EnvCube{}, increasing the candidate count from \(N=32\)
to \(N=256\) does not move the ceiling (\AlgCRL{} \(0.02\to0.02\),
\AlgGCIQL{} \(0.08\to0.08\)): \EnvCube{} stays inconclusive because the
GCBC proposal itself is floored at \(0.033\) mean success, which limits
any critic-based selection. On \EnvAntMazeM{}, \AlgCRL{}'s peak success under
this extractor (\(0.800\)) is well above the GCBC proposal alone
(\(0.367\)). \AlgGCIQL{}'s peak (\(0.260\)) and mean (\(0.139\)) are
\emph{below} that baseline, so its critic misranks the BC candidates.

\paragraph{FQL (flow-based, no \AWR{}).}
As a further extraction-mechanism check, we evaluate FQL~\citep{park-icml25a} (a flow-based behavior-constrained actor) on \EnvAntMazeM{} over an independent \(32\)-configuration Sobol draw on \((\eta,\alpha)\), where \(\alpha\) is FQL's distillation weight (sampled log-uniformly in \([1,300]\)) rather than an \AWR{} temperature (\(3\) seeds, seed mean, trained for \AlgGCIQL{}'s final-phase budget of \(820{,}277\) steps without phase resets).
Since the configurations are independent of the main sweep, we report success rates only. 
FQL reaches peak/mean success \(0.553/0.297\). 
Its actor objective, \(\alpha\)-related quantities and configuration count differ from the shared-\AWR{} sweeps, so this serves only as an additional reference point.

Taken together, these alternative extractors indicate that the landscape
patterns largely carry over beyond \AWR{}. Under DDPG+BC, the near-peak
breadth classes of the algorithms with action-conditioned critics carry
over (\AlgGCIQL{} narrow, \AlgCRL{} broad), and the \EnvCube{} ordering is
preserved; \AlgQRL{} is the exception, likely because of its indirect
action gradient (\Cref{tab:a4_summary}). Both critic-guided and
BC-proposal reranking agree positively with the \AWR{} ordering of
checkpoints, although these pooled correlations also reflect differences
between phases, and the BC-proposal extractor depends on the configuration
only through its critic. Two cases call for care: \AlgCRL{}'s DDPG+BC gain
is confounded by training budget, and \AlgGCIQL{}'s critic ranks BC
candidates below the GCBC baseline on \EnvAntMazeM{}. Extending these
checks beyond the two environments is a natural next step.

\end{document}